%% file: ms.tex
\pdfoutput=1
\documentclass{article}
\pdfoutput=1

\newcommand{\reviewer}[3]{
	\expandafter\newcommand\csname #1\endcsname[1]{
		\textcolor{#3}{[#2: ##1]}
	}
}
\reviewer{km}{KM}{blue}
\usepackage{mwe}
\usepackage{graphics}
\usepackage{graphicx}
\usepackage{hyperref}
\usepackage[dvipsnames]{xcolor}
\usepackage{algorithm}
\usepackage{algorithmic}

\usepackage{tablefootnote}

\usepackage{booktabs} 
\usepackage{amsmath,amssymb,enumitem,multirow,mathtools}
\usepackage{microtype}
\usepackage{amsmath,amsfonts,amssymb,amsthm,epsfig,epstopdf,titling,url,array}
\usepackage{subfigure}
\usepackage{booktabs} 
\usepackage[dvipsnames]{xcolor}
\usepackage{booktabs} 
\usepackage{makecell}
\usepackage{enumitem}
\usepackage{tablefootnote}
\usepackage{hyperref}
\usepackage[preprint]{neurips_2020}
\title{Model-Augmented $Q$-Learning}

\newcommand{\reward}{\mathcal{R}\xspace}
\newcommand{\transition}{\mathcal{T}\xspace}

\newcommand{\pred}{\mathcal{M}\xspace}
\newcommand{\rewardcoefone}{\zeta_1\xspace}
\newcommand{\rewardcoeftwo}{\zeta_2\xspace}
\newcommand{\losscoefzero}{\xi_1\xspace}
\newcommand{\losscoefone}{\xi_2\xspace}
\newcommand{\losscoeftwo}{\xi_3\xspace}

\newtheorem{theorem}{Theorem}
\usepackage{amsmath, amsthm, amssymb, amsfonts}
\usepackage{xspace}
\usepackage{mathtools}
\author{%
  Youngmin Oh \\
  Samsung Advanced Institute of Technology 
  \and
  Jinwoo Shin \quad Eunho Yang \\Korea Advanced Institute of Science and Technology
 \and
	Sung Ju Hwang\\
  Korea Advanced Institute of Science and Technology\\
 \texttt{sjhwang82@kaist.ac.kr}
}

\begin{document}
\maketitle
\input{0-abstract}
\input{1-introduction}
\input{2-method}
\input{3-exp-arxiv}
\input{5-conclusion}
\bibliography{ms}
\bibliographystyle{authordate1}

\input{6-appendix-arxiv}

\end{document}

%% file: 0-abstract.tex
\begin{abstract}
In recent years, $Q$-learning has become indispensable for model-free reinforcement learning (MFRL). However, it suffers from well-known problems such as under- and overestimation bias of the value, which may adversely affect the policy learning. To resolve this issue, we propose a MFRL framework that is augmented with the components of model-based RL. Specifically, we propose to estimate not only the $Q$-values but also both the transition and the reward with a shared network. We further utilize the estimated reward from the model estimators for $Q$-learning, which promotes interaction between the estimators. We show that the proposed scheme, called Model-augmented $Q$-learning (MQL), obtains a policy-invariant solution which is identical to the solution obtained by learning with true reward. Finally, we also provide a trick to prioritize past experiences in the replay buffer by utilizing model-estimation errors. We experimentally validate MQL built upon state-of-the-art off-policy MFRL methods, and show that MQL largely improves their performance and convergence. The proposed scheme is simple to implement and does not require additional training cost. \end{abstract}

%% file: 1-introduction.tex
\section{Introduction} \label{sec:introduction}

Model-free reinforcement learning (MFRL), which does not learn a model about the environment, has achieved remarkable success on various tasks, e.g., Atari games \citep{deep_q-learning_with_per, deep_q-learning_with_er, rainbow, efficientrainbow}, Starcraft \citep{vinyals2019grandmaster}, and robotic control tasks \citep{sac, td3, ppo, trpo}, due to its simplicity that eliminates the needs of hand-designing or learning a model, and many techniques that enable sample-efficient learning without a model.

Since MFRL algorithms learn agents' policy without environment models, $Q$-learning is critical for policy learning. Some recent MFRL methods such as Rainbow~\citep{rainbow, efficientrainbow} achieve impressive performance with $Q$-learning without learning the policy networks. MFRL algorithms~\citep{sac,sac2,td3} with actor-critic models~\citep{ac}, which learns the policy with the actor, still relies on the critic that estimates the $Q$-values to guide its learning. Here, an accurate estimation of the values of the state-action pairs with $Q$-learning is an essential part of many MFRL algorithms for their success, since if they are over- or underestimated, then the bias will adversely affect the policy learning \citep{doubleqlearning, td3, underestimation}. 

However, the lack of model in MFRL makes them sample-inefficient compared to model-based RL, which can generate samples from the model of the environment that is either manually defined or learned. Many prior works~\citep{mbpo, nagabandi2018neural} have proposed to combine the two methods in a single framework, which first train a model-based controller and collects trajectories to initializes policy network that continually learns using the trajectories from the controller.\footnote{We provide more detailed literature survey in the supplementary material.}

\begin{figure}[!t]
\centering
\begin{tabular}{c}
    \makecell{\includegraphics[width=1\columnwidth]{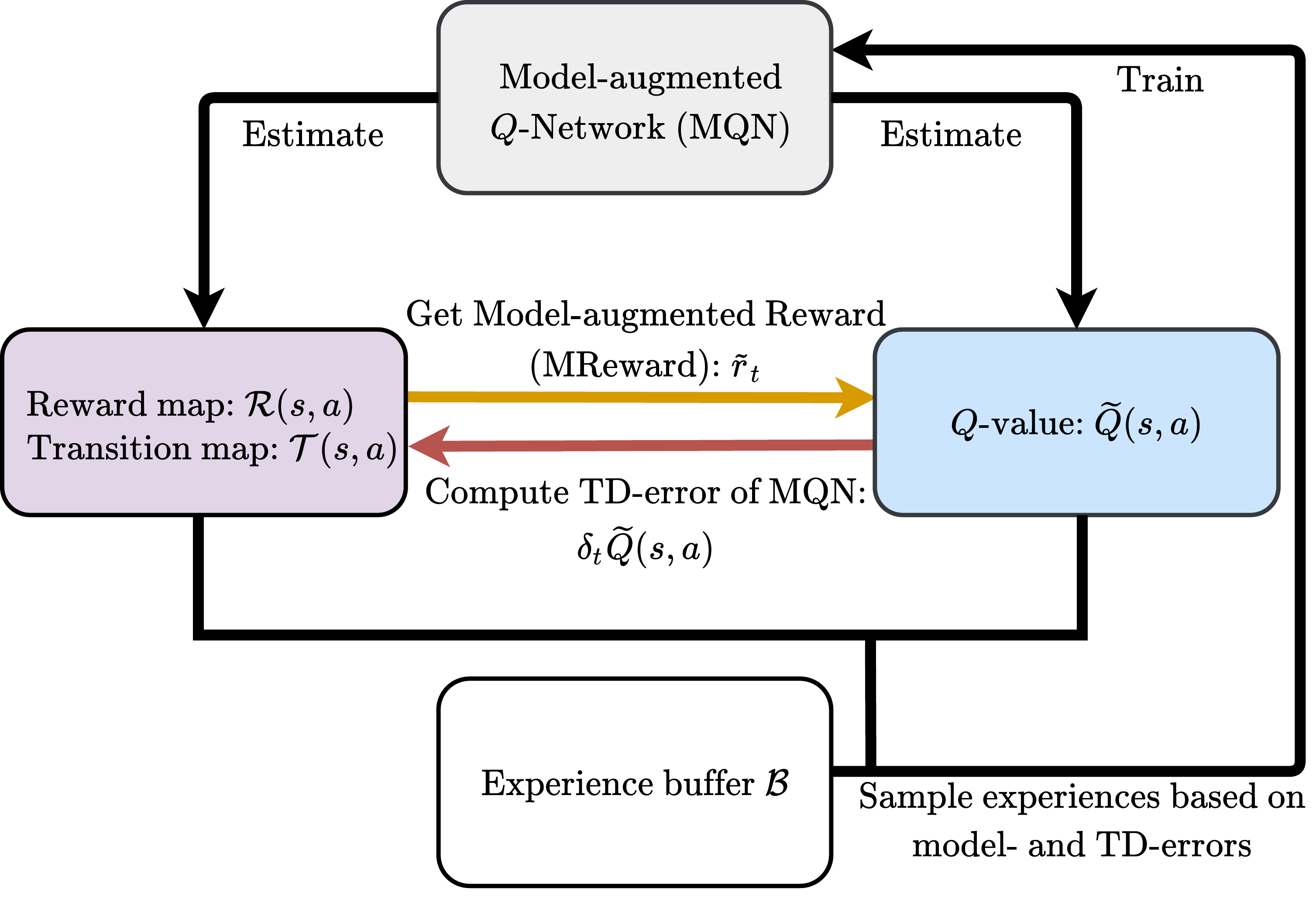}}
\end{tabular}
\caption{\textbf{Overview of Model-Augmented $Q$-Learning.} Our proposed method improves upon the MFRL algorithms. First, we modify a given $Q$-network to a model-augmented $Q$-network (MQN), which also learns the environment model via parameter sharing. The estimators interact with each other via Model-augmented TD learning, where the TD errors of MQN are calculated using the estimated reward (MReward) as well as the transition map. Further, we use the model estimation errors compute priorities of experiences for sampling transitions in a buffer.}
\vspace{-0.3cm}
\label{fig:arch}
\end{figure}

Yet, instead of utilizing the generated transitions as in those methods, we propose to use the components of the model-based RL to \emph{augment} the learning of model-free RL. Specifically, we note that the reward $\reward(s,a)$, and the transition $\transition(s,a)$ are functions that have the same input space as the $Q$-function $Q(s,a)$, and modify the $Q$-network into a \emph{Model-augmented $Q$-network} (MQN) which additionally predicts the reward and the transition function (see Figure~\ref{fig:arch}). However, instead of naively performing multi-task learning via shared network weights, we promote interactions between the $Q$-learner and the model estimators. Specifically, we use the estimated reward from the model (MReward) to calculate the {temporal difference (TD)} target for $Q$-learning, and use the estimated $Q$-value to learn the reward. Moreover, we further utilize the estimation errors for the reward and transition function to prioritize the past experiences with high errors from the replay buffer, which we say  \emph{Model-augmented Prioritized Experience Replay} (MPER). 

Model-augmented $Q$-learning has the following advantages over the existing MFRL algorithms on various environments. First, learning to predict $Q$-values in the long-term is difficult in general, and thus estimating the short-term reward with the shared model helps it learn both short-term and long-term viewpoints. Second, the estimators are highly related in the way that they help with the others' prediction task. Specifically, by utilizing the estimated reward and the transition, we can calculate the $Q$-values for a given policy \emph{without} interacting with the environment. Third, the proposed MPER can sample more diverse experiences to improve the sample-efficiency for not only MQN but also policy networks since we set the sampling priority as a weighted sum of the TD and the model estimation errors.

In summary, our contributions are followings: 
\begin{itemize}[leftmargin=*]
\item We propose \emph{Model-Augmented Q-Learning} (MQL), which not only augments the conventional $Q$-learning with the model estimation tasks via parameter sharing, but also utilizes the estimated values in each task to the computation of the objective in the other task. 

\item We show that MQL satisfies policy invariance and converges to an identical solution obtainable with $Q$-learning with the true reward.

\item We further propose to use the model estimation errors when prioritizing the the past transitions to sample in the experience replay (MPER).

\item We show that MQL is applicable to any modern MFRL algorithms (e.g., SAC, TD3, and Rainbow) with neural network estimators, and significantly improves their performance and sample-efficiency under various environments: MQL
is easy-to-implement and does not incur additional training cost.
\end{itemize}


%% file: 2-method.tex
\section{Method} \label{sec:method}
In this section, we first introduce model-free reinforcement learning (MFRL) with the actor-critic architecture and prioritized experience replay (PER). Then we propose our model-augmented $Q$-network (MQN) and define model-augmented $Q$-learning (MQL) by proposing the model-augmented reward (MReward) and the corresponding TD learning objective with the MReward. By utilizing MQN's TD-error equation (Eq.~\eqref{eq:deltaQ}), we promote interactions between MQN's estimators, such that estimating the models helps better estimate the $Q$-value, and vice versa. We then prove that our method satisfies policy invariance. Finally, we present the model-augmented prioritized experience replay (MPER) which samples the past experiences based on the model estimation errors.

\subsection{Preliminary}

\textbf{Model-Free Reinforcement Learning (MFRL).}  
In a standard reinforcement learning framework, 
an agent interacts with a given environment over discrete timesteps $t$. The agent selects action $a_{t}\in A$ by its policy $\pi$ on the current state $s_{t}\in S$ to receive the next state $s_{t+1}\in S$ and a reward $r_{t}$ from the environment, where $A$ and $B$ are action and state spaces, respectively. The agent's objective is to learn a 
policy $\pi$ that maximizes the return $R_{t}=\sum_{k=0}^{\infty}\gamma^{k}r_{t+k}$, which is the discounted cumulative rewards from the timestep $t$ with
a discount factor of $\gamma\in[0,1)$, at each state $s_{t}$.

Although our method is applicable to $Q$-learning in general, throughout this section, we focus on off-policy actor-critic RL algorithms with an experience replay buffer $\mathcal{B}$, which consist of the policy  (i.e., actor) $\pi_{\Theta}(a|s)$ and $Q$-function networks (i.e., critic) $Q_{\theta}(s,a)$, where $\Theta$ and $\theta$ are their parameters, respectively. The most commonly used loss for $Q_{\theta}(s,a)$ \citep{deep_q-learning_with_er, sac, td3} is 
\begin{align}
    \mathbb{E}_{(s_t, a_t, r_{t}, s_{t+1})\sim\mathcal{B}}\left[\| \delta_{t}Q_{\theta}^{\pi_{\Theta}} \|_{2}^{2}\right], \label{eq:originalQLoss}
\end{align} 
where $\|\cdot\|_{2}^{2}$ is the mean-square error and $\delta_{t}Q_{\theta}$ is the Temporal Difference error (TD-error) defined as follows: for a given transition $(s_t, a_t, r_t, s_{t+1})$, its TD-error $\delta_t Q_{\theta}$ is:
\begin{align}
&\delta_t Q_{\theta}^{\pi_{\Theta}}=\delta Q_{\theta}^{\pi_{\Theta}}(s_t, a_t, r_t, s_{t+1})\nonumber\\&=r_{t}+\gamma Q_{\theta}\left(s_{t+1},a'\right)-Q_{\theta}\left(s_{t},a_{t}\right), \label{eq:TD}
\end{align}where $a'\sim\pi_{\Theta}(\cdot|s_{t+1})$.
We can interpret the TD-error as a measure of how surprising
or `unexpected' the transition is.

\textbf{Prioritized Experience Replay.}  We now explain the concept of PER since we will later augment it with model-based components. Let $[n]$ be defined as  
set $\{1,\cdots,n\}$ for a positive integer $n$. Without
loss of generality, we can suppose that the replay buffer $\mathcal{B}$
stores the following information as its $i$-th transition:
\begin{align}\mathcal{B}_{i}=\left(s_{\kappa(i)},a_{\kappa(i)},r_{\kappa(i)},s_{\kappa(i)+1}\right),\label{eq:transition}\end{align}
with a function from the index of $\mathcal{B}$ to a corresponding timestep: $\kappa\left(i\right)$. PER calculates each $\mathcal{B}_{i}$'s priority $\sigma_{i}$ as the recently computed TD-error about itself and collects a set of priority scores:
\begin{align}
\mathcal{P}_{\mathcal{B}}=\{\sigma_{1},\cdots,\sigma_{\left|\mathcal{B}\right|}\}, \label{eq:priority}
\end{align} where each priority $\sigma_i$ is updated whenever the corresponding transition is sampled for training the actor and critic networks. The TD-error \eqref{eq:TD} is the most frequently used quantity to make the priority set \eqref{eq:priority} \citep{deep_q-learning_with_per,PSER, rainbow, efficientrainbow}.
Then the sampling strategy of PER is to determine an index set $I$ in $\left[\left|\mathcal{B}\right|\right]$ from the probability $p_{i}$ of $i$-th transition defined by the priority set:
\begin{equation}
p_{i}=\frac{\sigma^{\alpha}_{i}}{\sum_{k\in\left[\left|\mathcal{B}\right|\right]}\sigma^{\alpha}_{k}}, \label{eq:samplingprob}
\end{equation} with a hyper-parameter $\alpha>0$. Next, to compensate the bias of probabilities, we calculate the importance-sampling weights as follows:
\begin{equation}
w_{i} = \left(\frac{1}{| \mathcal{B} | p(i)}\right)^{\beta}, \label{eq:computeweight}
\end{equation} where $\beta>0$ is also a hyper-parameter. 

\begin{algorithm}[t!]
  \caption{MQN, MPER, MReward to actor-critic RL methods}
  \label{alg:Alg1}
\begin{algorithmic}
    \STATE Initialize the network $\pred$ parameters  $\theta$, the actor parameters $\Theta$, a replay buffer $\mathcal{B}\leftarrow \emptyset$, priority set $\mathcal{P}_{\mathcal{B}}\leftarrow \emptyset$, and  index set $\mathcal{I}\leftarrow \emptyset$. 
    \STATE Set the batch size $m$. 
    \FOR {each timestep $t$}
    \STATE Choose $a_{t}$ from the actor and collect a transition $(s_t, a_t, r_{t},s_{t+1})$ from the environment.
    \STATE Update replay buffer $\mathcal{B} \leftarrow \mathcal{B}\cup \{(s_t, a_t, r_{t},s_{t+1})\}$ and priority set $\mathcal{P}_{\mathcal{B}}\leftarrow \mathcal{P}_{\mathcal{B}}\cup\{1.0\}$. 
    \FOR {each gradient step}
    \STATE Sample an index $I$ by the given set $\mathcal{P}_{\mathcal{B}}$ and Eq.\ (\ref{eq:samplingprob})  with $|I|=m$.
    \STATE Calculate weights $\{w_{i}\}_{i\in I}$ and a priority set $\{ \sigma_{k}\}_{k\in I}$ 
    by  (\ref{eq:computeweight}) and \eqref{eq:setpriority}, respectively.
    \STATE Train $\pred_{\theta}=(\widetilde{Q}_\theta, \reward_\theta, \transition_\theta)$ by \eqref{eq:TotalLoss} and the actor by $\widetilde{Q}_{\theta}$ with batch $\{\mathcal{B}_{i}\}_{i\in I}\subset \mathcal{B}$ and corresponding weights $\{w_{i}\}_{i\in I}$.
    \ENDFOR
    \ENDFOR
\end{algorithmic}
\end{algorithm}

\subsection{Model-augmented $Q$-learning} \label{sec:mql}
\textbf{Model-augmented $Q$-network.}
In an RL framework, an environment model consists of two maps: a reward map $\reward(s,a)$ and a transition model $\transition(s,a).$ We observe that these two maps and $Q_{\theta}(s,a)$ have the same input domain, i.e., $S\times A$, where $S$ and $A$ are state and action spaces. Motivated by this, we slightly modify $Q_{\theta}(s,a)$ as $\pred_{\theta} (s,a)$ to additionally predict the environment model: a reward network $\reward_{\theta}(s,a)$ and a transition model $\transition_{\theta}(s,a)$ in parallel via parameter sharing. We refer to $\pred_{\theta} (s,a)$ as a \emph{Model-augmented $Q$-network} (MQN) such that $$
\pred_{\theta}=\left(\widetilde{Q}_\theta, \reward_{\theta}, \transition_{\theta} \right),
$$ where $\widetilde{Q}_\theta$ measures $Q$-values with MReward with the additional estimators, which we will explain soon below. 

\textbf{Model-Augmented $Q$-Learning (MQL).}
Instead of simply resorting to weight sharing, we formulate a new objective for MQN to promote the interaction between the estimators. To this end, we first propose the following losses the estimation of $\reward_{\theta}$ and $\transition_{\theta}$:
\begin{align}
    L_{\reward_{\theta}}&=\mathbb{E}_{(s_t, a_t, r_{t}, s_{t+1})\sim\mathcal{B}}\left[\| \delta_{t}\reward_{\theta}\|_{2}^{2}\right], \label{eq:rewardloss} \\
    L_{\transition{\theta}}&=\mathbb{E}_{(s_t, a_t, r_{t}, s_{t+1})\sim\mathcal{B}}\left[\| \delta_{t}\transition_{\theta}\|_{2}^{2}\right] \label{eq:transitionloss},
\end{align} where
\begin{align}
    \delta_{t}\reward_{\theta}&=\reward_{\theta}(s_t,a_t) - r_{t}, \label{eq:diffreward} \\
    \delta_{t}\transition_{\theta}&=\transition_{\theta}(s_t,a_t) - s_{t+1}\label{eq:difftransition},
\end{align}
Then by using the estimation errors in \eqref{eq:diffreward}-\eqref{eq:difftransition}, we shape the new reward $\widetilde{r}_{t}$ (MReward) instead of using $r_{t}$:
\begin{align}
    \widetilde{r}_{t}& = \reward_{\theta} + \rewardcoefone \delta_{t}\reward_{\theta} + \rewardcoeftwo \delta_{t}\transition_{\theta}.& \label{eq:modifiedreward}
\end{align} with some positive coefficients $\rewardcoefone$ and $\rewardcoeftwo.$ We use MReward to develop a new TD-error about MQN, which encourage agent's exploration compared to the original $Q$-network. However, using MReward may change the direction of agent's learning. However, we will prove that learning of MQN with MReward does not change the set of optimal policies in the last paragraph of this section.

We formulate the loss for $\widetilde{Q}_{\theta}$ using the MReward \eqref{eq:modifiedreward}, that is similar to the TD-error in \eqref{eq:TD}:
\begin{align}
    L_{\widetilde{Q}_{\theta}}=\mathbb{E}_{(s_t,a_t,s_{t+1})\sim\mathcal{B}}\left[\| \delta_{t} \widetilde{Q}_\theta^{\pi_{\Theta}}
     \|_{L^2}^2\right], \label{eq:QLoss}
\end{align}where
\begin{align}
&{\delta}_{t}\widetilde{Q}_{\theta}^{\pi_{\Theta}}={\delta}\widetilde{Q}_{\theta}^{\pi_{\Theta}}(s_{t},a_{t},r_{t},s_{t+1})\nonumber\\&=\widetilde{Q}_{\theta}(s_t,a_t) - \left( \widetilde{r}_{t} + \gamma \widetilde{Q}_{\theta}(s_{t+1},a')\right). \label{eq:deltaQ}
\end{align}
Here $a'$ denotes the action sampled by the current policy on the current state, i.e., $a'\sim \pi_{\Theta}(\cdot|s_{t+1})$.
Then combining \eqref{eq:rewardloss}-\eqref{eq:transitionloss} and \eqref{eq:QLoss}, we obtain the loss for Model-augmented $Q$-learning (MQL) of $\pred_{\theta}$:
\begin{align}
    L_{\pred_{\theta}}=\losscoefzero L_{\widetilde{Q}_{\theta}}+\losscoefone L_{\reward_{\theta}}+\losscoeftwo L_{\transition_{\theta}}, \label{eq:TotalLoss}
\end{align}  with some positive coefficients $\losscoefzero$, $\losscoefone$, and $\losscoeftwo$. The coefficients adaptively changes by the quantity of each loss by employing a dynamic method in  \citep{liang2020simple}. We will specify how to determine the coefficients in the supplementary material.

\textbf{Model-augmented PER.}\label{subsec:MPER}
Using \eqref{eq:deltaQ}, we make the estimators in MQN interact positively with each other by proposing a suitable sampling method.
we formulate MPER from PER that compute the sampling probability of each transition based on a single metric, e.g., TD-error. Since our method uses $\pred_{\theta}(s,a)$ instead of the original action-value function, we modify the rule in obtaining the priority set in~\eqref{eq:priority} for experiences in the buffer $\mathcal{B}$ accordingly. To this end, we compute the priority of each transition as the weighted sum of the TD-errors and model errors using \eqref{eq:TD} and \eqref{eq:diffreward}-\eqref{eq:difftransition}:
\begin{equation}
\sigma_{i} =  \losscoefzero|\delta_{\kappa(i)}\widetilde{Q}_{\theta}|^2+ \losscoefone|\delta_{\kappa(i)}\reward_{\theta}|^2 + \losscoeftwo|\delta_{\kappa(i)}\transition_{\theta}|^2. \label{eq:setpriority}
\end{equation}
Then MPER computes probabilities and weights for corresponding transitions by \eqref{eq:samplingprob} and \eqref{eq:computeweight}. When using the priority scoring in~\eqref{eq:setpriority}, the buffer can sample various experiences that are useful for both long-term and short-term viewpoints.

\textbf{Interaction of MQN's estimators.}\label{subsec:int} Now then we can state that the equation \eqref{eq:deltaQ} makes MQN's estimators positively interact each other with MPER. For instance, if the model errors are large, then 
$\delta_t Q_{\theta}^{\pi_{\Theta}}$ in \eqref{eq:deltaQ} should be also large since MReward $\widetilde{r}_t$ deviates from the real reward. Since both errors are large, MPER samples experiences that are valuable to learn both the estimators adaptively by the priority set. In other words, due to the interaction between the two estimators in MQN, \textbf{reducing model errors also results in reducing TD errors simultaneously}. The interaction is not possible without both parameter sharing and the equation  \eqref{eq:deltaQ}, since it is nontrivial to select experiences that are valuable to all estimators. We validate our statement in Section~\ref{sec:exp}.

The detailed descriptions of our model-augmented $Q$-learning is provided in Algorithm \ref{alg:Alg1}.

\textbf{Convergence of MQL} \label{subsec:con}
We establish the following guarantee on convergence of Model-augmented $Q$-learning to a unique solution that is identical to the solution obtained with the true reward, under a fixed policy:
\begin{theorem}\label{mainthm}
Let $\pred=\left(\widetilde{Q}, \reward, \transition \right)$ be estimators of $Q$-value, reward, and transition maps defined as in Section~\ref{sec:mql}. If $\pred$ is updated by the following loss \eqref{eq:TotalLoss} with the MReward \eqref{eq:modifiedreward} for a given policy $\pi^*$, then $\pred$ converges to a unique $\left(\widetilde{Q}^*, \reward^*, \transition^* \right)$ satisfying 
\begin{align}
\widetilde{Q}^*(s_t,a_t)=  r_{t} + \gamma \widetilde{Q}^*(s_{t+1},a) \label{eq:thm}
\end{align} for any transition $(s_t, a_t, r_t, s_{t+1})$, where $a\sim\pi^*(\cdot|s_{t+1})$.
\end{theorem}

Theorem~\ref{mainthm} suggests that the convergence of $\widetilde{Q}$ is invariant with respect to MReward \eqref{eq:modifiedreward}. In other words, our method guarantees policy invariance, which means that the modification of the $Q$-network does not affect a set of Pareto optimal policies. We provide the proof of this theorem in the supplementary material.  

%% file: 3-exp-arxiv.tex
\section{Experiment}
\label{sec:experiment}

In this section, we conduct experiments to answer the following questions:

\begin{itemize}[leftmargin=*]
\item Can the proposed method enhance the performances of diverse off-policy MFRL algorithms in various environments?
\item What attributes to the success of our method the most?
\end{itemize}

We first describe our experimental setup and show the main results against relevant baselines to show that our method is generally applicable to any off-policy MFRL methods with $Q$-learning and experience replays, and largely improves their performance. Then we perform an ablation study of MQN to analyze the most crucial components of it.

\subsection{Experimental setup}\label{sec:exp}

\textbf{Off-policy MFRL algorithms.} We validate the effectiveness of our model-augmented $Q$-learning with the following algorithms: Soft Actor-Critic (SAC)~\citep{sac}, Twin Delayed Deep Deterministic (TD3), and Rainbow \citep{rainbow}. Here, we apply data-efficient Rainbow \citep{efficientrainbow} since its sample efficiency is dramatically higher than the original and SimPLe. We modify these base algorithms to obtain different variants of our method, namely Model-augmented SAC (MSAC), Model-augmented TD3 (MTD3), and Model-augmented Rainbow (MRainbow).
We consider SAC, TD3, and Rainbow because they and their variants are state-of-the-art off-policy RL algorithms for continuous and discrete control tasks, respectively. We emphasize that when applying our method to given algorithms, we do not alter the original hyper-parameters to show that our method effectively improves the base algorithm's performance without any hyperparameter tuning. The detailed configuration of the hyper-parameters are provided in the \textbf{supplementary material}.

\textbf{Environments.} 
Although most RL algorithms frequently have used MuJoCo environments \citep{todorov2012mujoco}, they are not freely available to everyone since they
belong to a commercial physics engine. Accordingly, we consider alternative free implementations of the original MuJoCo environments, called PyBullet Gymperium\footnote{https://github.com/benelot/pybullet-gym}, and other free environments in the OpenAI Gym~\citep{brockman2016openai}.
We validate our method 
on the following standard continuous control tasks: HumanoidPyBulletEnv-0 (Humanoid), HalfCheetahPyBulletEnv-v0 (HalfCheetah), HopperPyBulletEnv-v0 (Hopper), BipedalWalkerHardcore-v3,
and Pendulum*. All of environments are supported by PyBullet Gymperium except for the last two tasks. 
BipedalWalkerHardcore-v3 belongs to Box2D continuous control tasks in the OpenAI gym and 
Pendulum$^*$ is a sparse reward environment which is a modification from Pendulum-v0 from the OpenAI Gym. 
In Pendulum$^*$, agents receive the reward of $+1$ only if the rod is at the upright position $(-3/\pi, 3/\pi)$ in 100 continuous steps. In the case of discrete control tasks, we validate our method on Atari games. We provide the details of the environments we used for the experiments in the \textbf{supplementary material}.

\begin{figure*}[!t]
\centering
\begin{tabular}{cccc}
    \makecell{\includegraphics[width=0.22\columnwidth]{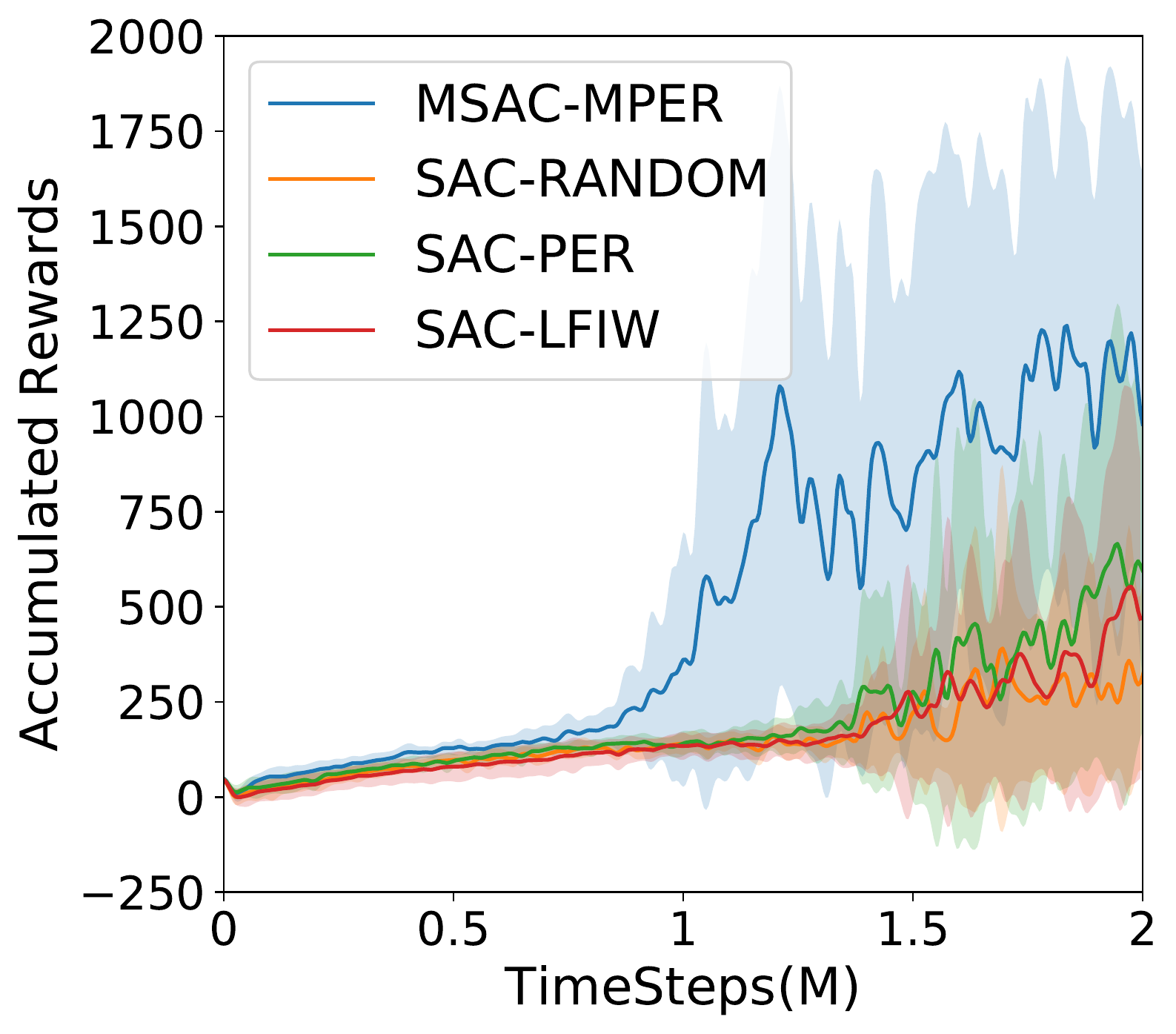}} &
    \makecell{\includegraphics[width=0.22\columnwidth]{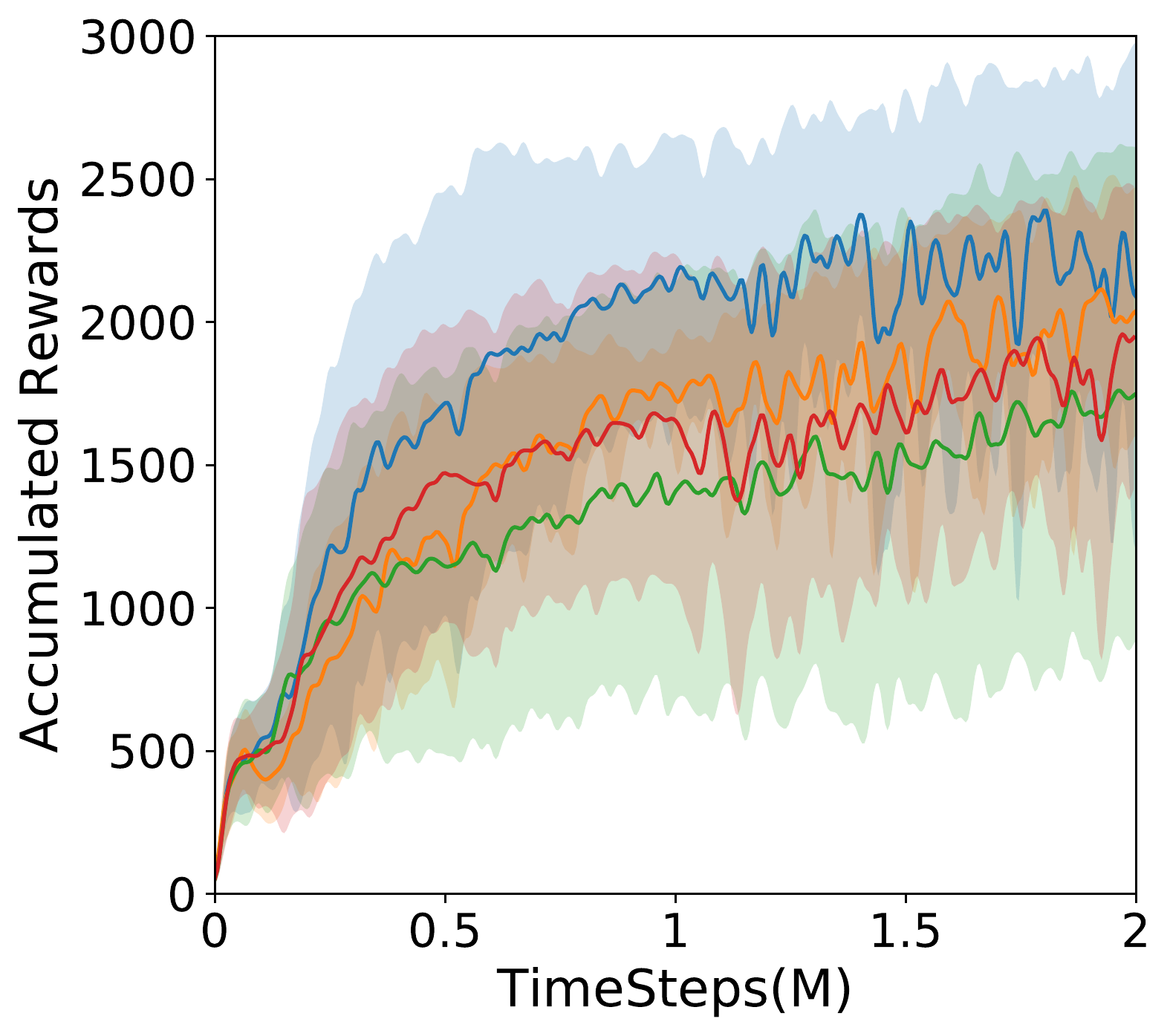}} &
    \makecell{\includegraphics[width=0.22\columnwidth]{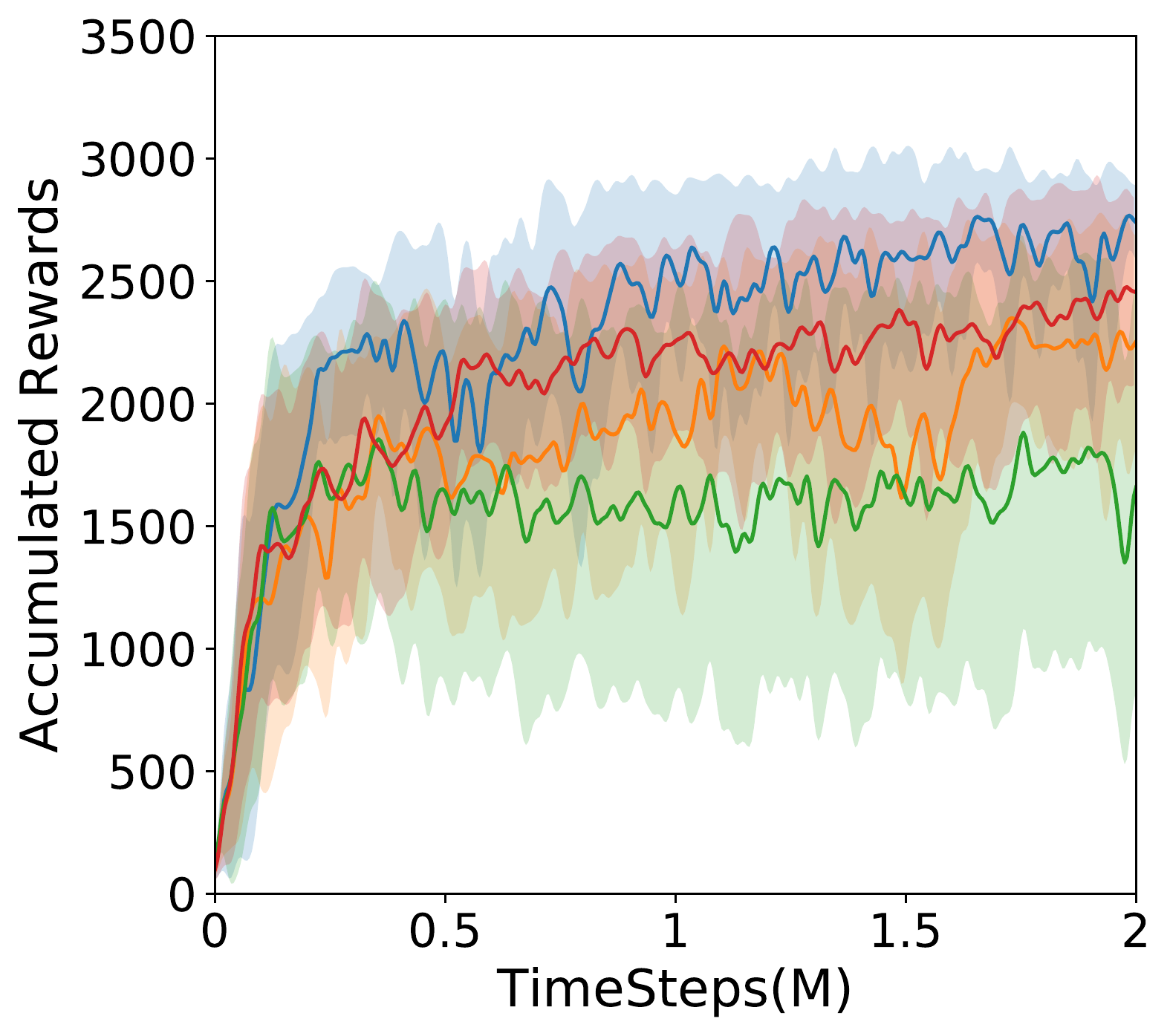}} & \makecell{\includegraphics[width=0.22\columnwidth]{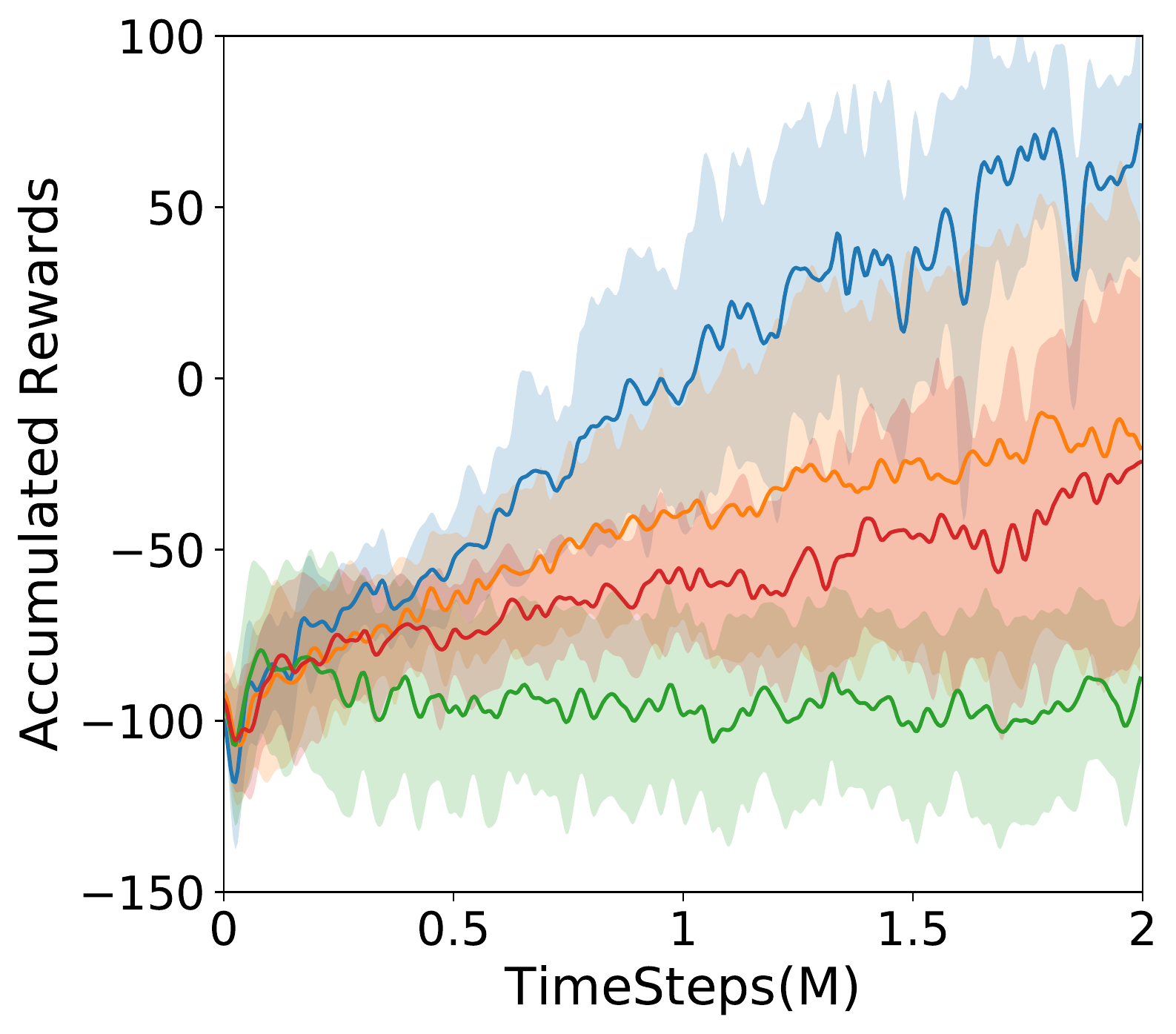}} 
    \\
    (a) Humanoid (SAC) & 
    (b) HalfCheetah (SAC) & 
    (c) Hopper (SAC) & 
    (d) BWH (SAC)
    \\
    \makecell{\includegraphics[width=0.22\columnwidth]{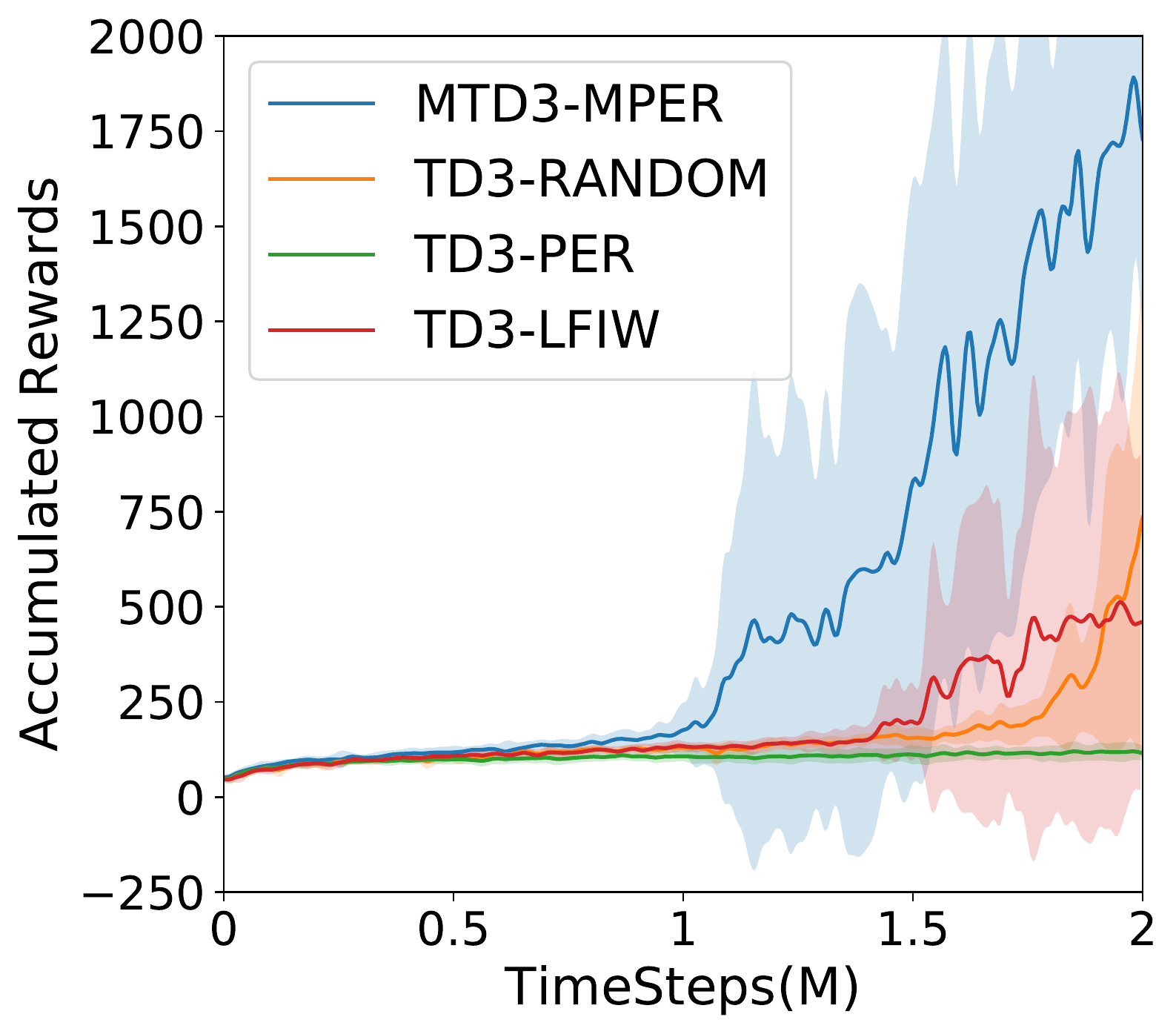}} &
    \makecell{\includegraphics[width=0.22\columnwidth]{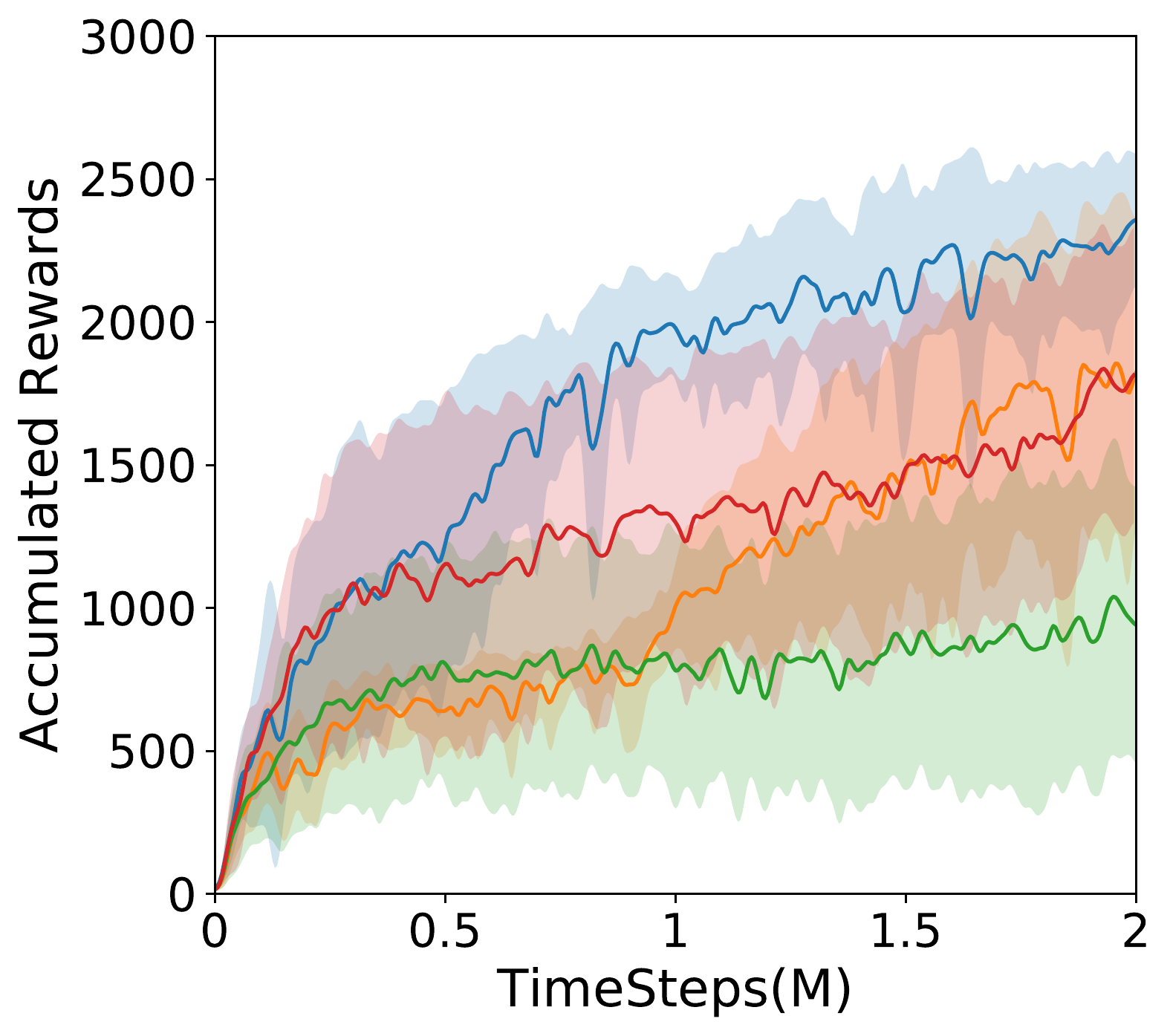}} &
    \makecell{\includegraphics[width=0.22\columnwidth]{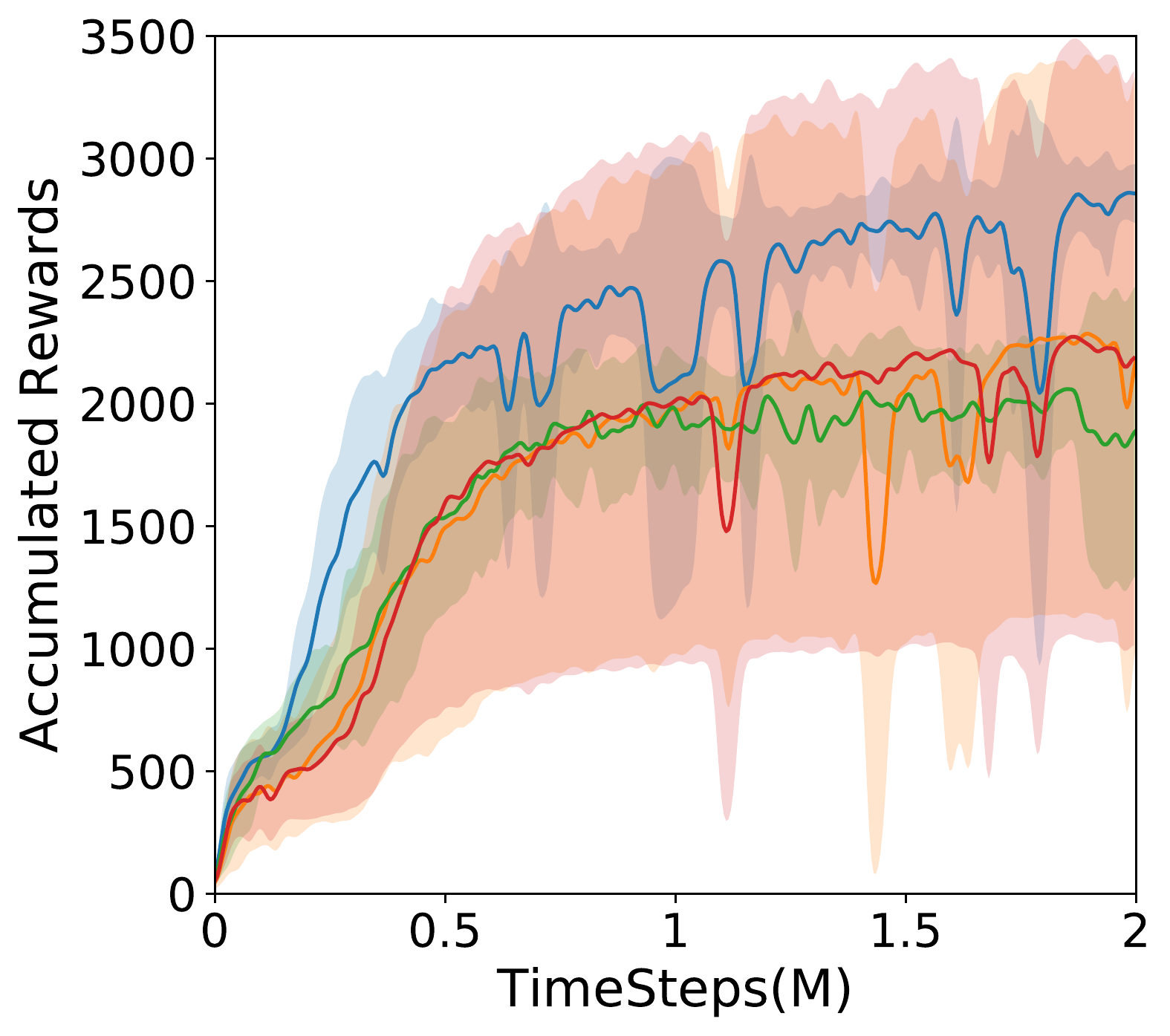}} & \makecell{\includegraphics[width=0.22\columnwidth]{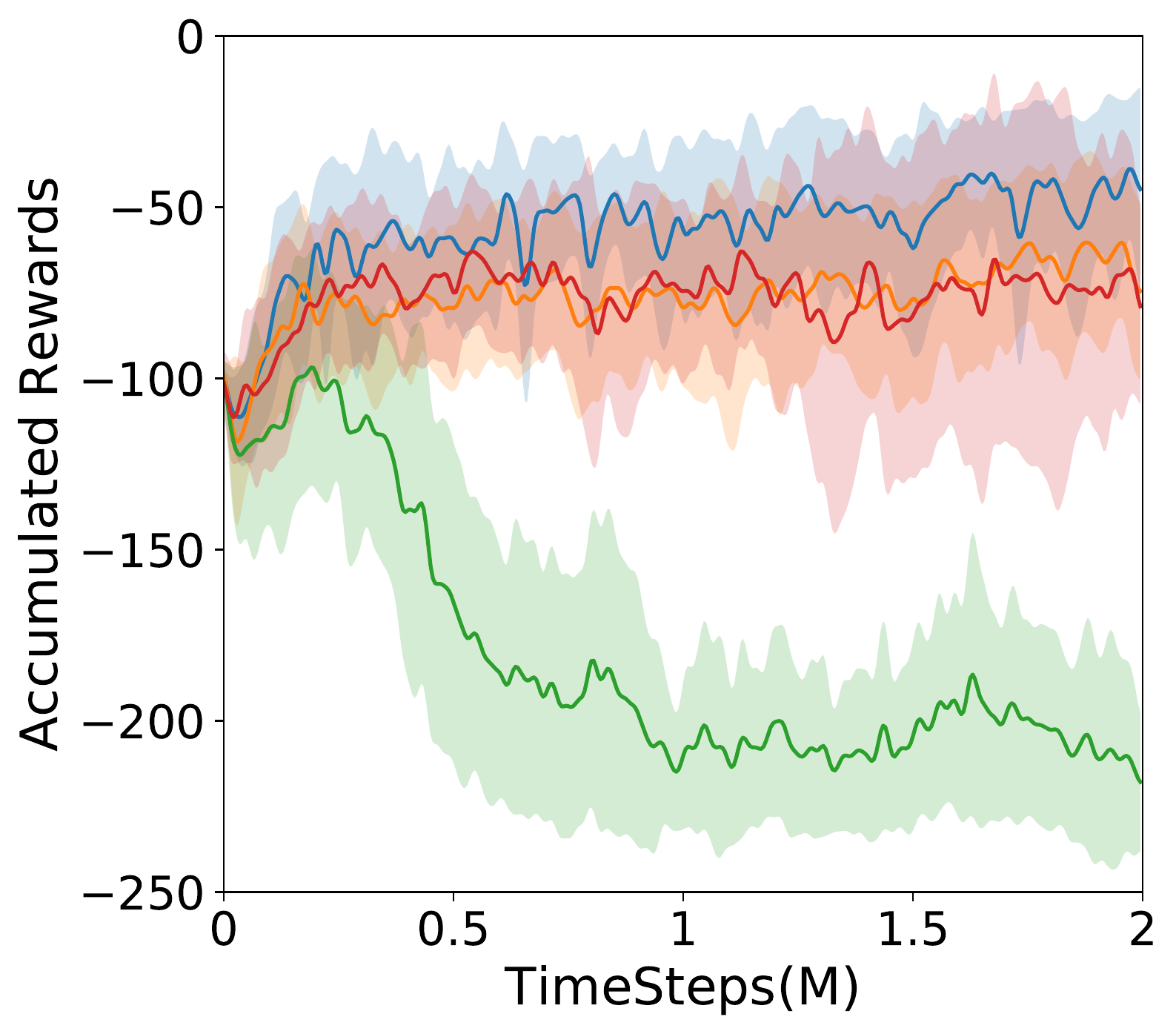}} 
    \\
    (e) Humanoid (TD3)& (f) HalfCheetah (TD3)& (g) Hopper (TD3)& (h) BWH (TD3)  
\end{tabular}
\caption{Learning curves of off-policy RL algorithms SAC and TD3 on PyBullet environments (Humanoid, HalfCheetah, and Hopper) and BWH.
Here, MSAC-MPER and MTD3-MPER, which represent variants of SAC and TD3 by our method, respectively, outperform other RL algorithms. CDE and PB denote reward shaping methods: curiosity-driven exploration \citep{curiositydriven} and potential-based reward shaping \citep{zou2019reward}, respectively.
The solid line and shaded regions represent the mean and standard deviation, respectively, across five runs with random seeds. 
}
\label{fig:exp1}
\end{figure*}

\textbf{Reward shaping methods.} 
We compare our MReward against the following reward shaping methods.
\begin{itemize}[leftmargin=*]
\item Curiosity-driven exploration (CDE) \citep{curiositydriven}: This method uses additional intrinsic rewards to promote exploration:
$$
r_{t}^{i} = \frac{\eta}{2}\| \widehat{\phi}\left(s_{t+1}\right)-\phi\left(s_{t+1}\right) \|,
$$
where $
\phi(s_t)$ is a feature vector and $
\widehat{
\phi}(s_t)$ is a predicted feature vector for $s_t$, respectively.
\item Potential-based method (PB) \citep{ng1999policy,wiewiora2003principled,devlin2012dynamic}: This method uses a modified reward $R_t=r_t+F$, where 
$$
F_t(s,a,s',a') = 
-\gamma \Psi(s',a') + 
\Psi(s,a)
$$ and $\Psi$ is a real-valued function. Although there exist various methods for choosing $\Psi$, we use the value network as $\Psi$ following~\citep{zou2019reward}, as we do not assume any prior knowledge about the given task.
\end{itemize}

\textbf{Sampling methods.} 
We compare the performance of our MPER against the following experience replays.
\begin{itemize}[leftmargin=*]
\item Experience Replay with Uniform Sampling at Random (RANDOM) :  Sampling transitions uniformly at random.
\item Priotized Experience Replay (PER) \citep{deep_q-learning_with_per}: Rule-based prioritized sampling of the transitions
based on TD-errors.
\item Experience Replay with Likelihood-free Importance Weights (LFIW) 
\citep{sinha2020experience}: Learning-based sampling method which predicts the importance of each experience. Since the authors only validated it on continuous control tasks with SAC and TD3, we compare against it in the same settings.
\end{itemize}

\begin{figure*}[ht!]
\centering
\begin{tabular}{cccc}
    \makecell{\includegraphics[width=0.22\columnwidth]{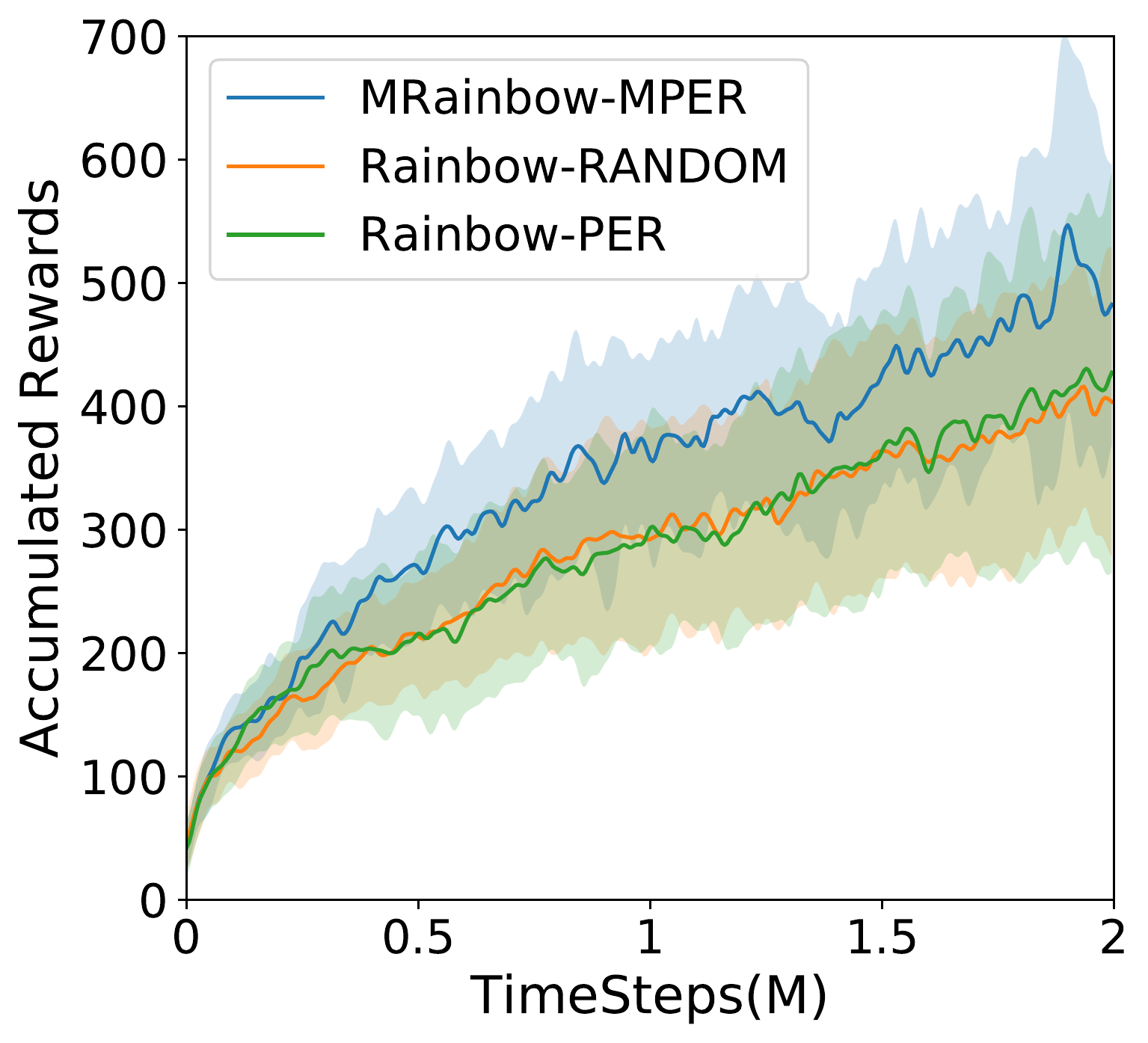}} &
    \makecell{\includegraphics[width=0.22\columnwidth]{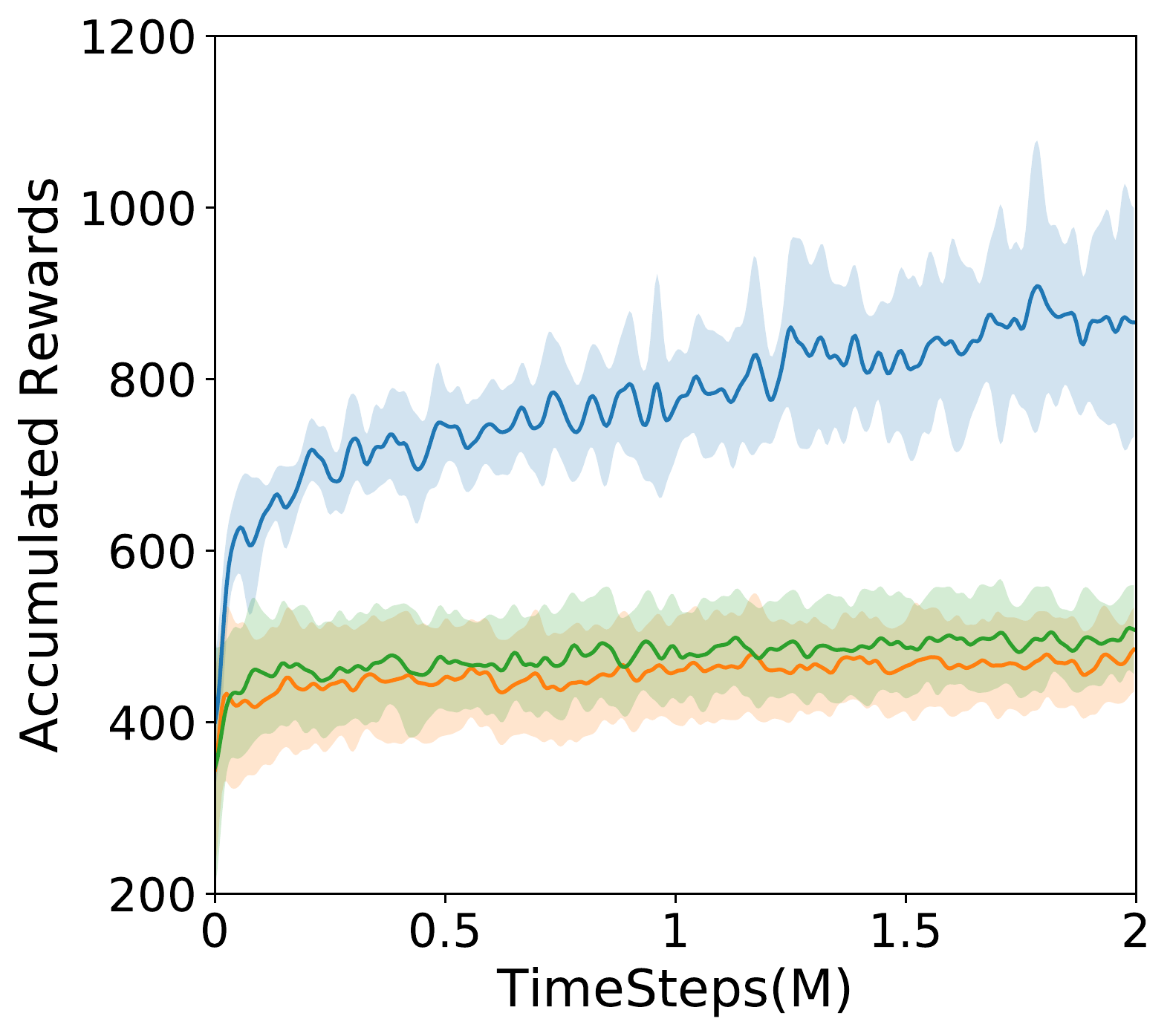}} &
    \makecell{\includegraphics[width=0.22\columnwidth]{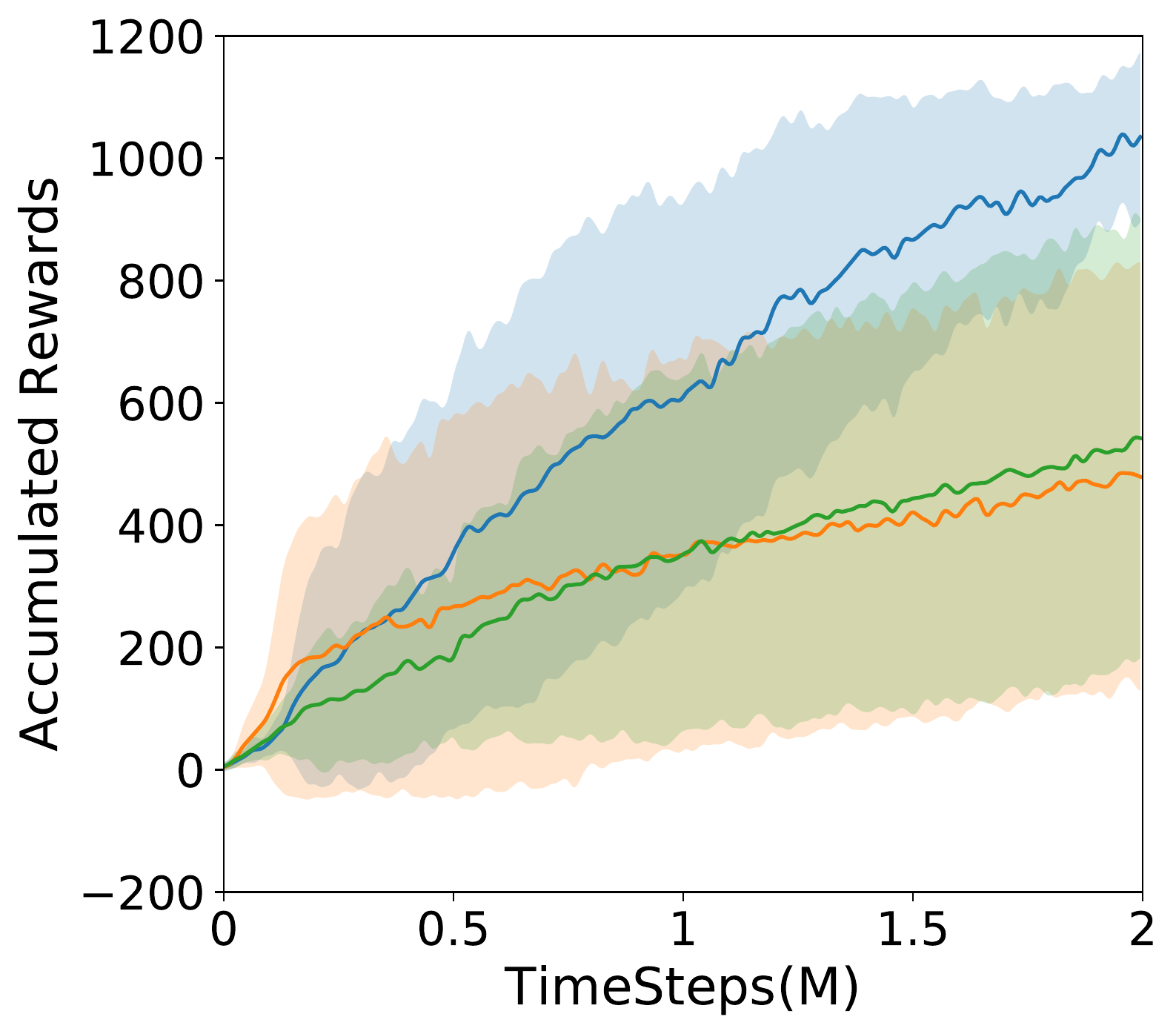}} &
    \makecell{\includegraphics[width=0.22\columnwidth]{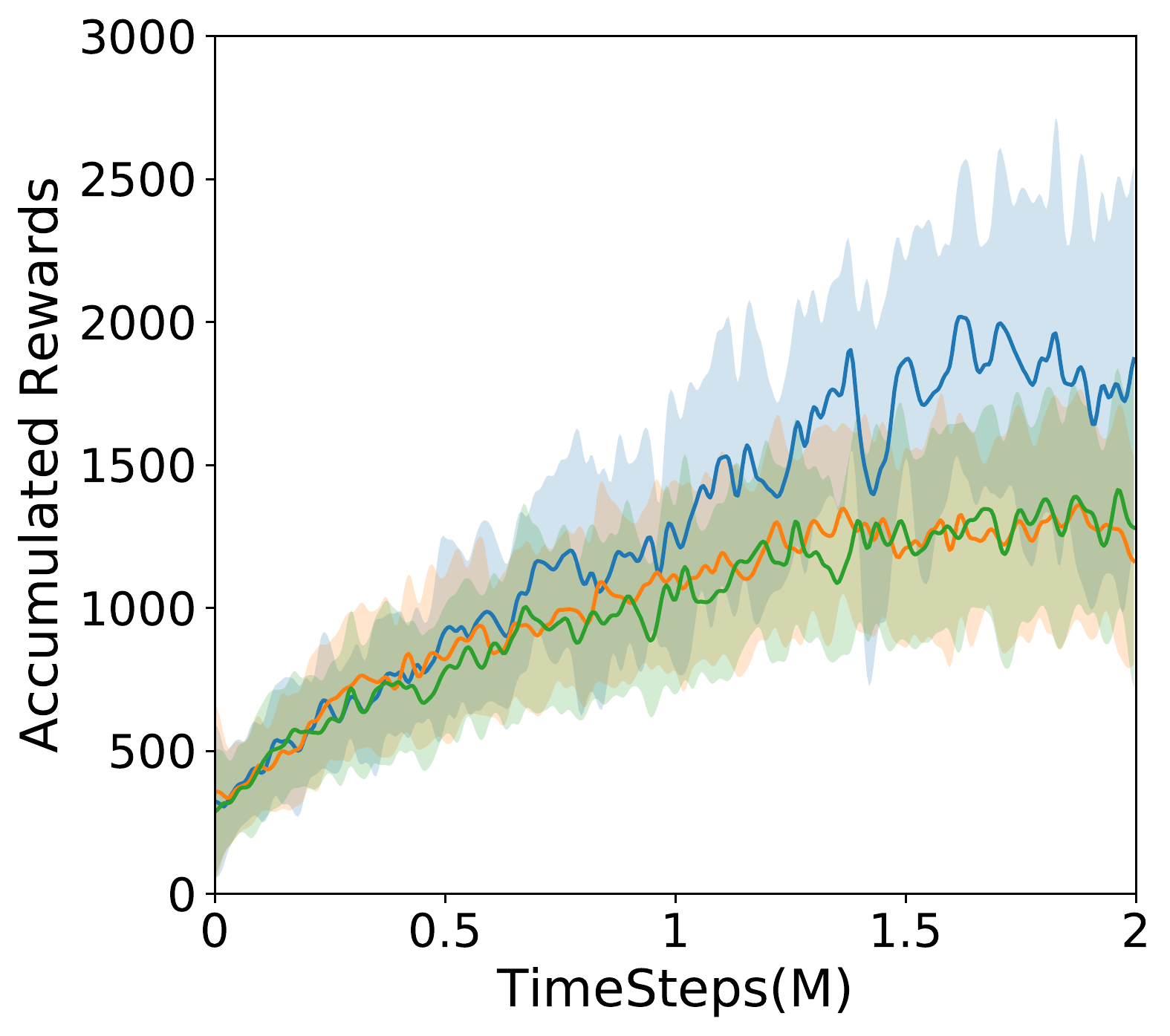}} 
    \\
    (a) Amidar & (b) Assault & (c) BankHeist & (d) DemonAttack\\
    \makecell{\includegraphics[width=0.22\columnwidth]{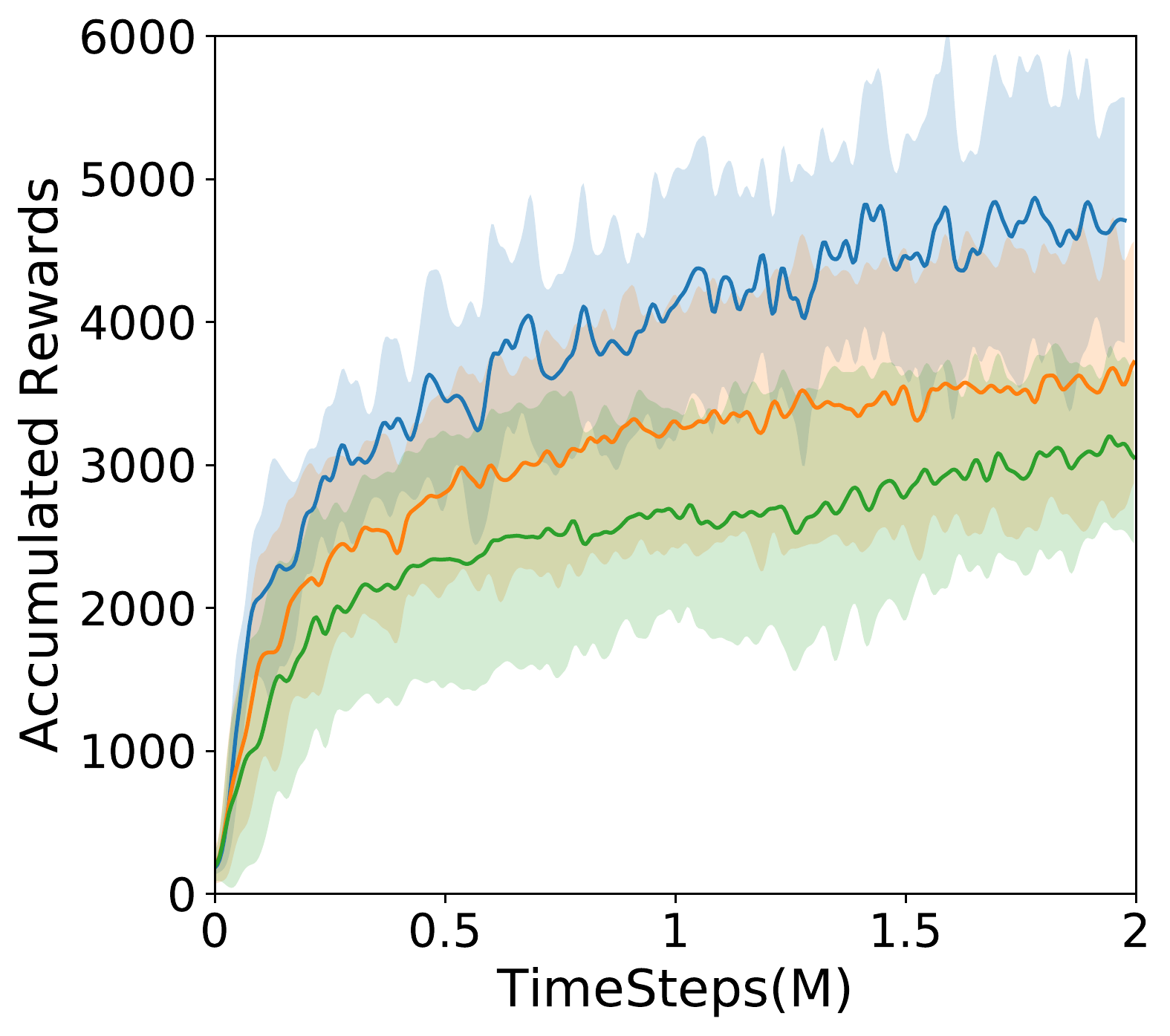}} &
    \makecell{\includegraphics[width=0.22\columnwidth]{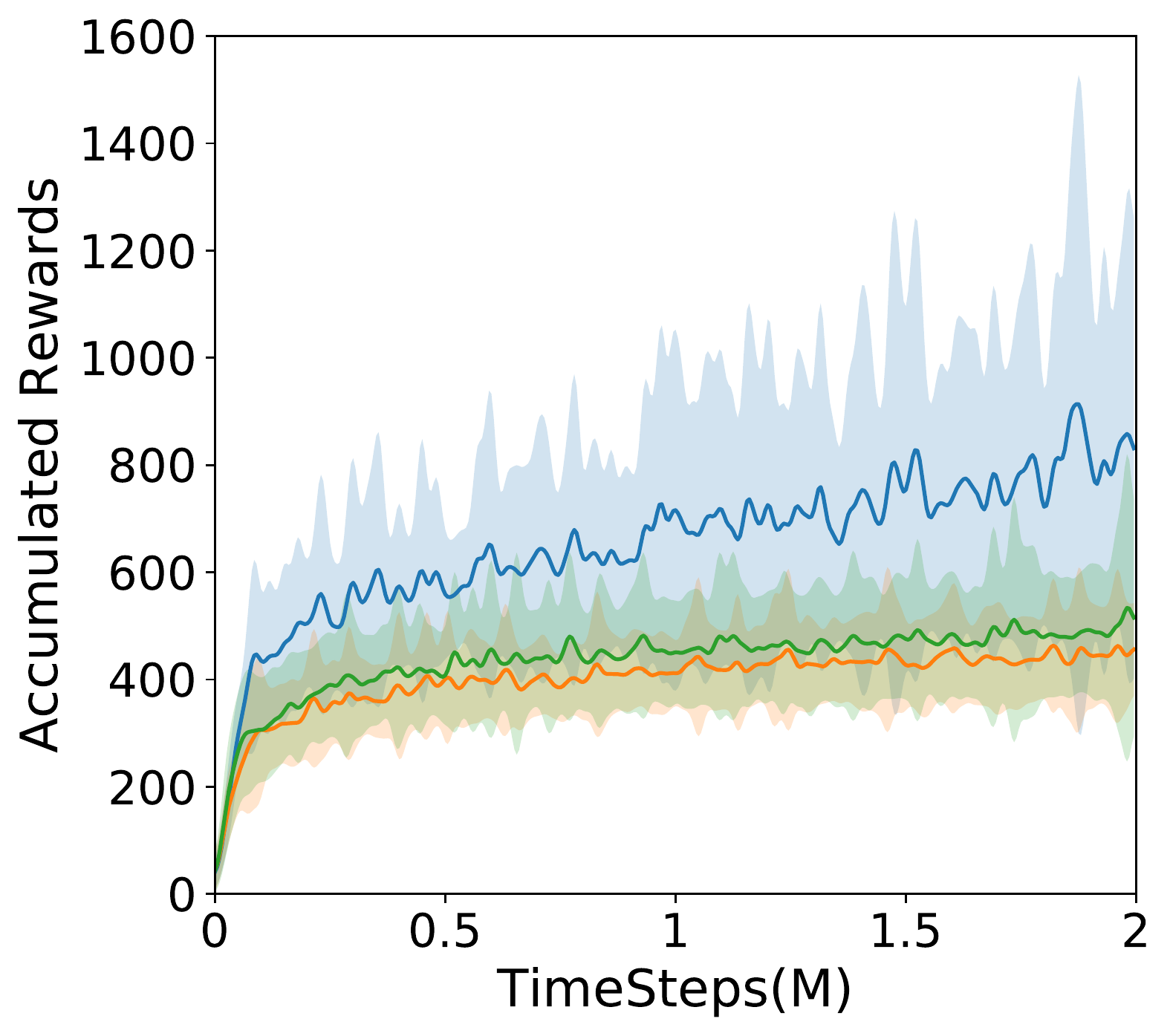}} &
    \makecell{\includegraphics[width=0.22\columnwidth]{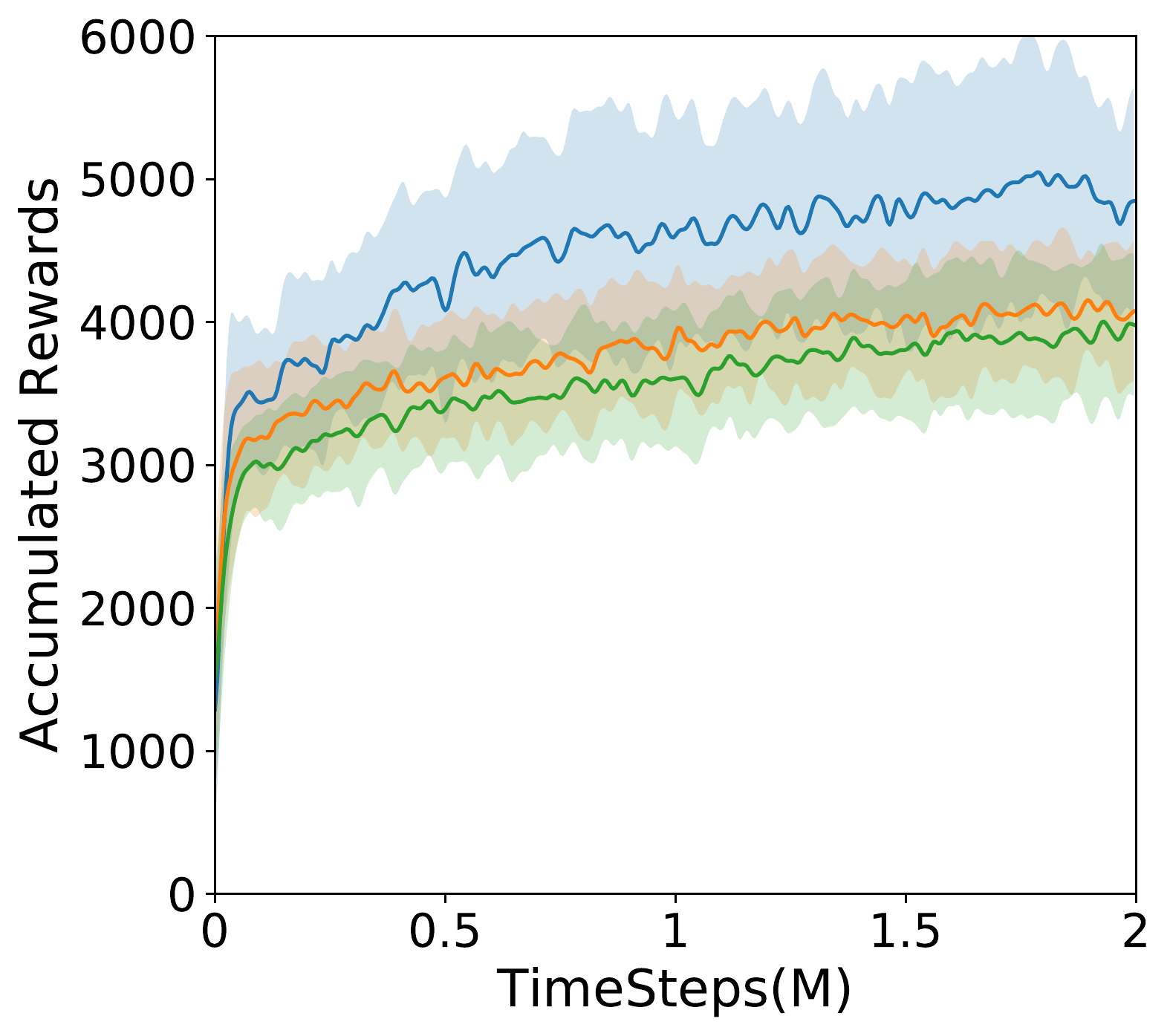}}& 
    \makecell{\includegraphics[width=0.22\columnwidth]{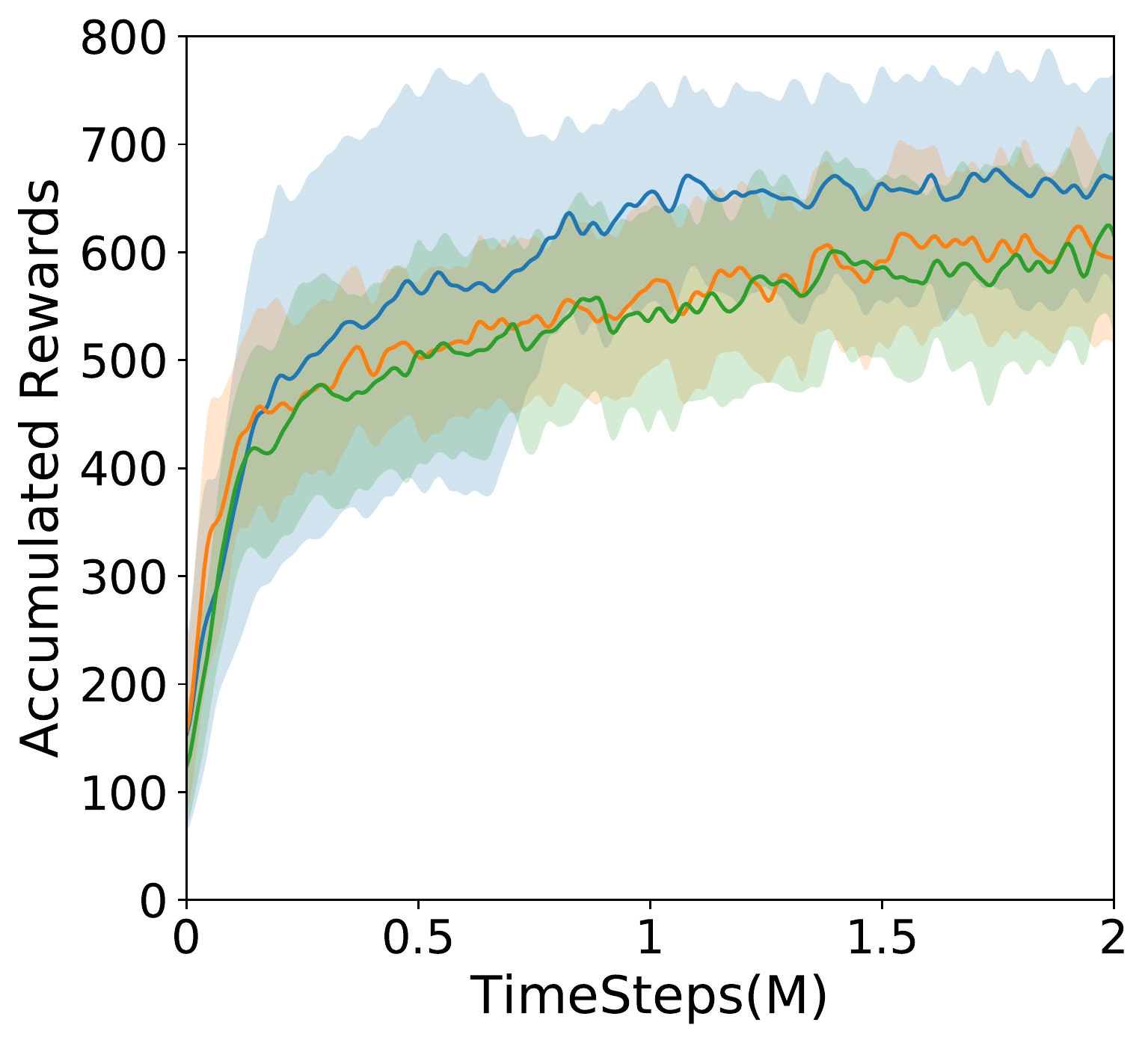}}\\
    (e) Frostbite & (f) Jamesbond & (g) Krull & (h) Seaquest
\end{tabular}
\caption{Learning curves of Rainbow on Atari games. Here, MRainbow-MPER is a variant of Rainbow, where all of MQN, MPER, and MReward are applied. Although Rainbow basically adopts PER, we denote it as Rainbow-PER for consistency. One can observe that MRainbow-MPER overwhelms other methods.
The solid line
and shaded regions represent the mean and standard deviation, respectively, across five runs  with random seeds. 
}
\label{fig:exp1-atari}
\end{figure*}

\begin{figure}[!t]
\centering
\begin{tabular}{cc}
    \makecell{\includegraphics[width=0.48\columnwidth]{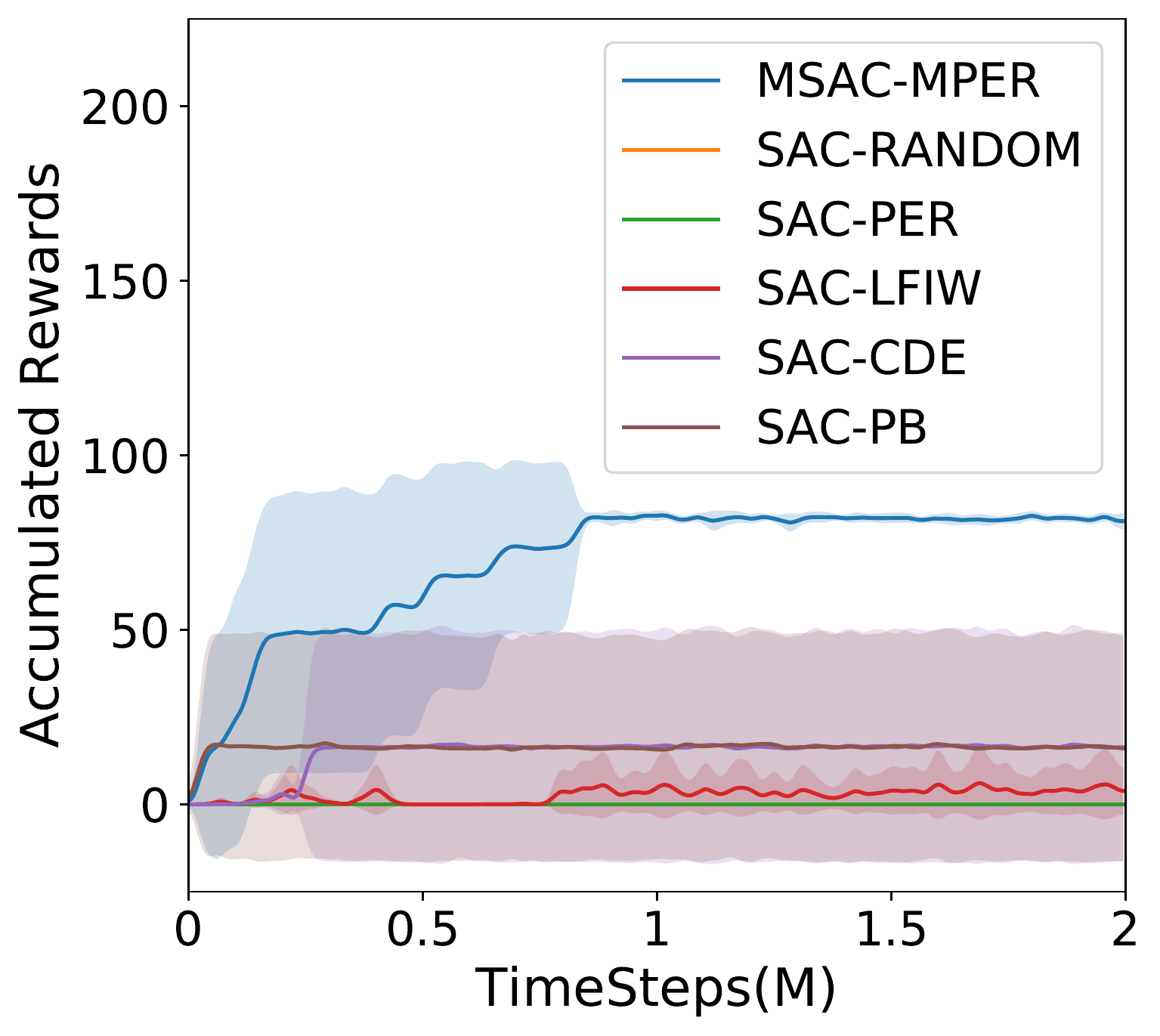}} & 
    \makecell{\includegraphics[width=0.48\columnwidth]{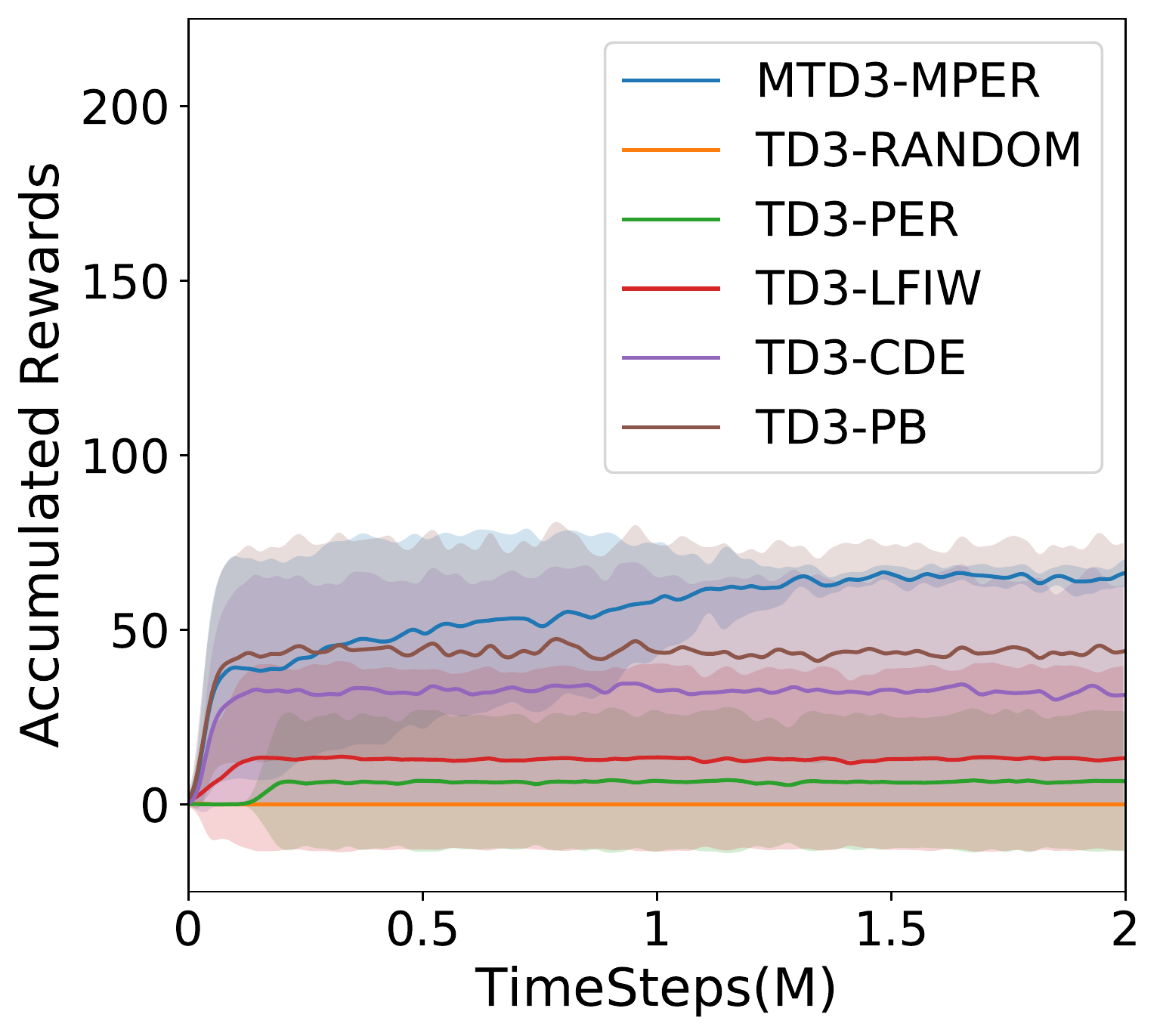}}
    \\
    (a) Pendulum$^{*}$ (SAC) &
    (b) Pendulum$^{*}$ (TD3)
\end{tabular}
\caption{Learning curves of off-policy RL algorithms SAC and TD3  on Pendulum$^{*}$.
Here, MSAC-MPER and MTD3-MPER, which represent variants of SAC and TD3 by our method, respectively, outperform other RL algorithms. CDE and PB denote reward shaping methods: curiosity-driven exploration \citep{curiositydriven} and potential-based reward shaping \citep{zou2019reward}, respectively.
The solid line and shaded regions represent the mean and standard deviation, respectively, across five runs with random seeds. 
}
\label{fig:exp2}
\end{figure}

\subsection{Main Experimental Results}
We consider three types of environments for our main experiments: Pybullet, sparse Pendulumn, and Atari games.

\textbf{Non-sparse reward environment}. Figure \ref{fig:exp1}.(a)-(d) and Figure \ref{fig:exp1}.(e)-(h) show the learning curves of SAC and TD3 on PyBullet environments and BipdalWalkerHardcore-v3, respectively.
The sampling method LFIW drives better results compared to SAC and TD3 with PER and RANDOM, but its variance in performance is much larger, which means that the training is less stable. On the other hand, our MAQ consistently outperforms baselines in all tested cases. In particular, It significantly
improves the performance of all off-policy RL algorithms on the environment with high-dimensional state and action spaces, i.e., HumanoidPyBulletEnv-v0. This impressive performance gain is made possible by two factors. 1) Our representation for the state-action pair $(s,a)$, which is the input for the $Q$-function, improves by parallel unification of MQN, MReward, and MPER. 2) The Model-augmented TD error in Eq.~\ref{eq:deltaQ} is improved, as the model estimation for the reward and the transition improves. 3) MPER samples the transitions that allows for further improvements in the accurate estimation of the Model-augmented TD error in Eq.~\ref{eq:deltaQ}. 

\textbf{Atari game environments}.
Next, since we focus on how to improve $Q$-learning, we validate our method with Rainbow~\citep{efficientrainbow}, which is a state-of-the-art $Q$-learning framework without a policy network. Figure~\ref{fig:exp1-atari} shows the learning curves of data-efficient Rainbow \citep{efficientrainbow} on various Atari games. MRainbow in the figure is a variant of Rainbow modified with our MAQ, that consists of MQN, MPER, and MReward. We can observe that MRainbow achieves overwhelmingly larger performance compared with the base Rainbow, with both PER and RANDOM on various Atari games. {We provide additional Atari results in the supplementary material.}

\textbf{Sparse reward environment}. Finally, we compare ours on the sparse reward environment, i.e., Pendulum$^{*}$ against two reward shaping methods, CDE, PE in Figure~\ref{fig:exp2}. Similarly to the experimental results on environments with dense rewards, MSAC-MPER and MTD3-MPER, which utilize our method, achieves overwhelmingly better performance over the baseline algorithms. While our MQN succesfully trained on all five instances, most baseline algorithms and sampling methods were unable to learn how to get high cumulative rewards. Although CDE was able to successfully learn good policies on certain instances, it failed on others.

\subsection{Ablation Study}
We now analyze what components in our method are crucial to its improvement of MFRL algorithms' performance.

\begin{figure*}[t!]
\centering
\begin{tabular}{cccc}
    \hspace{5mm} Hopper & \hspace{5mm}HalfCheetah & \hspace{5mm}Hopper & \hspace{5mm}HalfCheetah \\
    \makecell{\includegraphics[width=0.22\columnwidth]{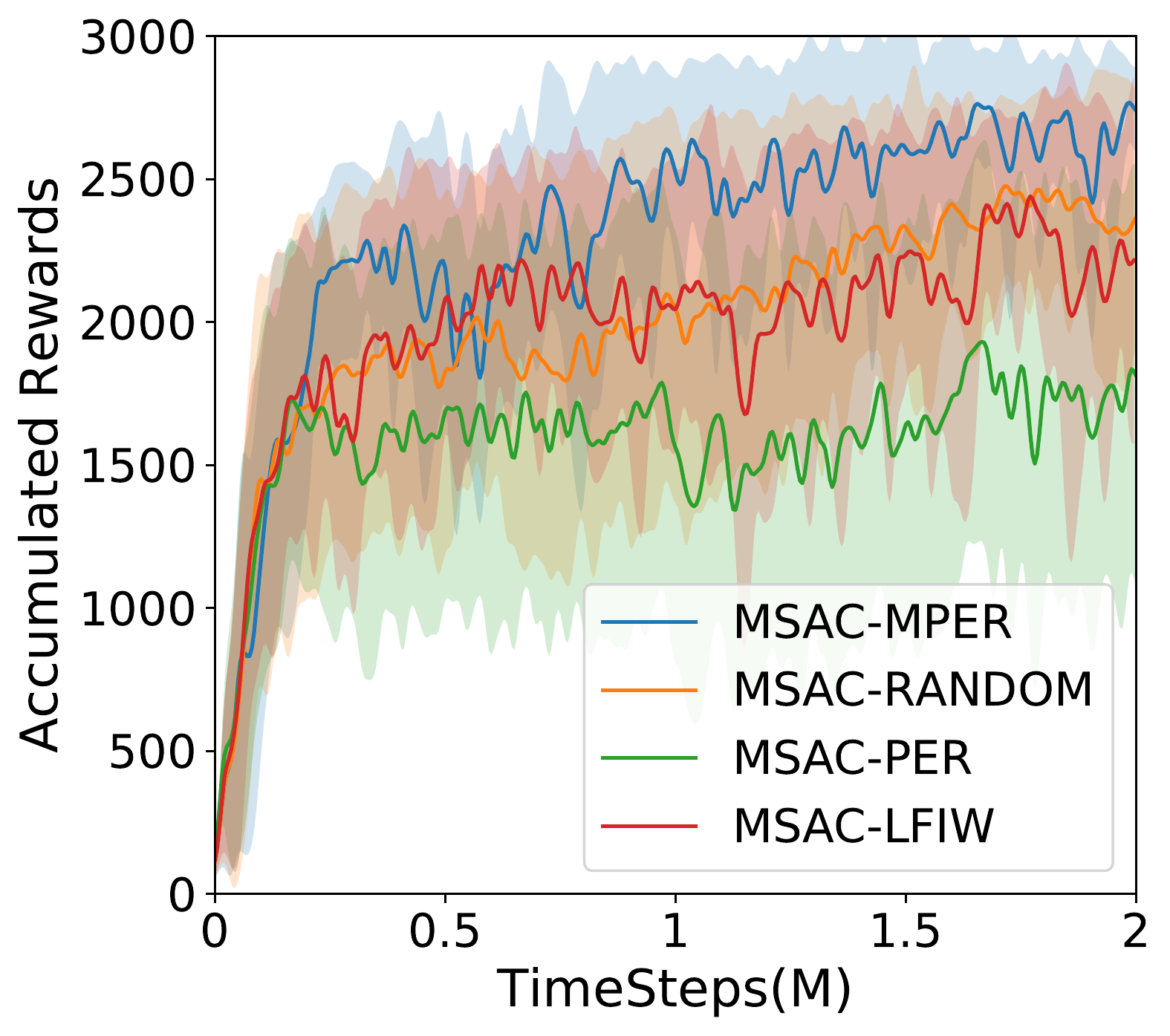}} &
    \makecell{\includegraphics[width=0.22\columnwidth]{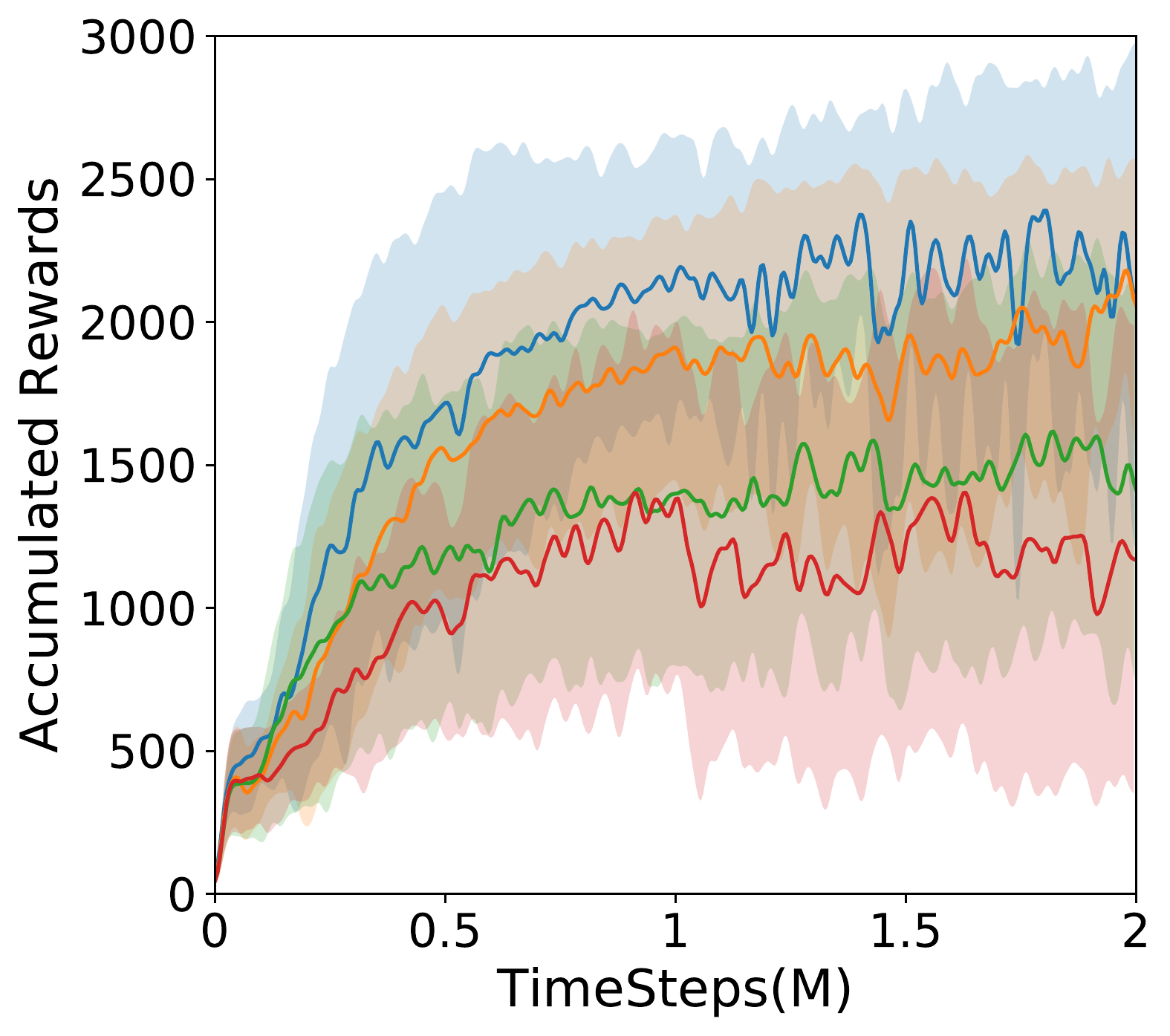}}&
    \makecell{\includegraphics[width=0.22\columnwidth]{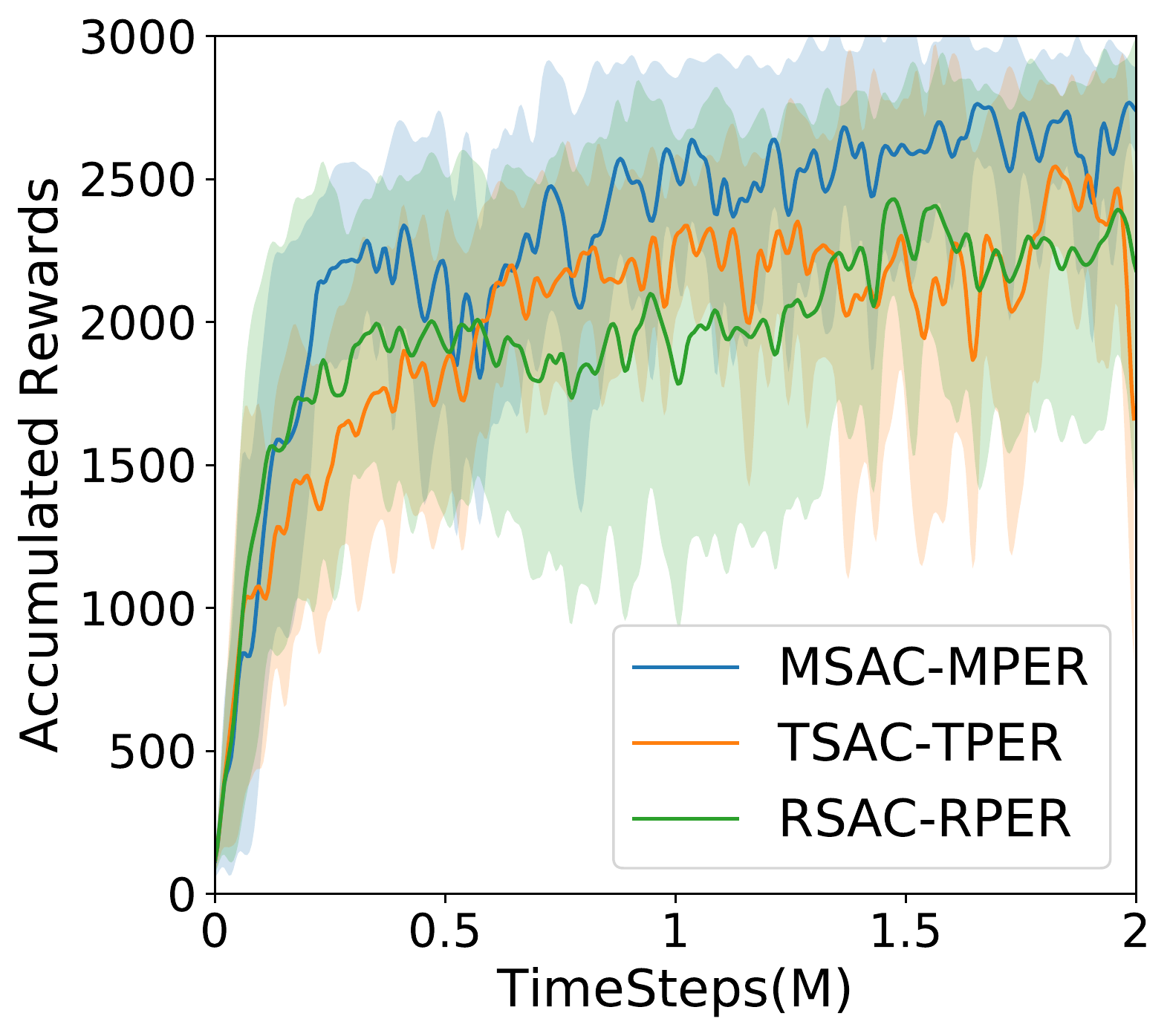}} &
    \makecell{\includegraphics[width=0.22\columnwidth]{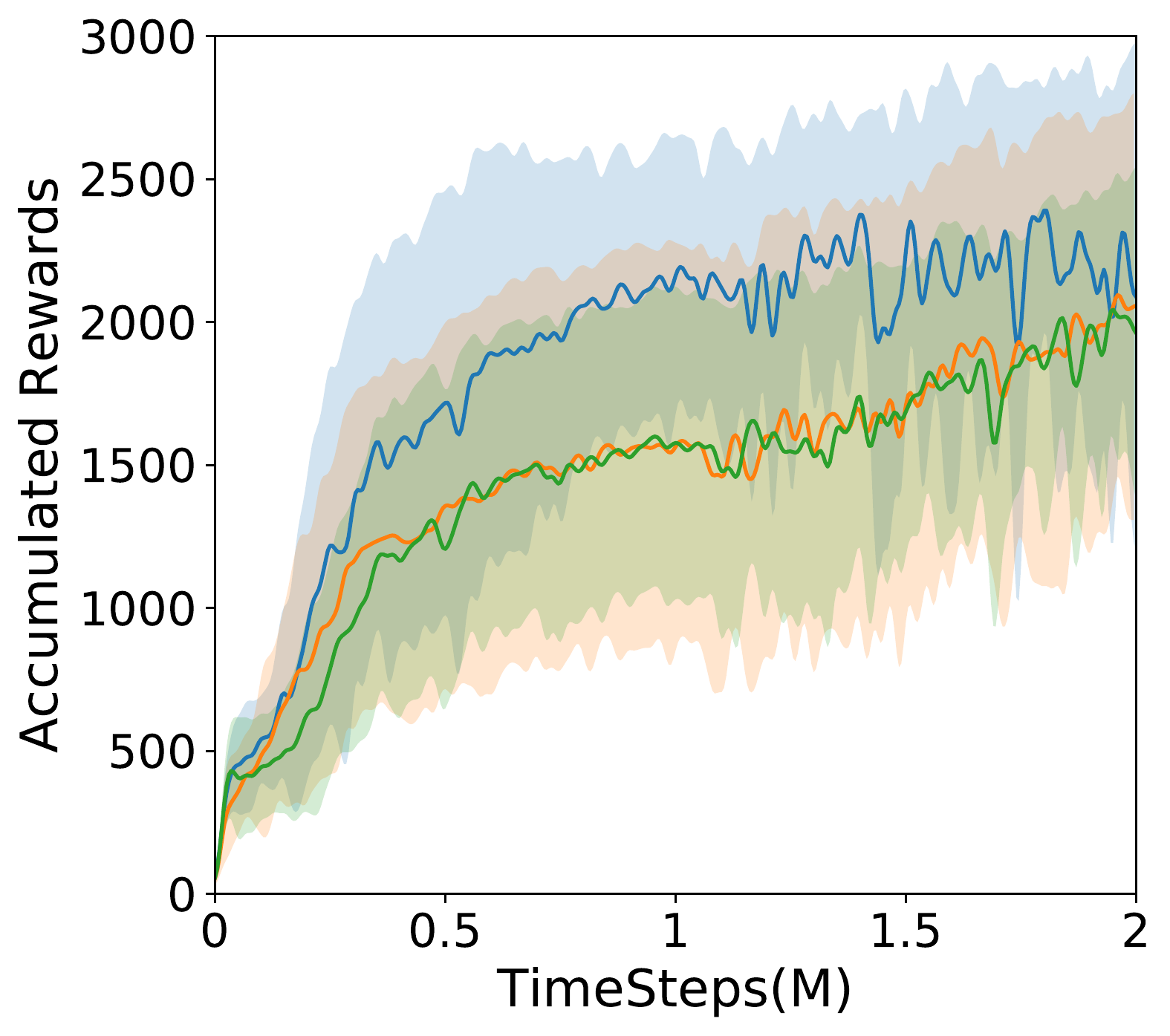}}\\
    {\small (a) Sampling effect}
    & {\small (b) Sampling effect}
    & {\small (c) Model's effect}
    & {\small (d) Model's effect}
\end{tabular}
\caption{\small (a/b): Learning curves of MSAC with different sampling methods. Under MSAC, we observe that the proposed MPER outperforms other sampling methods. (c/d): Learning curves for different variants of MPER for the ablation study. TSAC-TPER refers a variant of MPER with only the transition estimator that prioritizes the PER only with the transition estimation error, and RSAC-RPER refers to a version that only considers and estimates reward. One can observe that learning of both reward and transition maps is the most effective. 
The solid lines and shaded regions represent the mean and standard deviations across five runs  with random seeds.
}
\label{fig:ab1}
\end{figure*}

\textbf{Effectiveness of Sampling Methods.} First, we verify the effectivness of MPER. Since this method is only available with MQN, we set the baseline as MSAC and compare our sampling method MPER with RANDOM, PER, LFIW under MSAC and MTD3, respectively. Figure~\ref{fig:ab1}.(a)-(b) shows the learning curves of MSAC with different sampling methods. We can observe that LFIW and PER are inferior to RANDOM, even with the MQN $\pred_{\theta}$. The main reason is that these two methods focus on sampling transitions that are beneficial in updating the $Q$-network only (i.e., TD-errors and importance weights for the $Q$-loss). Since estimation of the $Q$-value is only a single component of MQN, this is suboptimal in improving our Mode-augmented TD error objective, and thus we need MPER which seeks to improve the estimation of the reward, state, and $Q$-values.

\textbf{Effectiveness of model learning.} Second, we analyze the effectiveness of the model estimation in MAQ. To verify it, we examine the performance of the MSAC variants without the reward or transition estimators. We refer to them as reward-considering (RSAC) with reward-considering PER (RPER)
and transition-considering (TSAC) with transition-considering PER (TPER) respectively. Figure~\ref{fig:ab1}.(c)-(d) show the learning curves of each method. SAC-PER performs the worst and adding each component leads to performance improvement. Also, TSAC-TPER outperforms RSAC-RPER, which suggests that learning the transition map may be more beneficial in improving the Q-value estimation or the representational power of the state-action representations. 

\begin{figure*}[!t]
\centering
\begin{tabular}{cccc}
    \hspace{5mm}Humanoid & \hspace{5mm}Humanoid & \hspace{5mm}Humanoid &
    \hspace{5mm}Humanoid\\
    \makecell{\includegraphics[width=0.205\columnwidth]{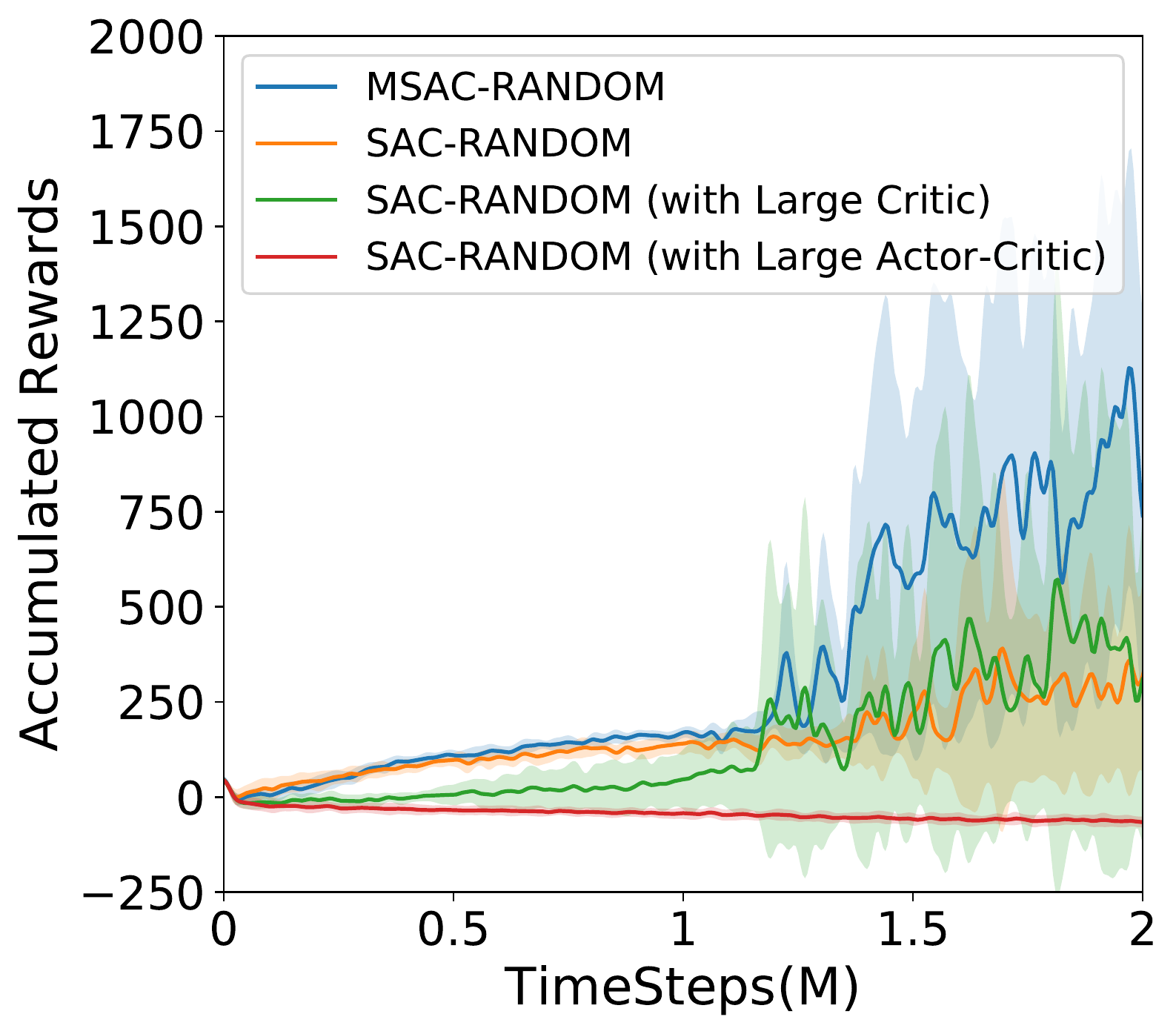}}& \makecell{\includegraphics[width=0.21\columnwidth]{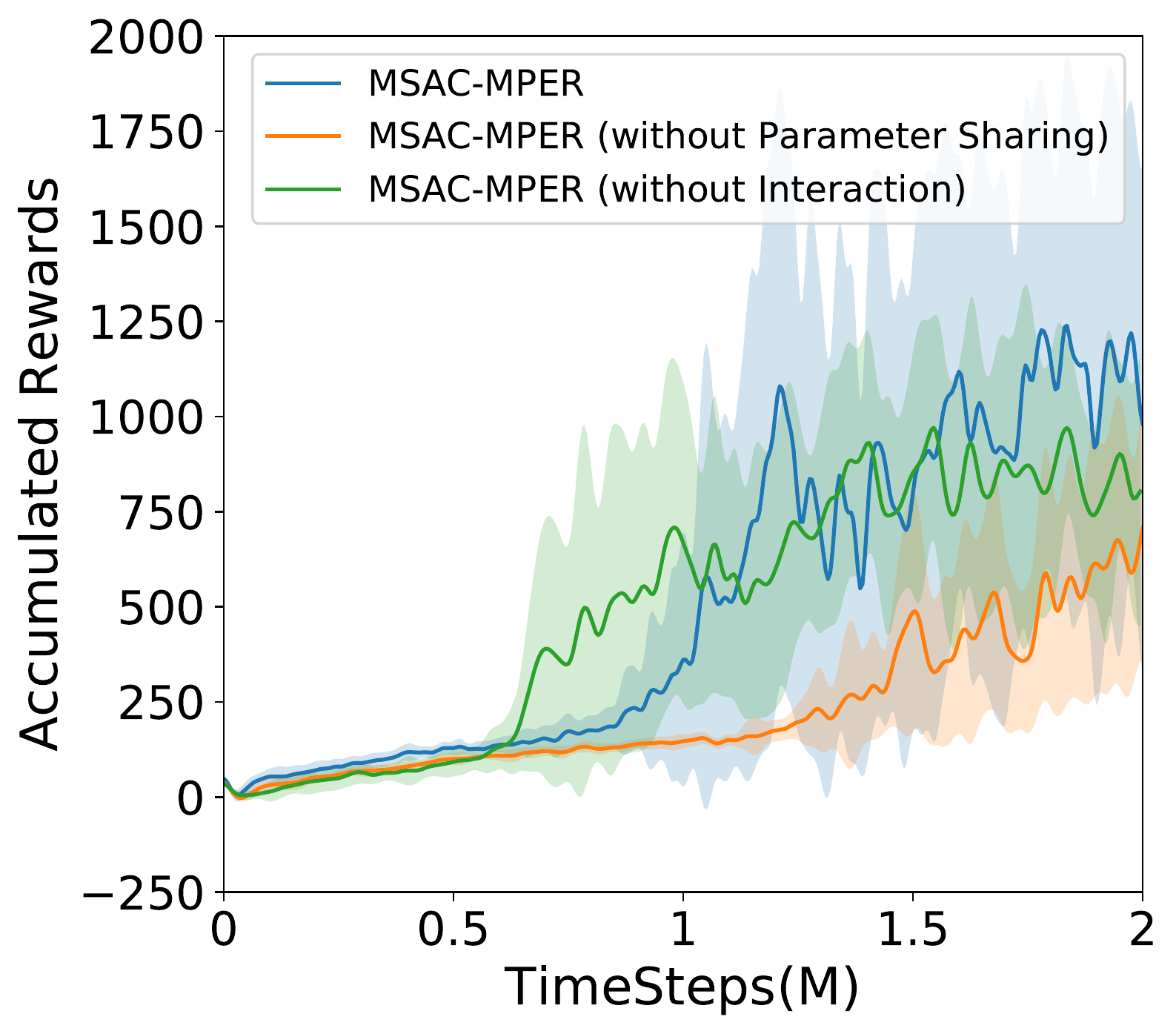}}&
    \makecell{\includegraphics[width=0.21\columnwidth]{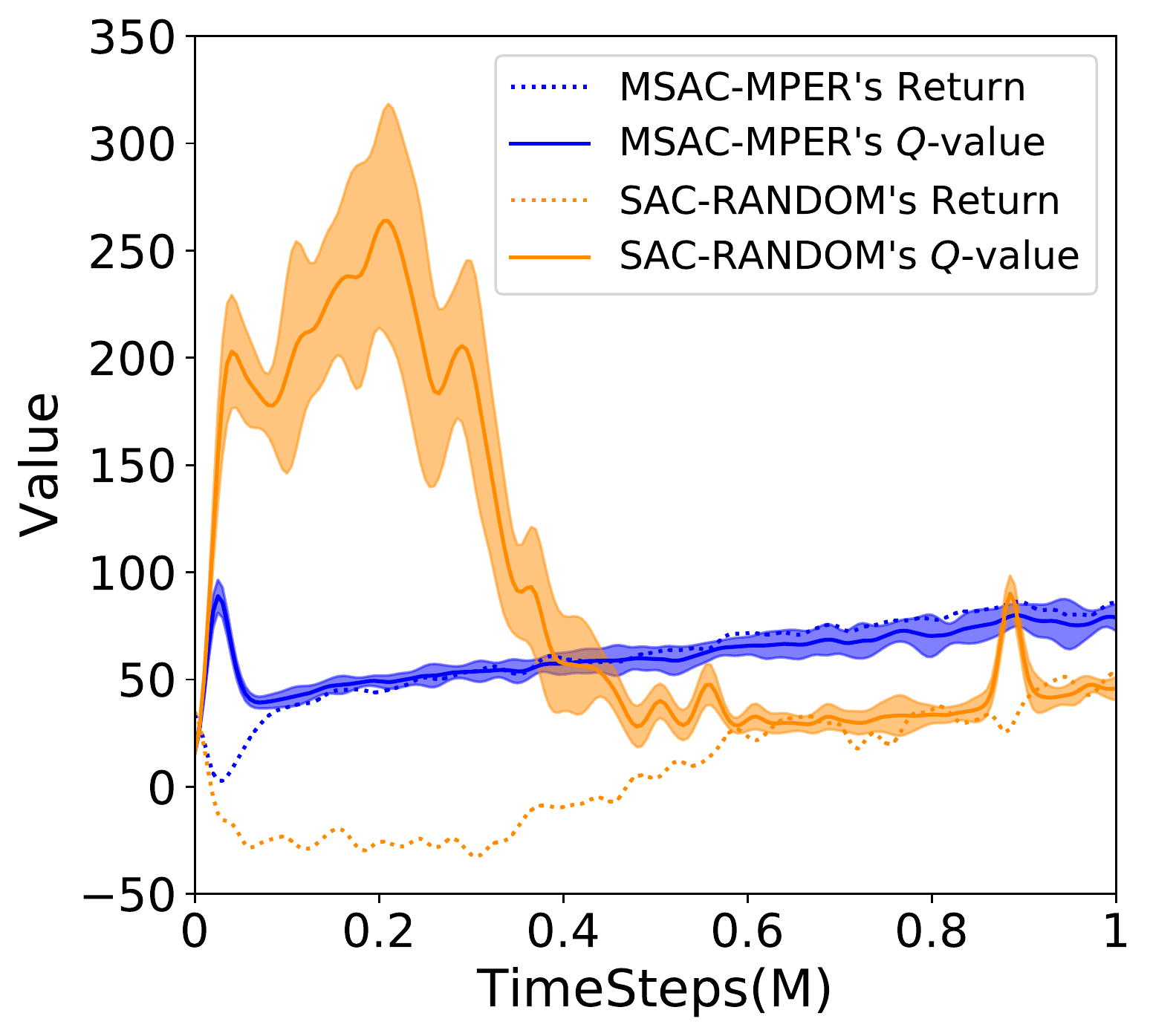}} &
    \makecell{\includegraphics[width=0.205\columnwidth]{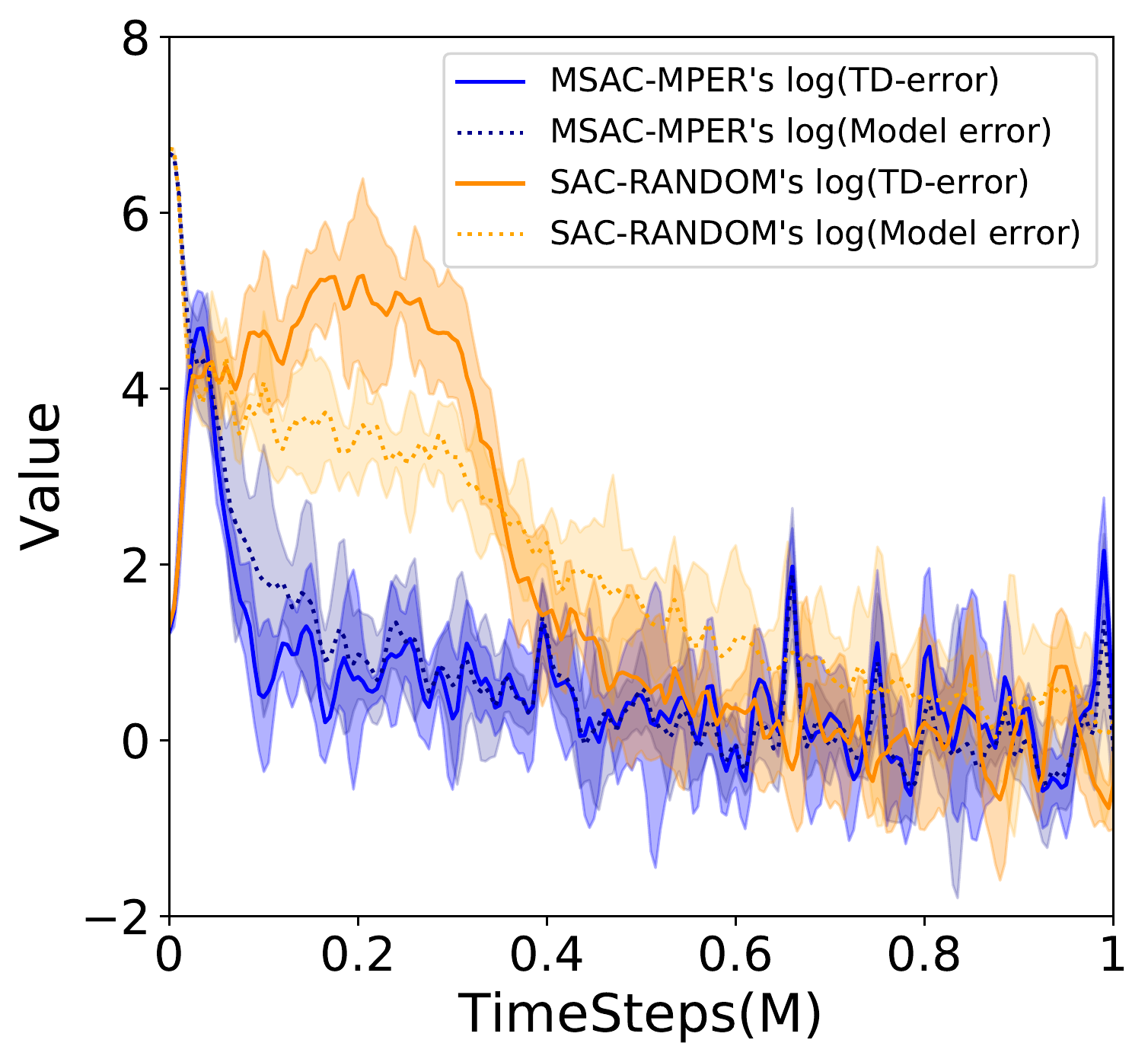}}\\
    {\footnotesize (a) Network size effect} & {\footnotesize (b) Interaction effect} & {\footnotesize (c) Estimation bias} & {\footnotesize (d) Model- and TD-errors}
\end{tabular}
\caption{\small 
(a): Learning curves of SAC with larger networks. Here, large actor and critic are networks whose hidden layer contains $(2500, 2500)$ hidden units, respectively. This shows that simply increasing the number of parameters in the value and the policy network leads to marginal or worse performance. (b): Measured estimation bias in the $Q$-values for MSAC-MPER and SAC-RANDOM, respectively. We observe that MSAC-MPER quickly converges to the true returns, with less than 0.1M steps. On the contrary, SAC-RANDOM largely overestimate the value in earlier training steps. (c): Comparison with and without the interaction of the model and value estimator in~\eqref{eq:deltaQ} or the parameter sharing between the estimators of MQN. MSAC-MPER without interaction leverages the true reward to train the model, and MSAC-MPER without parameter sharing leverages separate networks for each estimator. (d): Normalized model- and TD-error curves.
We can observe that the model- and TD-errors have the same tendency under MSAC-MPER unlike SAC-RANDOM with a separate model-estimation network, which is a direct evidence that the model estimation in MQL is indeed helpful in reducing the TD-errors. The solid lines and shaded regions represent the mean and standard deviations across five runs  with random seeds. 
}
\label{fig:ab2}
\end{figure*}

\textbf{Effectiveness of Increased Parameters.} One may suspect that if MSAC achieves improved performance due to the increased size of the $Q$-network, with additional estimators. To show that this is not the case, we use a base SAC with its hidden layers dramatically increased, from $(300,300)$ to $(2500, 2500)$ for both the Q-network and the policy network. Figure~\ref{fig:ab2}.(a) shows the effect of changes by learning curves. We observe that increasing the hidden layer size alone does not improve the performance and may even lead to the performance degeneration. From these results, it is clear that the improvements of our method is not simply coming from the increased size of hidden layers.

\textbf{Effectiveness of interaction and parameter sharing between the estimators.} We further verify the effectiveness of parameter sharing and interaction across the model and the $Q$-value estimators in MQN, to see where the improvements come from. For this experience, we consider a MQN with separate networks for model and $Q$-value estimation (without parameter sharing) in Figure~\ref{fig:ab2}.(b), and a MQN that utilizes the real reward instead of MReward in  \eqref{eq:deltaQ} to remove the interactions across estimators in Figure~\ref{fig:ab2}.(b). One can observe that removing the parameter sharing critically degrades the sample efficiency of MSAC-MPER. Moreover, although MSAC without the interaction learns well, MSAC-MPER with the interaction finally overtakes it, which shows that learning with the estimator interactions is also useful.

\textbf{Ananalys of value estimation errors.}
To check if our proposed method effectively deals with the overestimation problem, we check how close are the estimated $Q$-values to the true returns. To this end, we compute $Q$-values of initial states and computed true returns via five instances in Figure~\ref{fig:ab2}.(c). We can observe that MSAC-MPER's $Q$-estimation converges to the real returns until less than 0.1M steps. On the contrary, SAC-RANDOM's $Q$-estimation is largely overestimated and does not converge to the true returns before 0.5M steps. These results show that our method dramatically improves the precision of $Q$-learning.

\textbf{Analysis of the relation between model- and TD-errors.}
Figure~\ref{fig:ab2}.(d) shows that model- and TD-errors are highly related, with our MSAC-MPER. We observe that small model-errors generally lead to small TD-errors estimated with the true reward (Eq.~\ref{eq:TD}). Notice that at the beginning, as model-errors decreases rapidly, TD-errors also decreases sharply in succession. This is a direct evidence that estimation of the model with our MQL objective~\eqref{eq:deltaQ} is indeed helpful with the $Q$-learning itself. However, in the case of SAC-RANDOM, which uses a separate model network without the interaction term, the TD-errors and model estimation errors are uncorrelated, and TD-error decreases much slowly compared to MSAC-MPER.

%% file: 5-conclusion.tex
\section{Conclusion} \label{sec:conclusion}
We proposed model-augmented $Q$-learning (MQL), which utilizes a model-augmented $Q$-network (MQN) that estimates not only the $Q$-value but also the reward and transition maps via parameter sharing. However, instead of simply relying on shared parameters, we promote interactions between MQN's estimators by deriving the equation \eqref{eq:deltaQ} given a model-augmented reward (MReward). We also proved that MQL is policy-invariance, i.e., it does not affect the set of Pareto-optimal policies. We further propose Model-augmented Prioritized Experience Replay (MPER), which samples the past experiences based on the model estimation errors, as well as the TD errors. 

The advantages of our proposed method are as follows.
First, It simple to implement and is generally applicable to any MFRL algorithms that utilize $Q$-networks.
Second, MQL dramatically increases sample efficiency of state-of-the-art MFRL algorithms: SAC, TD3, and Rainbow. Third, it largely alleviates the understimation or overestimation of the value with the conventional $Q$-learning. Fourth, the computational cost of our algorithms is almost the same as the original algorithms. Finally, MQL is effective in various environments, including ones with the sparse rewards.

%% file: 6-appendix-arxiv.tex
\appendix
\onecolumn
\clearpage
\counterwithin{figure}{section}
\counterwithin{table}{section}

\begin{center}{\bf {\LARGE Supplementary Material:}}
\end{center}

\begin{center}{\bf {\Large Model-Augmented $Q$-learning}}
\end{center}
\section{Environment Description}
\subsection{PyBullet Environments}
PyBullet environments are the open-source implementations of the OpenAI Gym MuJoCo \citep{todorov2012mujoco} environments, which is currently one of the most widely used toolkits for developing and comparing reinforcement learning algorithms. Since MuJoCo environments are commercial, they hinder open research, so we do not use MuJoCo environments in the manuscript. Besides, these environments are known to be more difficult due to the more realistic improvement, e.g., adding energy cost,  of the existing MuJoCo environments. We provide PyBullet tasks applied in the manuscript:

{\bf HumanoidPyBulletEnv-v0} is an environment to control a three-dimensional bipedal robot to walk quickly without falling over.

{\bf HalfCheetahPyBulletEnv-v0} is an environment to control a two-dimensional cheetah robot for learning sprint.

{\bf HopperPyBulletEnv-v0} is an environment to control a two-dimensional one-legged robot hop-forward quickly without falling over.

\begin{figure*}[!h]
\centering
\begin{tabular}{ccc}
    \makecell{\includegraphics[width=0.33\columnwidth]{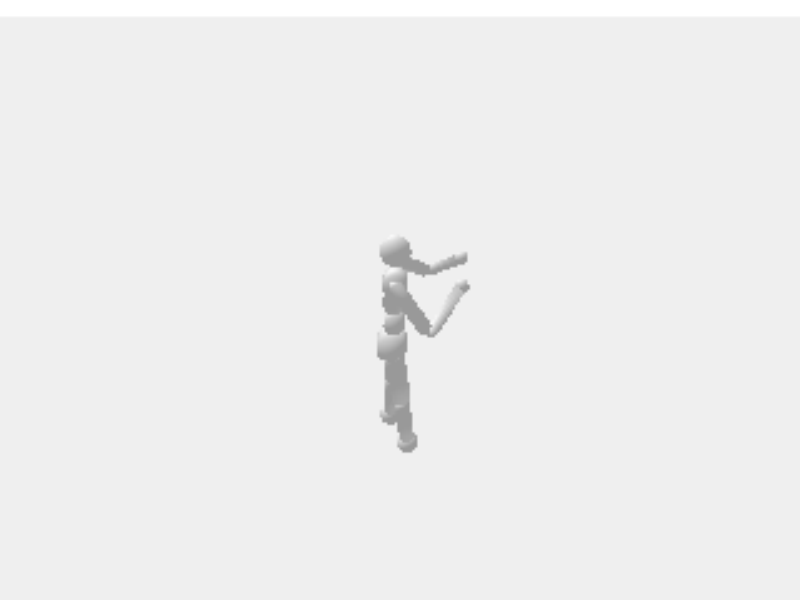}} &
    \makecell{\includegraphics[width=0.33\columnwidth]{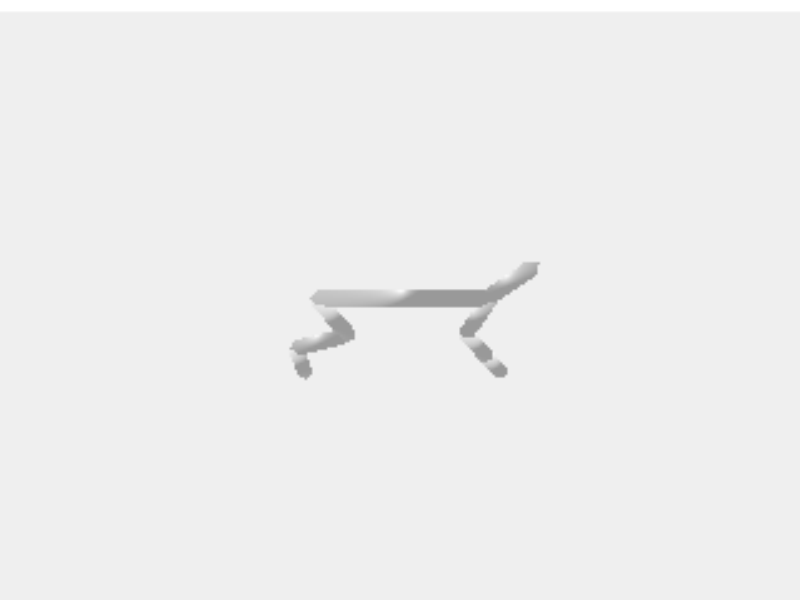}} &
    \makecell{\includegraphics[width=0.33\columnwidth]{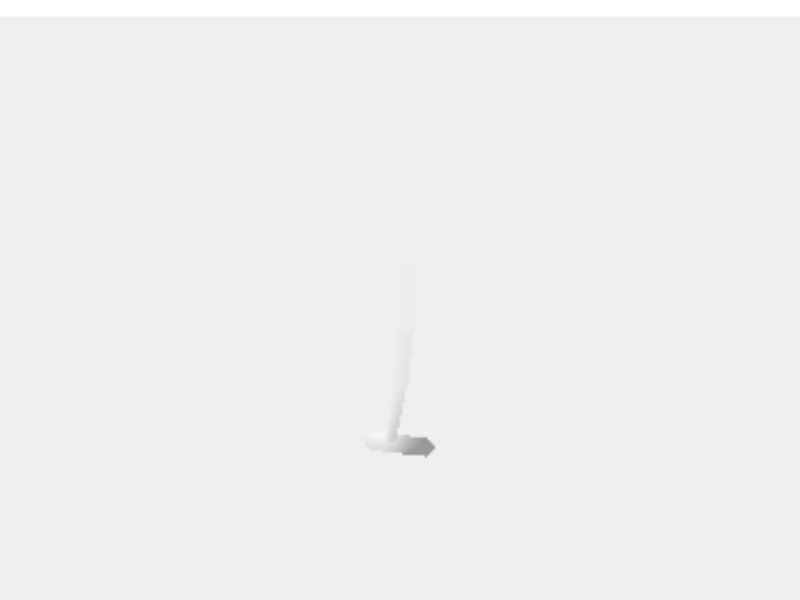}}
    \\
    (a) HumanoidPyBulletEnv-v0 & 
    (b) HalfCheetahPyBulletEnv-v0 & 
    (c) HopperPyBulletEnv-v0
\end{tabular}
\caption{Pybullet environments.
}
\label{fig:exp1}
\end{figure*}

\subsection{OpenAI Gym Environments}
OpenAI Gym \citep{openai} supports continuous control environments that belong to classic or Box2D simulators. We conduct experiments on the following environments among them.

{\bf Pendulum$^{*}$} is an environment which objective is to balance a rod in the upright position as long as possible. It is a variant of Pendulum-v0 that is supported by OpenAI gym. To make it sparser, we impose the following condition: The pendulum begins to receive +1 reward if maintaining the rod in the upright position more than 100 steps continuously.

{\bf BipedalWalkerHardcore-v3} is an environment to control a robot, a variant of BipedalWalker-v3 in OpenAI gym. The robot's objective is to move forward as far as possible while solving many obstacles. 

\begin{figure*}[!h]
\centering
\begin{tabular}{cc}
    \makecell{\includegraphics[width=0.33\columnwidth]{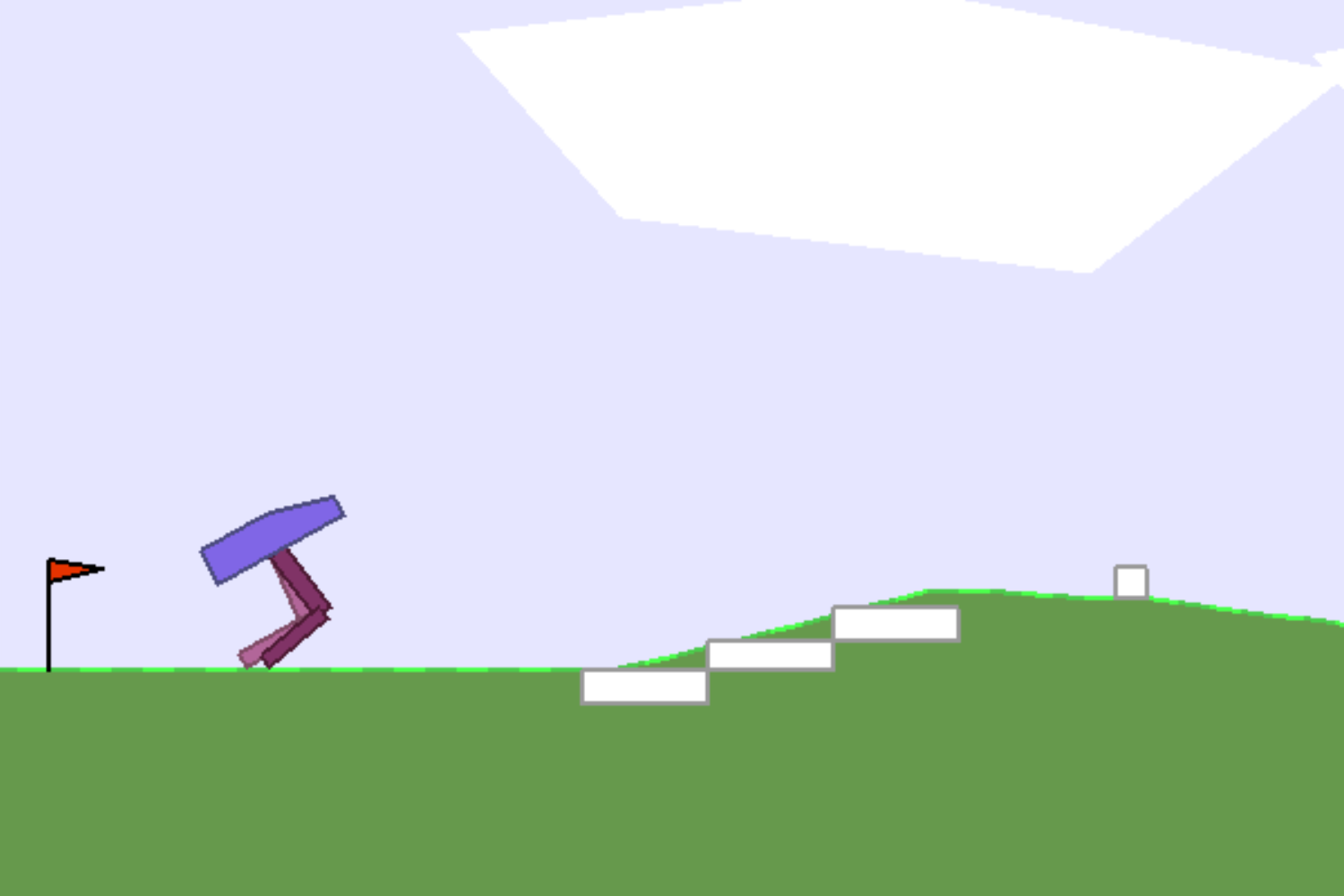}} &
    \makecell{\includegraphics[width=0.33\columnwidth]{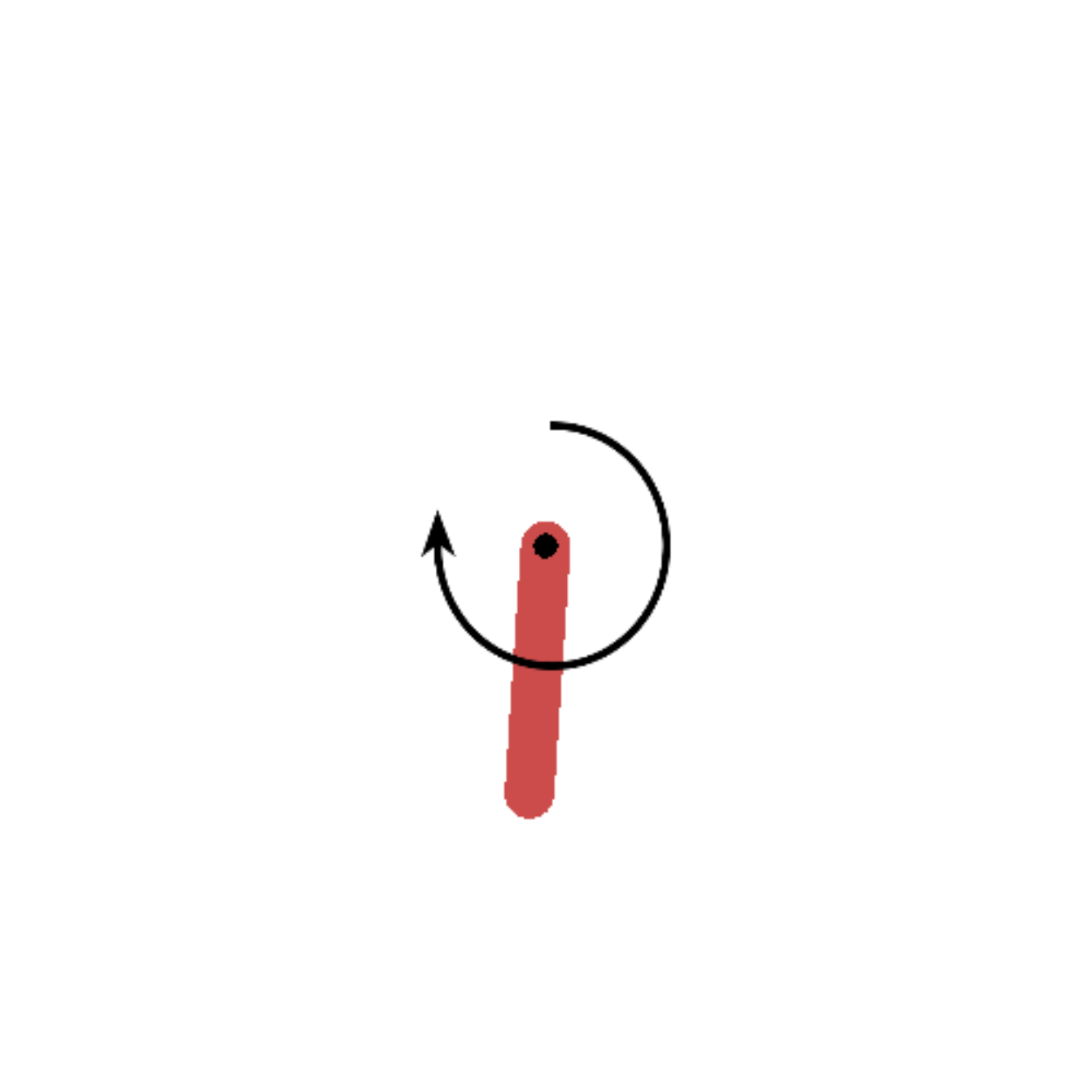}}
    \\
    (a) BipedalWalkerHardcore-v3 & 
    (b) Pendulum-v0
\end{tabular}
\caption{OpenAi Gym environments.
}
\label{fig:exp1}
\end{figure*}

Table~\ref{tab:mujocotable} shows the observation and action spaces and the maximum steps for each episode (horizon) in Pybullet and OpenAI gym environments that we considered. Here, $\mathbb{R}$ and $[-1,1]$ denote sets of real numbers and those between $0$ and $1$, respectively.

\begin{table}[h!]
\begin{center}
\begin{tabular}{lcccr}
\toprule
Environment & Observation space & Action space & Horizon \\
\midrule
HumanoidPybulletEnv-v0 & $\mathbb{R}^{^{44}}$ & $\left[-1,1\right]^{17}$ & $1000$ \\
HalfCheetahPybulletEnv-v0 & $\mathbb{R}^{^{26}}$ & $\left[-1,1\right]^6$ & $1000$ \\
HopperCheetahPybulletEnv-v0  & $\mathbb{R}^{^{15}}$ & $\left[-1,1\right]^3$ & $1000$ \\
BipedalWalkerHardcore-v3 & $\mathbb{R}^{^{24}}$ & $\left[-1,1\right]^4$ & $2000$ \\
Pendulum$^*$ & $\mathbb{R}^{^{3}}$ & $\left[-1,1\right]^1$ & 200\\
\bottomrule
\end{tabular}
\end{center}
\caption{Dimensions of observation and action spaces for continuous control environments}
\label{tab:mujocotable}
\end{table}

\newpage
\subsection{Discrete Control Environments}
We explain Atari environments that we considered in the manuscript and this supplementary material. The objective of RL agents is to learn a policy by observing the screen (RGB) to get high cumulative rewards.

{\bf Alien}: An environment where player should destroy all alien eggs in the RGB screen with escaping aliens.

{\bf Amidar}: An environment similar to MsPacman. In this environment, agents control a monkey in a fixed rectilinear lattice to eat pellets as much as possible while avoiding chasing masters.

{\bf Assault}:  An environment to control a spaceship. which objective is to eliminate the enemies.

{\bf Asterix}: An environment to control a tornado. Its objective is to eat hamburgers in the screen with avoiding dynamites. 

\begin{figure*}[!h]
\centering
\begin{tabular}{cccc}
    \makecell{\includegraphics[width=0.24\columnwidth]{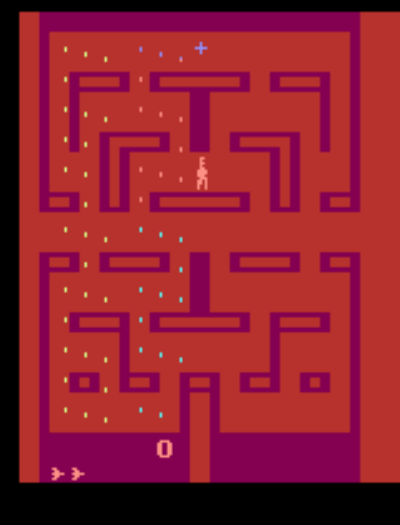}} &
    \makecell{\includegraphics[width=0.20\columnwidth]{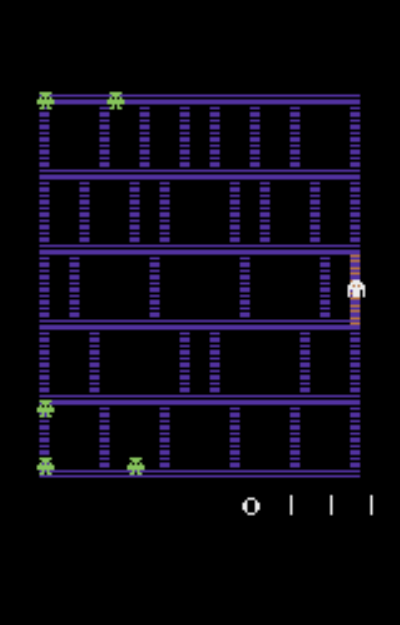}} &
    \makecell{\includegraphics[width=0.20\columnwidth]{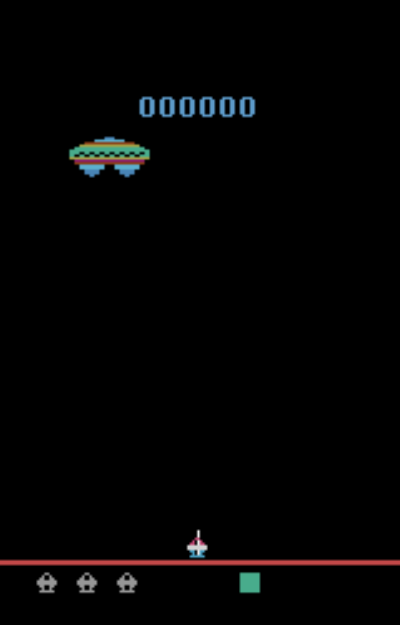}} & 
    \makecell{\includegraphics[width=0.24\columnwidth]{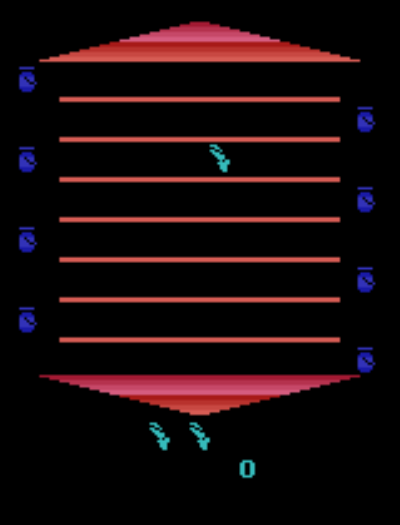}} \\ 
    (a) Alien & (b) Amidar & (c) Assault & (d) Asterix 
    
\end{tabular}
\caption{\small 
Atari Games: Alien, Amidar, Assault, and Asterix.
}
\label{fig:app-atari1}
\end{figure*}

{\bf BankHeist}: An environment to control a robber. The objective of the game is to rob banks as many as possible while avoiding the police in maze-like cities..

{\bf DemonAttack}: An environment to control a guardian.The guardian should kill demons that attacks from above.

{\bf Frostbite}: An environment to control a man who should collect ice blocks to make his igloo. His objective is to collect 15 ice blocks while avoiding some opponents, e.g., crabs and birds.

{\bf Gopher}: An environment to control a farmer. The farmer should protect three carrots from a gopher.

\begin{figure*}[!h]
\centering
\begin{tabular}{cccc}
    \makecell{\includegraphics[width=0.185\columnwidth]{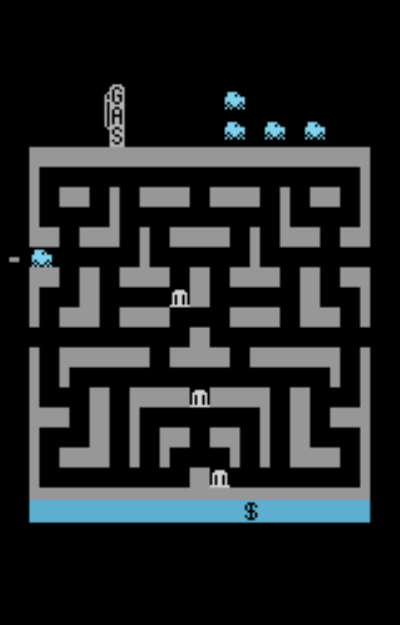}} &
    \makecell{\includegraphics[width=0.22\columnwidth]{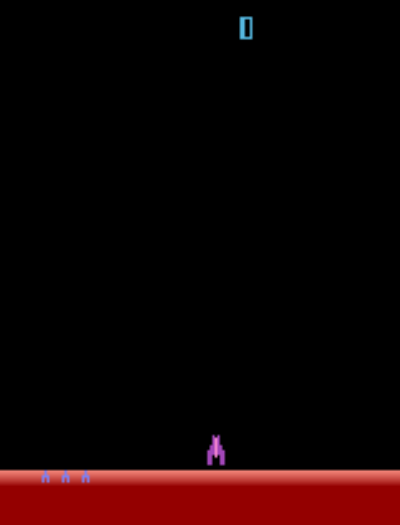}} &
    \makecell{\includegraphics[width=0.22\columnwidth]{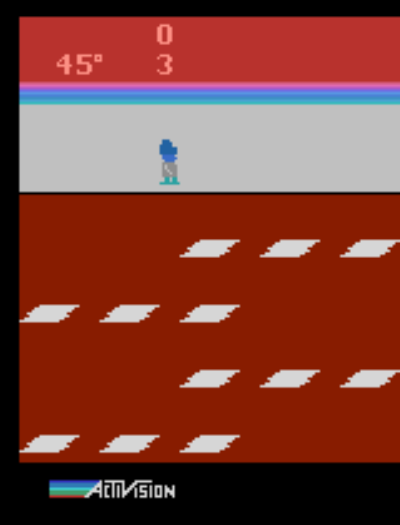}} & 
    \makecell{\includegraphics[width=0.185\columnwidth]{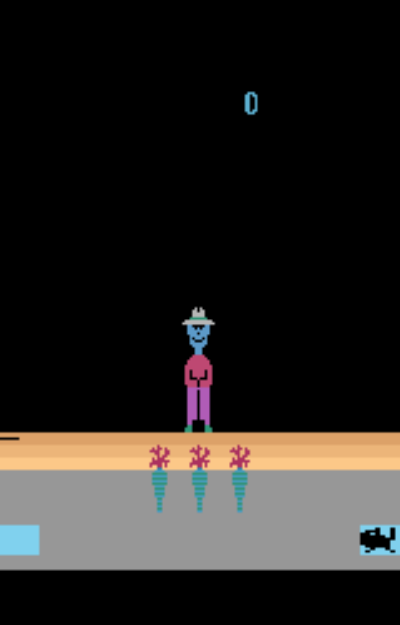}} \\ 
    (a) BankHeist & (b) DemonAttack & (c) FrostBite & (d) Gopher
    
\end{tabular}
\caption{\small 
Atari Games: BankHesit, DemonAttack, Frostbite, and Gopher.
}
\label{fig:app-atari2}
\end{figure*}

{\bf Jamesbond}:  An environment to control a vehicle. The objective is to move forward while avoiding and attacking enemies.

{\bf Kangaroo}:  An environment to control a mother kangaroo. The kangaroo's objective is to rescue her son while climbing.

{\bf Krull}:  An environment to control a player following a file of the same title. The player should complete stages, which are main parts in the film.

{\bf Seaquest}:  An environment to control a submarine. Its objective is to rescue divers while attacking enemies by missiles. 
\begin{figure*}[!h]
\centering
\begin{tabular}{cccc}
    \makecell{\includegraphics[width=0.2\columnwidth]{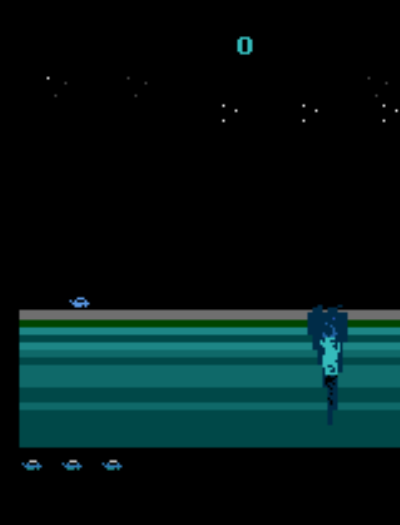}} &
    \makecell{\includegraphics[width=0.2\columnwidth]{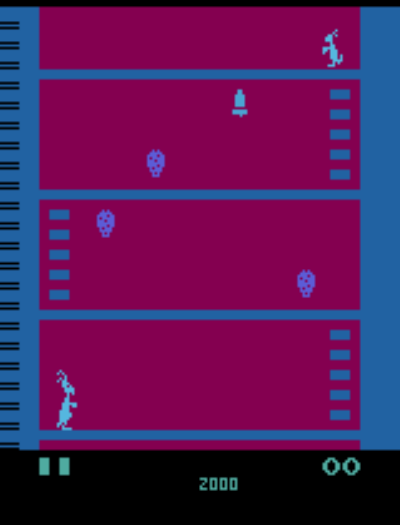}} &
    \makecell{\includegraphics[width=0.2\columnwidth]{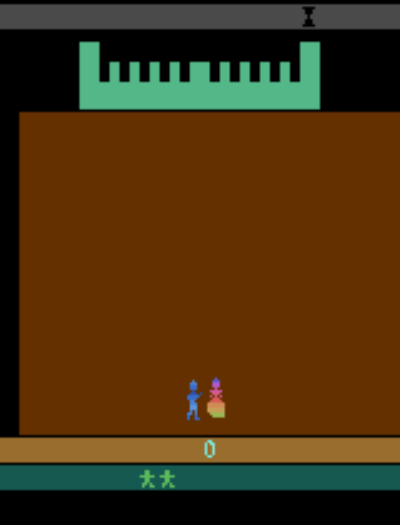}} & 
    \makecell{\includegraphics[width=0.2\columnwidth]{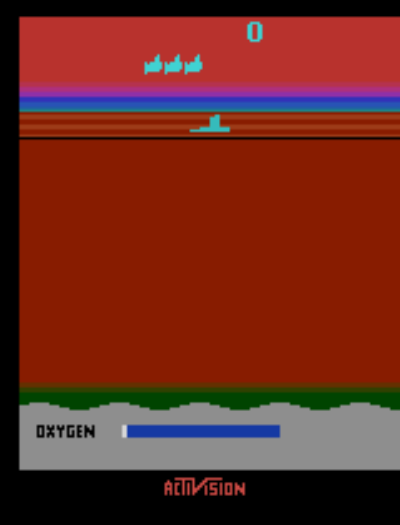}} \\ 
    (a) Jamesbond & (b) Kangaroo & (c) Krull & (d) Seaquest
    
\end{tabular}
\caption{\small 
Atari Games: Jamesbond, Kangaroo, Krull, and Seaquest.
}
\label{fig:ab2}
\end{figure*}

\clearpage
\section{Training details} \label{sec:training-details}
Table~\ref{tab:hp} shows hyper-parameters that we used in experiments of the manuscript. Notice that MQN barely needs hyper-parameters. In the case of Rainbow, instead of taking RGB array as input and output, MQN's transition map takes a context by CNN layers for the current and next RGB arrays as input and output.

\begin{table}[h!]
\label{tab:commontable}
\begin{center}
\begin{small}
\begin{tabular}{lc}
\toprule
Parameter & Value \\
\midrule
Shared \\
\hspace{5mm} Batch size (continuous control environments)    &  $256$ (SAC), $100$ (TD3), $32$ (Rainbow)\\
\hspace{5mm} Buffer size    & $10^{6}$ \\
\hspace{5mm} Target smoothing coefficient $(\tau)$ for soft update & $5\times10^{-3}$ \\
\hspace{5mm} Initial prioritized experience replay buffer exponents $(\alpha, \beta)$ \tablefootnote{$\beta$ increases to $1.0$ by the rule $\beta=0.4\eta+1.0(1-\eta)$, where $\eta =$ the current step/the maximum steps.}    &  $(0.7,$ $0.4)$ (SAC/TD3), $(0.5,$ $0.4)$ (Rainbow)\\ 
\hspace{5mm} Discount factor for the agent reward ($\gamma$) & $0.98$ (SAC/TD3), $0.99$ (Rainbow) \\
\hspace{5mm} Number of initial random actions (continuous control environments) & $5 \times 10^{3}$ \\ 
\hspace{5mm} Number of initial random actions (discrete control environments) & $10,000$ (SAC/TD3) $1,600$ (Rainbow) \\ 
\hspace{5mm} Optimizer & Adam \citep{kingma2014adam}\\
\hspace{5mm} Nonlinearity & ReLU \\
\hspace{5mm} Replay period & 64 (SAC/TD3), 1 (Rainbow) \\
\hspace{5mm} Gradient step & 64 (SAC/TD3), 1 (Rainbow) \\
\midrule
MQN \\
\hspace{5mm} MReward coefficients $(\rewardcoefone$, $\rewardcoeftwo)$ &
$(10^{-3}$, $10^{-3})$\\
\midrule
Likelihood-free Importance Weights\\
\hspace{5mm} Temperature $(T)$ & 5\\
\hspace{5mm} Hidden units per layer & 256, 256\\
\hspace{5mm} Fast replay buffer size & $10^{4}$\\
\midrule
TD3 \\
\hspace{5mm} Hidden units per layer & 400, 300 \\
\hspace{5mm} Learning rate & $10^{-3}$\\
\hspace{5mm} Policy update frequency & $2$\\
\hspace{5mm} Gaussian action and target noises & $0.1,$ $0.2$ \\
\hspace{5mm} Target noise clip & 0.5\\
\hspace{5mm} Target network update& Soft update with interval $1$\\
\midrule
SAC \\
\hspace{5mm} Hidden units per layer & 400, 300 \\
\hspace{5mm} Learning rate & $7.3\times10^{-4}$\\
\hspace{5mm} Target entropy& $-\dim A$  ($A$ is action space)\\
\hspace{5mm} Soft update with interval & $1$
\\
\midrule
Rainbow\\
\hspace{5mm} Observation down-sampling for Atari RGB & $84\times84$ with grey-scaling\\
\hspace{5mm} CNN channels for Atari environments & 32, 64 \\
\hspace{5mm} CNN filter size for Atari environments & $5\times5$, $5\times5$\\
\hspace{5mm} CNN stride for Atari environments & 5, 5 \\
\hspace{5mm} Action repetitions and Frame stack & 4 \\ 
\hspace{5mm} Reward clipping & True ($[-1,1]$) \\
\hspace{5mm} Terminal on loss of life & True\\
\hspace{5mm} Max frames per episode & $1.08\times 10^{5}$\\
\hspace{5mm} Target network update & Hard update (every 2,000 updates) \\
\hspace{5mm} Support of $Q$-distribution & 51 \\
\hspace{5mm} $\epsilon$ for Adam optimizer & $1.5\times 10^{4}$ \\
\hspace{5mm} Learning Rate & $10^{-4}$\\
\hspace{5mm} Max gradient norm & 10 \\
\hspace{5mm} Noisy nets parameter & 0.1 \\
\hspace{5mm} Multi-step return length & 20 \\
\hspace{5mm} $Q$-network's hidden units per layer & 256\\
\bottomrule
\end{tabular}
\end{small}
\end{center}
\caption{Hyper-parameters} \label{tab:hp}
\end{table}

\section{The Proof of Theorem 1} \label{sec:related}
\begin{proof}
We first define the following notations
\begin{align*}
\|(x_{1},\cdots,x_{n})\|_{2}^{2} & =\frac{1}{2}\sum_{i=1}^{n}x_{i}^{2},\\
\|(x_{1},x_{2},x_{3},\cdots,x_{n})|| & =||x_{1}||_{\infty(S)}+\sum_{i=1}^{n}||x_{i}||_{2}^{2},\\
\|x||_{\infty(S)} & =\max_{s\in S}x(s),
\end{align*}
where $S$ is a state space. Let $\boldsymbol{R}$ and $\boldsymbol{S}$
be defined by 
\begin{align*}
\boldsymbol{\boldsymbol{R}} & =\{\mathcal{R}(s,\pi(s))|s\in S\} \in\mathbb{R}^{\dim(S)},\\
\boldsymbol{\boldsymbol{S}} & =\{ \mathcal{T}(s,\pi(s))|s\in S\} \in\mathbb{R}^{\dim(S)},
\end{align*}
where $\left(\mathcal{R},\mathcal{T}\right)$ is an environment's
model. By defining an operator $\mathcal{P}^{\pi}$ by 
\begin{align*}
\mathcal{P}^{\pi}(q,r,s) &=\left(\mathcal{P}_{1}^{\pi}(q,r,s),\mathcal{P}_{2}^{\pi}(q,r,s),\mathcal{P}_{3}^{\pi}(q,r,s)\right),\\
\mathcal{P}_{1}^{\pi}(q,r,s) & =r+\zeta_{1}\|r-\boldsymbol{\boldsymbol{R}}\|_{2}+\zeta_{2}\|s-\boldsymbol{S}\|_{2}\\&+\gamma q(\boldsymbol{S},\pi(\cdot|\boldsymbol{\boldsymbol{S}})),\\
\mathcal{P}_{2}^{\pi}(q,r,s) & =r-\kappa_{1}(r-\boldsymbol{\boldsymbol{R}}),\\
\mathcal{P}_{3}^{\pi}(q,r,s) & =s-\kappa_{2}(s-\boldsymbol{\boldsymbol{S}}),
\end{align*}
where $\kappa_{1},\kappa_{2}$ are small positive constants, $q\in F:=\{f|f:S\times A\rightarrow\mathbb{R}\},r\in\mathbb{R},s\in\mathbb{R}^{\dim(S)}$,
one can compute
\begin{align*}
\left|\mathcal{P}_{2}^{\pi}(q^{1},r^{1},s^{1})-\mathcal{P}_{2}^{\pi}(q^{2},r^{2},s^{2})\right| & \leq\left|(1-\kappa_{1})(r^{1}-r^{2})\right|,\\
\left|\mathcal{P}_{3}^{\pi}(q^{1},r^{1},s^{1})-\mathcal{P}_{3}^{\pi}(q^{2},r^{2},s^{2})\right| & \leq\left|(1-\kappa_{2})(s^{1}-s^{2})\right|.
\end{align*}
Notice that the following inequalities are obtained:
\begin{align}
\left|\|x||_{2}-\|y||_{2}\right| & \leq\|x-y||_{2},\label{eq:ite}\\
\|x+y||_{2} & \leq\|x||_{2}+\|y||_{2}\label{eq:te}
\end{align}
Using (\ref{eq:ite})-(\ref{eq:te}), one can also calculate
\begin{align*}
&||\mathcal{P}^{\pi}(q^{1},r^{1},s^{1})-\mathcal{P}^{\pi}(q^{2},r^{2},s^{2})|| \\ &\leq\left|\zeta_{1}\|r^{1}-r^{2}\|_{2}+\zeta_{2}\|s^{1}-s^{2}\|_{2}\right|\\
 & +\gamma\max_{s\in S}\left|q^{1}(s,\pi(\cdot|s))-\gamma q^{2}(s,\pi(\cdot|s)\right|,
\end{align*}
so that
\begin{align}
&||\mathcal{P}^{\pi}(q^{1},r^{1},s^{1})-\mathcal{P}^{\pi}(q^{2},r^{2},s^{2})|| \nonumber
\\
&\leq\max\left(\gamma,\zeta_{1},\zeta_{2},(1-\kappa_{1}),(1-\kappa_{2})\right)\nonumber\\
&\times ||q^{1}-q^{2},r^{1}-r^{2},s^{1}-s^{2}||.\label{eq:cont}
\end{align}
Therefore, the Banach fixed point theorem \citep{agarwal2018banach} is available
by (\ref{eq:cont}) due to $$\max\left(\gamma,\zeta_{1},\zeta_{2},(1-\kappa_{1}),(1-\kappa_{2})\right)<1,$$ so there exists a unique $\left(q^{*},r^{*},s^{*}\right)$
such that 
\[
\mathcal{P}^{\pi}\left(q^{*},r^{*},s^{*}\right)=\left(q^{*},r^{*},s^{*}\right).
\]
Finally, let $q^{0}\in F$ such that 
\[
q^{0}=\boldsymbol{r}+\gamma q^{0}(\boldsymbol{S},\pi(\cdot|\boldsymbol{\boldsymbol{S}})).
\]
Since 
\begin{align*}
r^{*} & =r^{*}-\kappa_{1}(r^{*}-\boldsymbol{\boldsymbol{R}}),\\
s^{*} & =s^{*}-\kappa_{2}(s^{*}-\boldsymbol{\boldsymbol{S}}),
\end{align*}
it is easy to get $r^{*}=\boldsymbol{\boldsymbol{R}},s^{*}=\boldsymbol{\boldsymbol{S}}.$
Therefore, 
\[
q^{*}=\boldsymbol{r}+\gamma q^{*}(\boldsymbol{S},\pi(\cdot|\boldsymbol{\boldsymbol{S}})).
\]
Since $q^{*}$ is a unique solution in $F,$ one can conclude that  $q^{*}\equiv q_{0}$. This completes the proof.
\end{proof}

\input{4-related-arxiv}
\clearpage
\section{Example of Implementating Our Method}
To verify that our method is easy to implement, as we mentioned in the manuscript, we explain how we can apply our method to Soft Actor Critic (SAC) \citep{sac, sac2} based on PyTorch 1.7.1. In SAC, it uses double critic networks and their target networks. We can implement model-augmented Q-networks (MQNs) from the given critic networks as 
\begin{figure*}[h!]
\centering
\begin{tabular}{c}
    \makecell{\includegraphics[width=1\columnwidth]{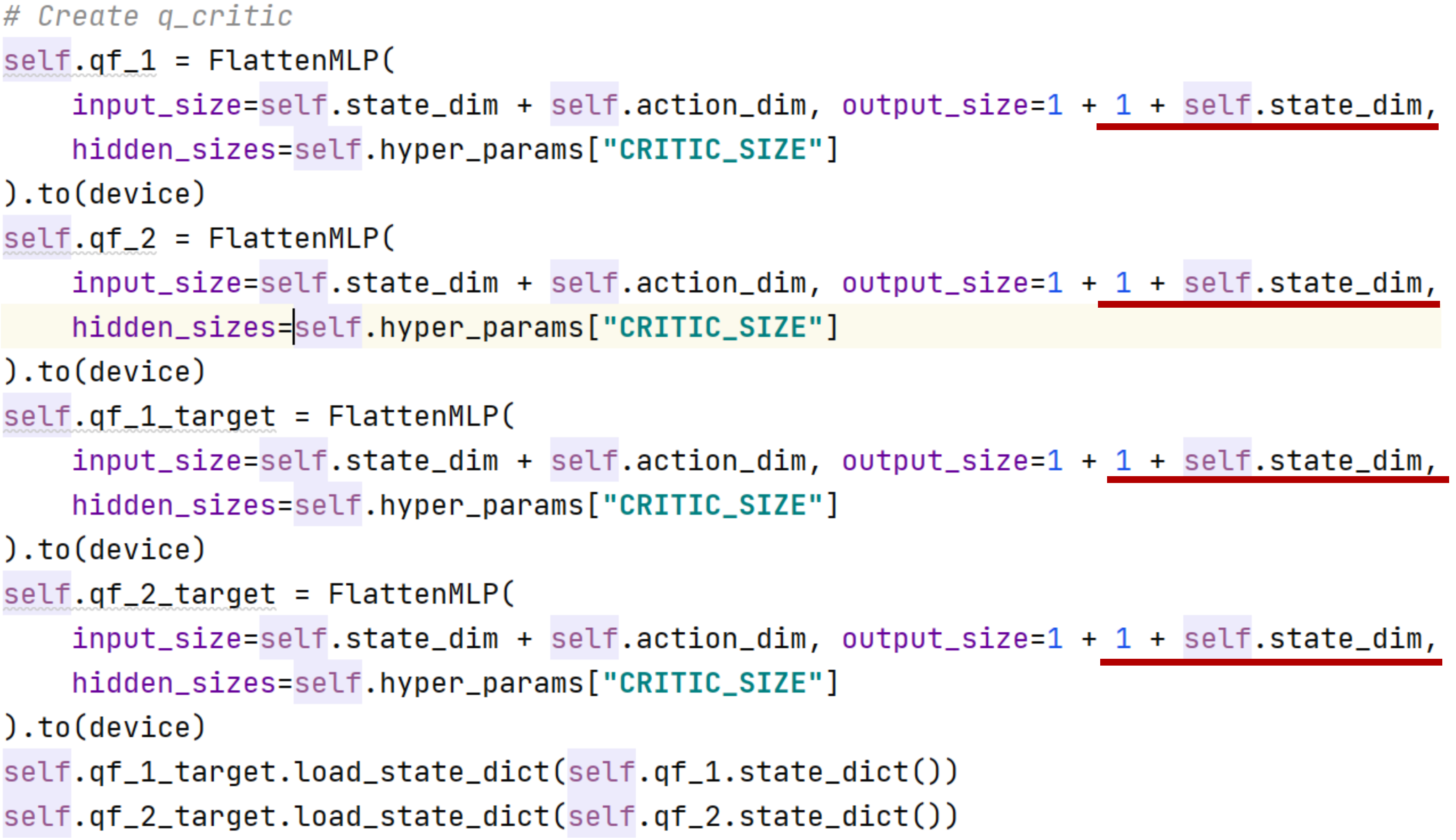}}
\end{tabular}
\caption{Implementation of MQNs. The parts with red lines are the newly added parts, respectively.
}
\label{fig:code1}
\end{figure*}

\begin{figure*}[t!]
\centering
\begin{tabular}{c}
    \makecell{\includegraphics[width=1\columnwidth]{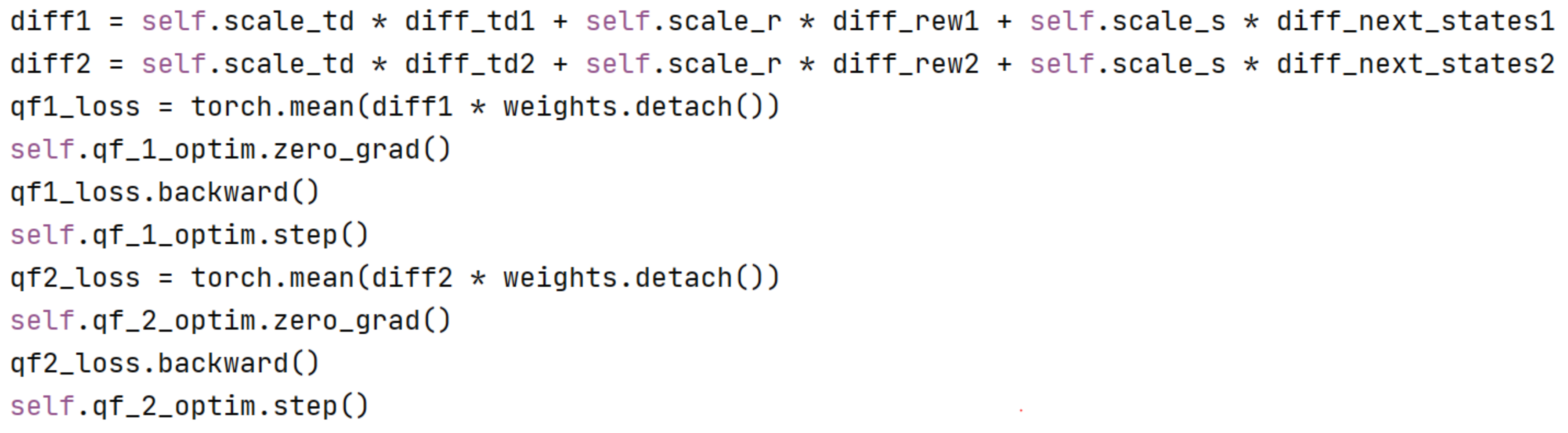}}
\end{tabular}
\caption{Formulating MQN's loss.
}
\label{fig:code3}
\end{figure*}

\begin{figure*}[t!]
\centering
\begin{tabular}{c}
    \makecell{\includegraphics[width=1\columnwidth]{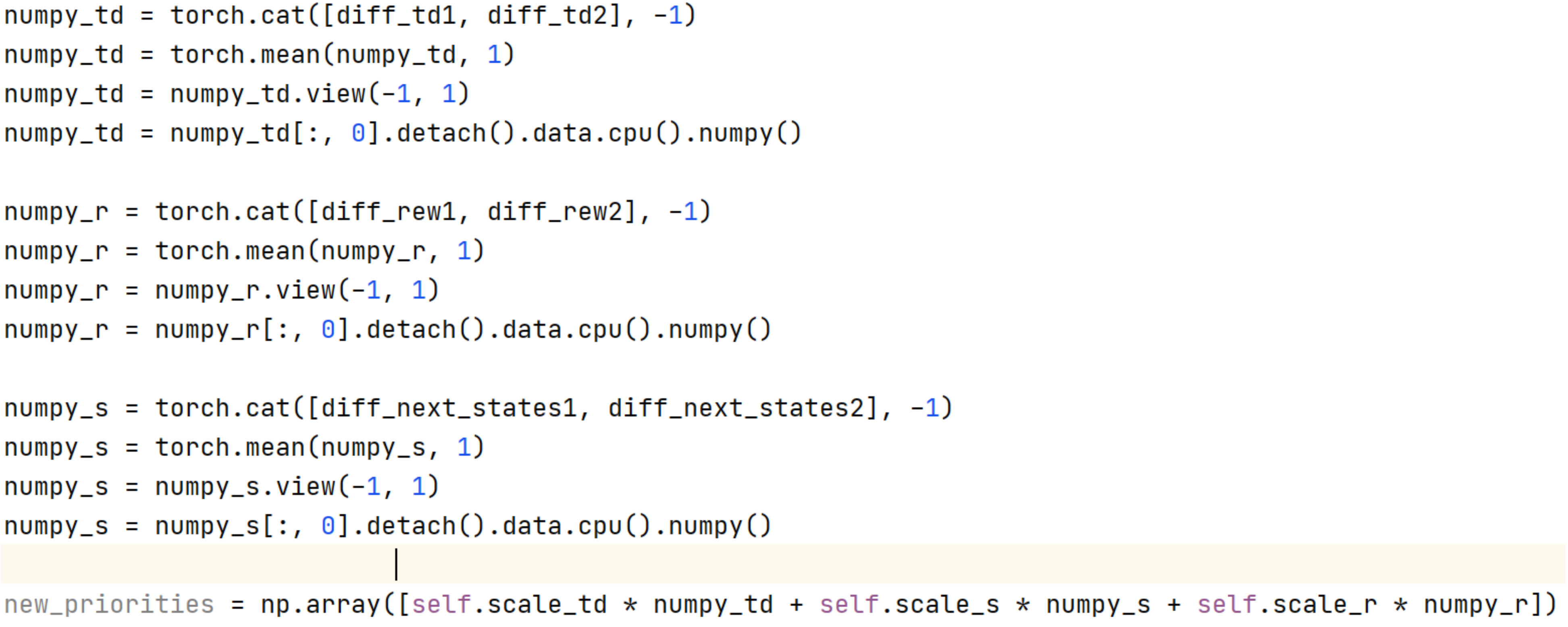}}
\end{tabular}
\caption{Formulating priorities of samples.
}
\label{fig:code4}
\end{figure*}

In the update step of SAC, we can compute model errors as follows:

\begin{figure*}[h!]
\centering
\begin{tabular}{c}
    \makecell{\includegraphics[width=1\columnwidth]{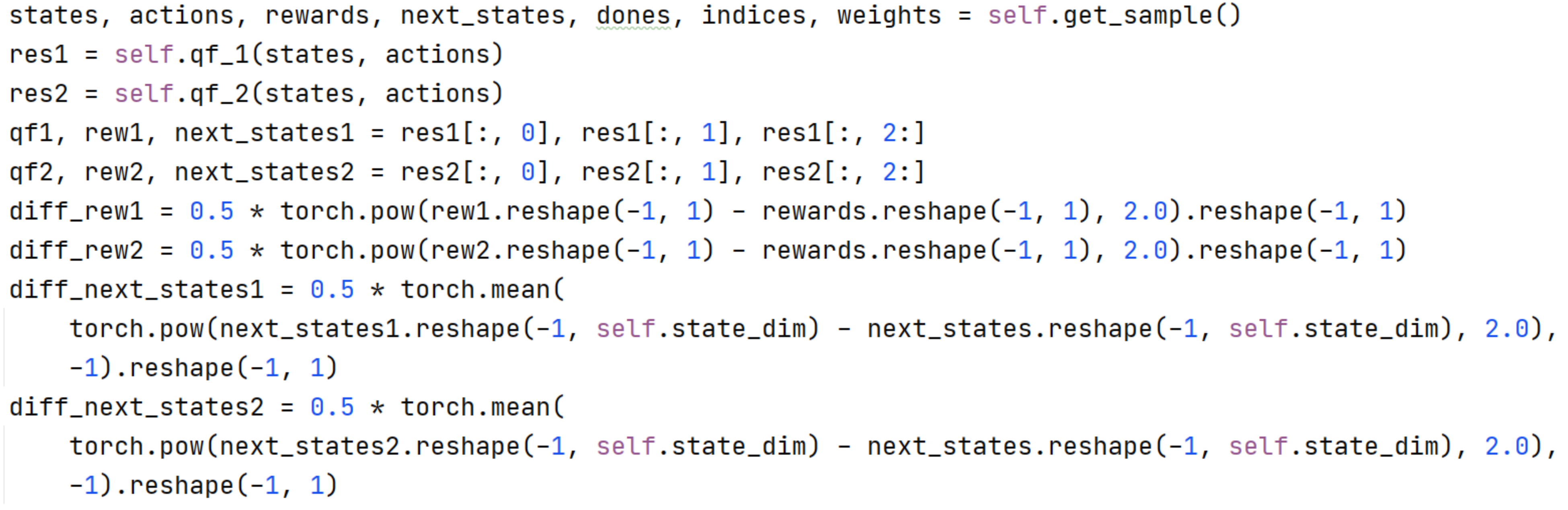}}
\end{tabular}
\caption{Implementation of model errors in MQNs.
}
\label{fig:code2}
\end{figure*}
Then we can formulate MQN's loss by summing model- and TD-errors as in Figure~\ref{fig:code3}. The coefficient of each loss adaptively changes by a dynamic method in \citep{liang2020simple}. Furthermore, we use the quantity to impose priorities to corresponding samples as in Figure~\ref{fig:code4}. These modifications are all for applying model-augmented $Q$-learning (MQL). Note that there is no hyper-parameters change. In a similar manner, we can apply MQL to other algorithms, e.g., TD3 \citep{td3}, and Rainbow \citep{rainbow, efficientrainbow}. In the case of Rainbow, we use the context obtained by convolutional layers. 

\newpage

\section{Additional Experimental Results}
We add efficient-Rainbow \citep{efficientrainbow}'s results on other Atari games. MRainbow-MPER also outperforms baselines in all tested cases. 

\begin{figure*}[ht!]
\centering
\begin{tabular}{cccc}
    \makecell{\includegraphics[width=0.24\columnwidth]{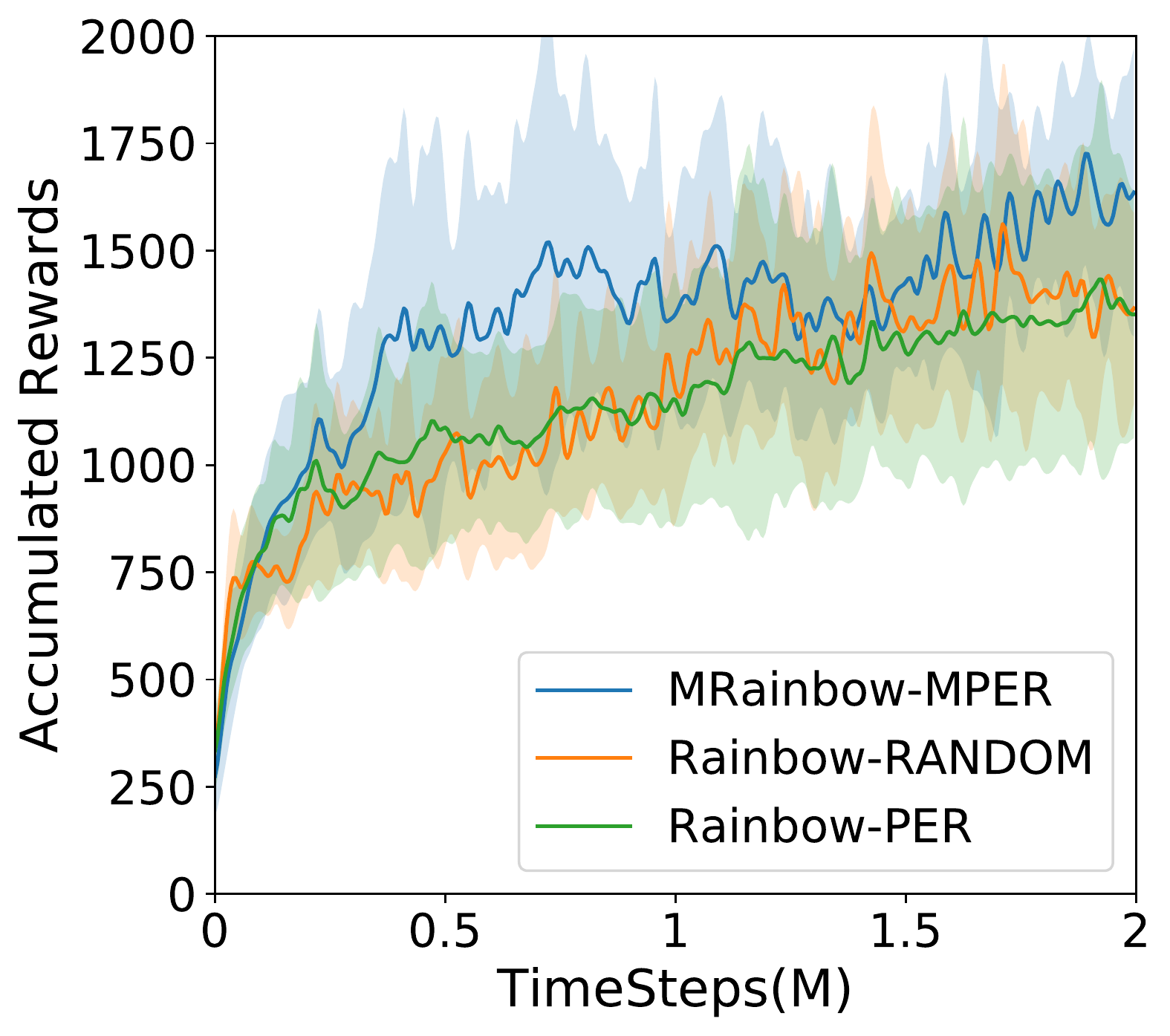}} &
    \makecell{\includegraphics[width=0.24\columnwidth]{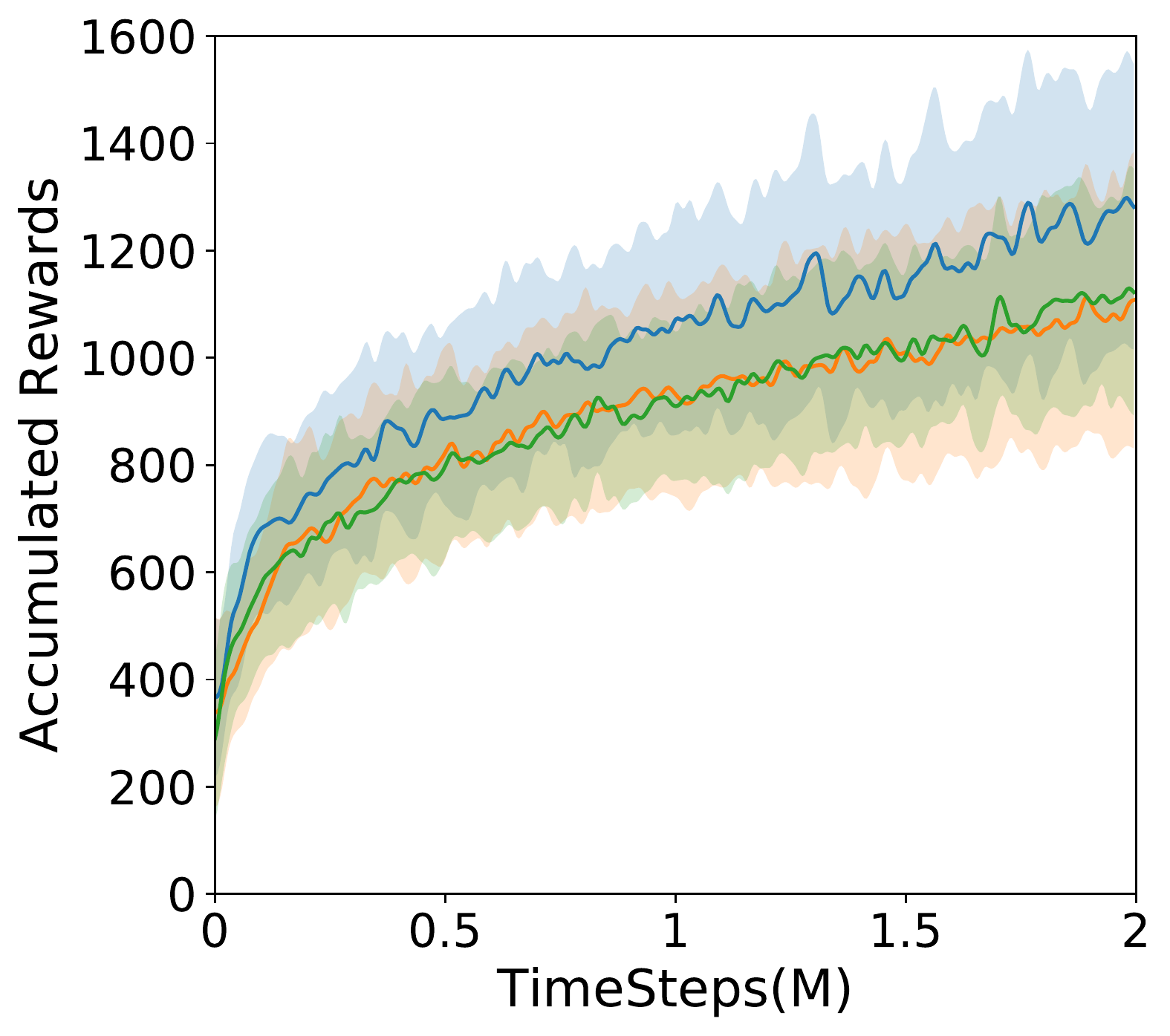}} &
    \makecell{\includegraphics[width=0.24\columnwidth]{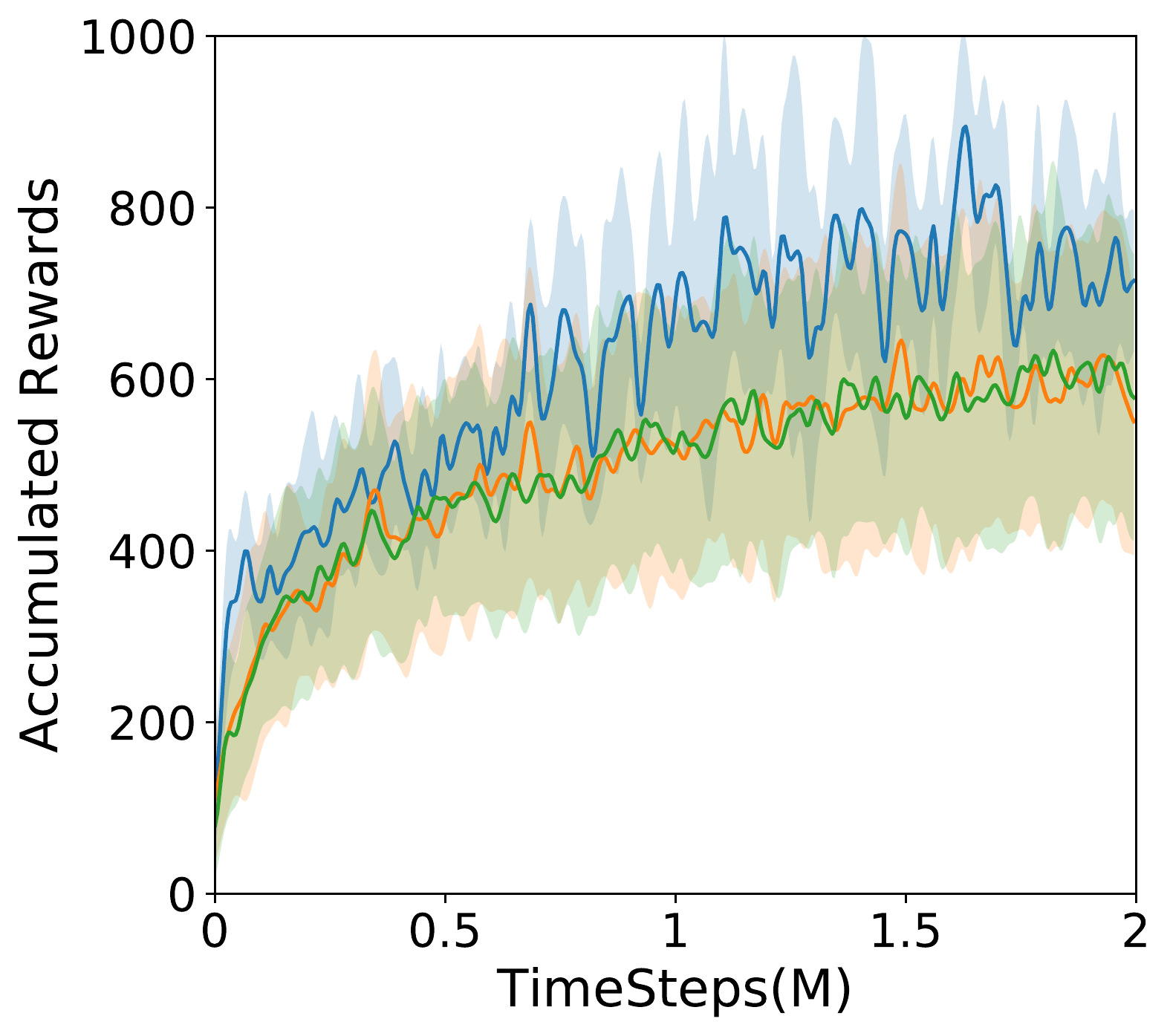}} &
    \makecell{\includegraphics[width=0.24\columnwidth]{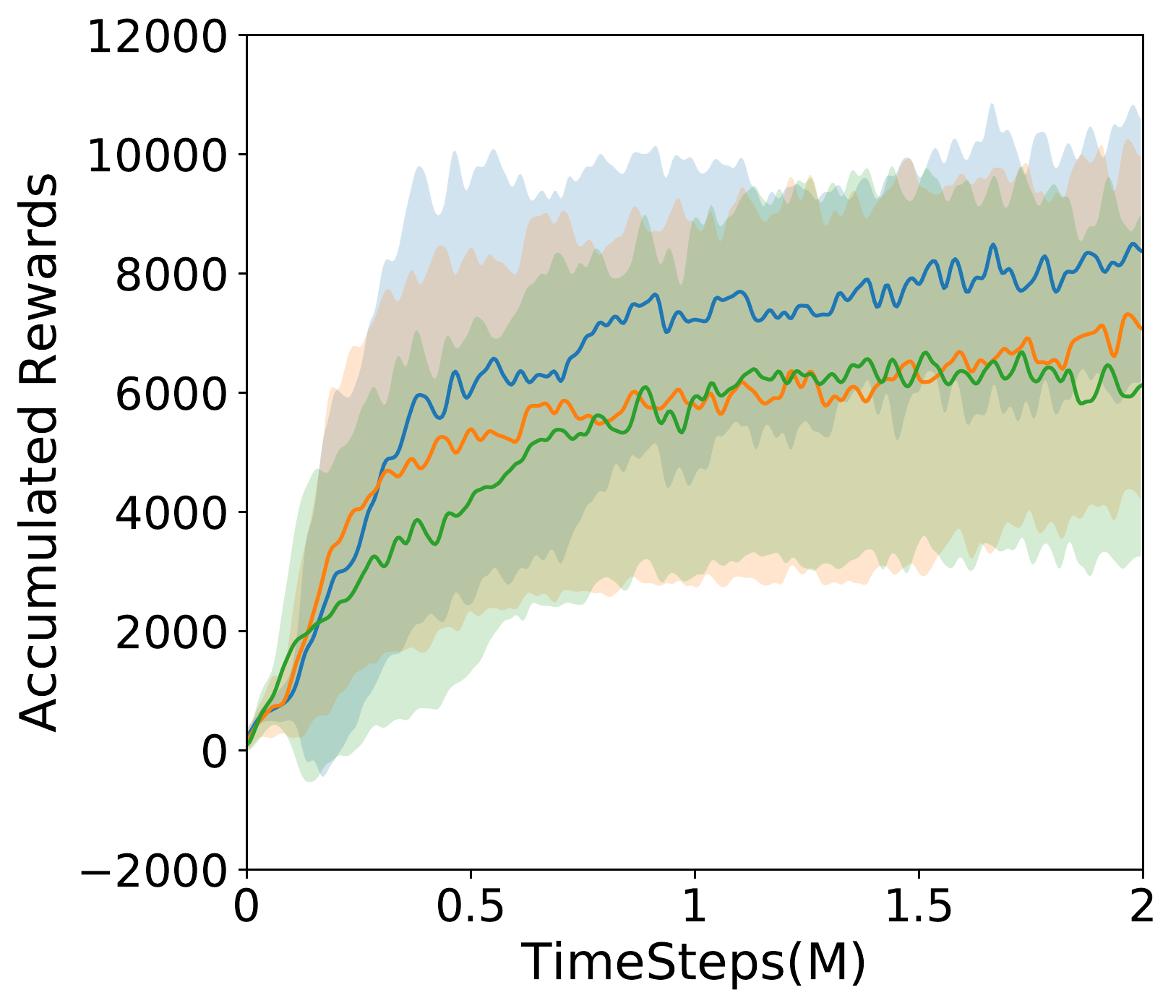}} 
    \\
    (a) Alien & (b) Asterix & (c) Gopher & (d) Kangaroo
\end{tabular}
\caption{Learning curves of Rainbow on Atari games. Here, MRainbow-MPER is a variant of Rainbow, where all of MQN, MPER, and MReward are applied. Although Rainbow basically adopts PER, we denote it as Rainbow-PER for consistency. One can observe that MRainbow-MPER overwhelms other methods.
The solid line
and shaded regions represent the mean and standard deviation, respectively, across five runs  with random seeds. 
}
\label{fig:app-exp1-atari}
\end{figure*}
\begin{figure*}[h!]
\centering
\begin{tabular}{c}
    \makecell{\includegraphics[width=0.6\columnwidth]{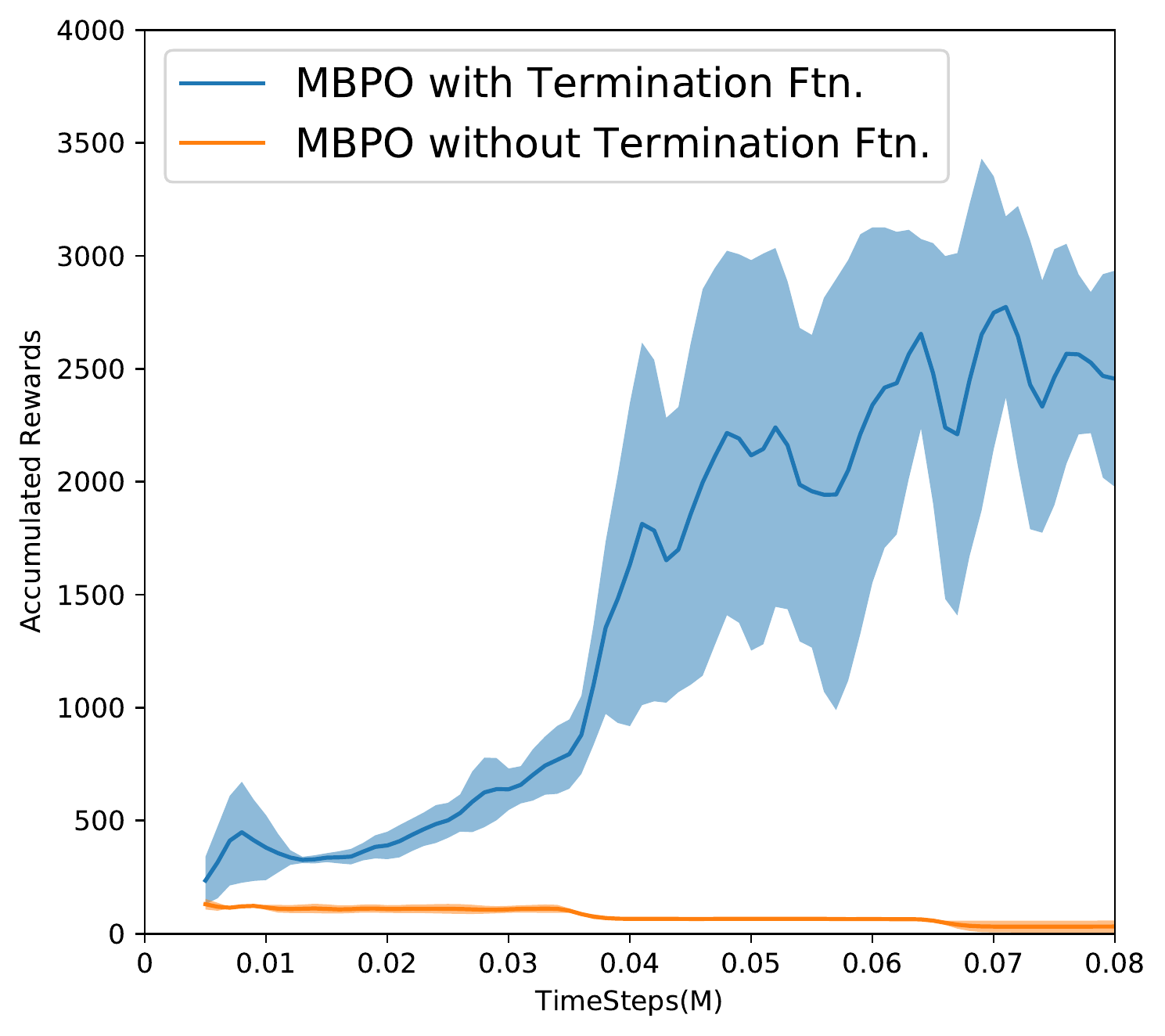}}
    \\
    Hopper-v3
\end{tabular}
\caption{Learning curves of MBPO on Hopper-v3 belonging to MuJoCo environments. One can observe that learning terminal functions is harmful to sample efficiency in MBPO.
The solid line
and shaded regions represent the mean and standard deviation, respectively, across five runs  with random seeds. 
}
\label{fig:MBPO}
\end{figure*}
Moreover, as we mentioned in Section~\ref{sec:related}, to show that prior knowledge is crucial to MBRL, we provide an experimental result about one of the state-of-the-art MBRL methods, i.e., MBPO \citep{mbpo} with and without learning a termination function. To verify it,  we increased the output dimension of an ensemble of dynamic models by one. Then we used the sigmoid function to learn the probability of the termination function with the binary cross-entropy loss. One can observe that the learning termination function is significantly harmful to the sample efficiency compared to the original MBPO in Figure~\ref{fig:MBPO}. The degraded result comes from wrongly computed TD-errors since computing the TD target depends heavily on the termination function.

%% file: 4-related-arxiv.tex
\section{Related Work} \label{sec:related}

\textbf{Reward shaping.} Our reward estimator in the MQN is closely related to the reward shaping methods. The most well-known and frequently used methods for reward shaping are the potential-based reward shaping (PBRS) \citep{ng1999policy} and its variants \citep{wiewiora2003principled, devlin2012dynamic, harutyunyan2015expressing}. \cite{ng1999policy, wiewiora2003principled,
devlin2012dynamic} prove that the policy-invariance is guaranteed even if the reward is modified by potential functions that depend only on states. 
Since reward shaping usually requires heavy prior knowledge of the given tasks, \cite{hu2020learning} proposed to learn the weight of a given potential function, and \cite{zou2019reward} used meta-learning to learn the reward estimator.

\textbf{Model-based RL.}
Although our approach is different from model-based RL algorithms, we briefly introduce some of them. There are different types of Model-based RL (MBRL), and the most popular approach is Dyna-style~\citep{dyna}, in which algorithms generate fictitious experiences to train agents. Although there exists various methods~\citep{metrpo, slbo, mbmpo, mbpo, gametheory} for MBRL, the common strategy is to first learn the environment model and use it to generate fictitious experiences for learning an agent's policy. Due to its ability to generate transitions, MBRL's sample efficiency is remarkable on certain tasks, but they require prior knowledge of environments such as a termination function on state and reward function, for a given state-action pair. Furthermore, MBRL requires much larger computing costs over MFRL algorithms.

\textbf{Off-policy model-free RL.} 
In the case of $Q$-learning without policy newtorks, Rainbow \citep{rainbow} that combines various techniques to extend the original DQN learning \citep{deep_q-learning_with_er} is one of the state-of-the-art methods in Atari game environments. 
Although there is a model-based approach for Atari games \citep{kaiser2019model}, data-efficient Rainbow \citep{efficientrainbow} that we use as a baseline outperforms it.
In the case of the actor-critic architecture, twin delayed
DDPG (TD3)~\citep{td3} and soft actor-critic (SAC)~\citep{sac} are frequently used the state-of-the art methods. TD3 employs double $Q$-networks, target policy smoothing,
and different frequencies to update a policy and $Q$-networks, to reduce overestimation bias. 
SAC also adopts double $Q$-learning and utilize the entropy measure of an agent policy to the reward to encourage the exploration of the agent. 

\textbf{Experience replay.} Prioritized experience replay (PER)  \citep{rainbow,ddpg_per,deep_q-learning_with_per,sac_boosting, PSER} is one of the most frequently used strategies to sample important transitions. Its effectiveness in $Q$-learning is verified on Atari
environments. 
Recently, learning-based sampling, which utilizes neural networks to generate priority scores, have shown to be outperform rule-based PER~\citep{experience_replay_optimization, sinha2020experience}. However, ERO~\citep{experience_replay_optimization}'s gain over PER is not very significant, and \citep{sinha2020experience} requires to determine some hyper-parameters and its effective on discrete control tasks, e.g., Atari games, is unclear.
